%% file: main.tex
\documentclass{ustthesis}

\usepackage{mathpazo,amsmath,amssymb,epsfig,enumerate,bbm,calc,color,ifthen,capt-of} 
\usepackage{algorithm}
\usepackage[center]{subfigure}
\usepackage{color,graphicx}

\usepackage{hyperref} 
\usepackage[margin=25mm,textheight=247mm,textwidth=145mm]{geometry}
\usepackage{times}
\usepackage{latexsym}
\usepackage{mathrsfs}
\usepackage{amsmath}
\usepackage{amssymb}
\usepackage[numbers]{natbib}
\usepackage{booktabs}
\usepackage{graphicx}
\usepackage{multirow}
\usepackage{bibentry} 
\usepackage{hyperref}
\usepackage{ltablex}
\usepackage{CJKutf8}
\usepackage{algpseudocode}
\usepackage{mathrsfs}

\usepackage{physics}
\usepackage{tikz}
\usepackage{mathdots}
\usepackage{yhmath}
\usepackage{cancel}
\usepackage{color}
\usepackage{siunitx}
\usepackage{array}
\usepackage{tabularx}
\usepackage{extarrows}
\usepackage{booktabs}
\usepackage[numbers]{natbib}

\usetikzlibrary{fadings}
\usetikzlibrary{patterns}
\usetikzlibrary{shadows.blur}
\usetikzlibrary{shapes}

\usepackage{pifont}
\newcommand{\cmark}{\ding{51}}
\newcommand{\xmark}{\ding{55}}
\usepackage{natbib}

\usepackage[scaled]{helvet} 
\usepackage{courier} 
\usepackage{eulervm} 
\normalfont
\usepackage[T1]{fontenc}

\newcommand{\tab}[1]{\hspace{3mm}}

\title{Greenformers: Improving Computation and Memory Efficiency in Transformer Models via Low-Rank Approximation}  

\author{Samuel Cahyawijaya}     
\degree{\MPhil}             
\subject{the Department of Electronic and Computer Engineering}      
\department{Electronic and Computer Engineering}       
\advisor{Prof. Pascale FUNG}     
\depthead{Prof. Andrew Wing On POON}    
\defencedate{2021}{08}{24}

\begin{document}


\maketitle

%



\input{chapter/sec-ack}

\tableofcontents


\listoffigures


\listoftables


\input{chapter/sec-abstract}




\input{chapter/sec-1-introduction}
\input{chapter/sec-2-related}
\input{chapter/sec-2.5-low-rank-factorized-transformer}
\input{chapter/sec-3-low-rank-transformer}
\input{chapter/sec-5-linformer-for-genomics}
\input{chapter/sec-6-conclusion}
\input{chapter/sec-publications}

\bibliographystyle{plain}
\bibliography{main}




\end{document}

%% file: chapter/sec-ack.tex
\acknowledgments
I would never have completed this work without the help from many people. First of all, I thank my advisor, Professor Pascale FUNG, for her years of mentoring, advice, and encouragement. I have learned from her how to develop, evaluate, express, and defend my ideas. These skills are important for my later PhD study. I thank the members of my thesis committee, Professor Qifeng Chen, and my thesis chairperson Professor Stuart Gietel-Basten, for their insightful comments on improving this work.

I thank my colleagues in HKUST -- Dr.~Genta Indra Winata, Andrea Madotto, Dai Wenliang, Yu Tiezheng, Xu Yan, Lin Zhaojiang, Zihan Liu, Etsuko Ishii, Yejin Bang, Dr.~Xu Peng, Dr.~Ilona Christy Unarta, Kharis Setiasabda, Bryan Wilie, Karissa Vincentio, Jacqueline Cheryl Sabrina, Darwin Lim, Kevin Chandra, and many others. We have finished a lot of valuable works and develop many insightful ideas altogether. In daily life, we have been very good friends. Without them, my graduate study in HKUST would not be so colorful.  Last but not least, I thank my parents and my brothers, for their support and encouragement along my MPhil study in HKUST.

\endacknowledgments

%% file: chapter/sec-abstract.tex
\begin{abstract}

In this thesis, we introduce Greenformers, a collection of model efficiency methods to improve the model efficiency of the recently renowned transformer models with a low-rank approximation approach. The development trend of deep learning models tends to results in a more complex and larger model. Although it leads to a better and more accurate prediction, the resulting model becomes even more costly, as it requires weeks of training with a huge amount of GPU resources. Particularly, the size and computational cost of transformer-based models have increased tremendously since its first debut in 2017 from $\sim$100 million parameters up to $\sim$1.6 trillion parameters in early 2021. This computationally hungry model also incurs a substantial cost to the environment and even reaches an alarming level of carbon footprint. Some of these models are so massive that it is even impossible to run the model without a GPU cluster.

Greenformers improve the model efficiency of transformer models by applying low-rank approximation approaches. Specifically, we propose a low-rank factorization approach to improve the efficiency of the transformer model called Low-Rank Transformer. We further compare our model with an existing low-rank factorization approach called Linformer. Based on our analysis, the Low-Rank Transformer model is suitable for improving both the time and memory efficiency in processing short-sequence ($\leq512$) input data, while the Linformer model is suitable for improving the efficiency in processing long-sequence input data ($\geq512$). We also show that Low-Rank Transformer is more suitable for on-device deployment, as it significantly reduces the model size. Additionally, we estimate that applying LRT to the existing BERT$_{BASE}$ model can significantly reduce the computational, economical, and environmental costs for developing such models by more than 30\% of its original costs.

Our Low-Rank Transformer can significantly reduce the computational time and memory usage on the speech recognition task. Specifically, our Low-Rank Transformer can halve the size of the model and increase the speed by up to 1.35x in the GPU and 1.25x in the CPU while maintaining the performance of the model compared to the original transformer model. Our finding suggests that transformer models tend to be over-parameterized and our Low-Rank Transformer can help to mitigate the over-parameterization problem, yielding a more efficient model with a better generalization.


Additionally, we extend the possibility of applying a low-rank approximation approach to a genomics study for Alzheimer's disease risk prediction. We apply sequence modeling techniques with the Linformer model to predict Alzheimer's disease in the Chinese cohort. We define our problem as a long sequence classification problem with various lengths up to $\sim$33,000 nucleotides long. Our result shows that Linformer models with Subword Tokenization can process very long sequence data and boost the evaluation performance by up to $\sim$5\% AUC compared to the existing FDA-approved risk scoring model and other deep learning variants. Based on our analysis, we further conclude that the choice of tokenization approach can also provide a huge computation and memory efficiency as much as the efficient model approach, which makes consideration of choosing tokenization approach more prominent for developing a more efficient transformer model.

\end{abstract}

%% file: chapter/sec-1-introduction.tex
\chapter{Introduction}\label{sec-introduction}

\section{Motivation and Research Problems}

Starting from AlexNet~\cite{krizhevsky2012alexnet} in 2012, deep learning models such as convolution neural network, recurrent neural network, and transformer have made significant progression in various fields. Along with its remarkable adoption and growth, the computational cost required for developing a deep learning model also rises significantly at an unprecedented rate. From 2012 to 2018, the computational cost required is estimated to increase by 300,000x~\cite{schwartz2019green}. From 2018 onward, the development of the transformer-based NLP model has shown an even sharper trend. Starting with ELMo ~\cite{peters2018elmo} with ~100M parameters in 2018, followed by BERT ~\cite{devlin2019bert} with ~340M parameters and ~\cite{radford2019gpt2} with ~1.5B parameters in 2019. Recently, two other gigantic models have been released: 1) GPT-3~\cite{brown2020gpt3} with ~175B parameters and 2) Switch Transformer~\cite{fedus2021switch} with ~1.6T parameters. This enormous model size requires a huge amount of computational cost. This computationally hungry model also incurs a substantial cost to the environment and even reaches an alarming level of carbon footprint ~\cite{strubell2019energy}. Some of these models are so massive that it is even impossible to run the model in real-time without a GPU cluster.

As shown in Table~\ref{tab:cost-of-transformer}, the growing trend of transformer-based models is so massive. Within just 3 years, the computational cost for training the most massive transformer-based model has increase by around 20,000 times from 0.181 petaflop/s-day for training the Transformer$_{BIG}$~\cite{vaswani2017transformer} model to 3,640 petaflop/s-day for training the GPT-3~\cite{brown2020gpt3} model. This enormous computational cost leads to a massive increase in terms of computation cost, economic cost, and CO$_2$ emission. For instance, in 2017, the price of developing the original transformer models is less than \$1,000 USD with less than 100 kg of CO$_2$ emission. While in 2020, GPT-3 model costs around  \$4,600,000 USD with about 552 tons of CO$_2$ emission. This massive growth of computational requirement of developing a transformer-based model is concerning and has attracted considerable attention in recent years.

\begin{table}[!t]
\centering
\resizebox{0.85\textwidth}{!}{
    \begin{tabular}{lcccc}
    \toprule
    \multirow{2}{*}{\textbf{Model}} & \textbf{Release} & \textbf{Compute Cost} & \textbf{Economical Cost} & \textbf{CO$_2$ emission} \\
     & \textbf{Year} & \textbf{(petaflop/s-day)} & \textbf{(USD)} &  \textbf{(kg)} \\
    \midrule
    Transformer$_{BASE}$ & 2017 & 0.008$^1$ & \$41 - \$140$^4$ & 11.8$^4$ \\
    Transformer$_{BIG}$ & 2017 & 0.181$^1$ & \$289 - \$981$^4$ & 87.1$^4$ \\
    BERT$_{BASE}$ & 2018 & 2.24$^2$ & \$2,074 - \$6,912$^4$ & 652.3$^4$ \\
    BERT$_{LARGE}$ & 2018 & 8.96$^2$ & \$8,296 - \$27,648$^\star$ & 2,609.2$^\star$ \\
    GPT-2 (1.5B) & 2018 & 10 - 100$^3$ & \$12,902 - \$43.008$^4$ & N/A \\
    GPT-3 & 2020 & 3,640$^3$ & \$4,600,000$^\dagger$ & 552,000$^4$ \\
    \bottomrule
    \end{tabular}
}
\caption[Cost of training a transformer-based model]{Cost of training a transformer-based model. $^1$Hernandez et al (2020)~\cite{hernandez2020measuring}. $^2$Aßenmacher et al. (2019)~\cite{abenmacher2020comparability}. $^3$Brown et al (2020)~\cite{brown2020gpt3}. $^4$Strubell et al (2019)~\cite{strubell2019energy}. $\star$ is estimated based on the computational cost comparison to the BERT$_{BASE}$ model. $\dagger$ is retrieved from \footnotemark.}
\label{tab:cost-of-transformer}
\end{table}

\footnotetext{https://lambdalabs.com/blog/demystifying-gpt-3/}
Several responses have been made to address this problem and raise people's awareness to improve the efficiency of a deep learning model and reduce the overall carbon footprint. SustainNLP is a shared task ~\cite{wang2020sustainnlp} released with the goal of building energy-efficient solutions for the NLP model. Schwartz et al. ~\cite{schwartz2019green} explored a methodology to measure efficiency and introduce the term Green AI which refers to AI research that focuses on increasing computational efficiency. Dodge et al. (2019) ~\cite{dodge2019show} introduces a method for estimating the model performance as a function of computation cost. Strubell et al. (2019) analyzes the expected carbon footprint of NLP research and provide actionable suggestions to reduce computational costs and improve equity. Hernandez et al. (2020) ~\cite{hernandez2020measuring} shows the importance of both hardware and algorithmic efficiency in reducing the computational cost of developing a deep learning model and provide recommendations for reporting modeling efficiency in AI research. Recent benchmarks\cite{tay2021lra,wilie2020indonlu,cahyawijaya2021indonlg} are also considering the model efficiency as one of the metrics to compare the models listed on the benchmarks.

Different efforts have been done to improve the hardware and algorithm efficiency of developing a deep learning model. As shown in Table 2 in Hernandez et al. (2020) ~\cite{hernandez2020measuring}, 44x less computation cost is required to achieve the same level of performance as AlexNet. While the architectural shift from recurrent neural network models such as Seq2seq~\cite{ilya2014seq2seq} and GNMT~\cite{johnson2017googles} to transformer model~\cite{vaswani2017transformer} leads to an increase of computational efficiency by 61x and 9x respectively. All these improvements are made possible through numbers of efficiency methods such as distillation~\cite{hinton2015distil,zhang020ternarybert,sun2020mobilebert}, pruning~\cite{lecun1990obd,hasibi1993sodpruning,hao2016pruning,han2015learningprune,frankle2019lottery,xu2021networkpruning}, quantization~\cite{jacob2018quantization,zhang020ternarybert,bai2020binarybert}, fast adaptation ~\cite{finn2017maml, winata2020fa-casr, winata2020mtl}, data cleansing~\cite{lee2019denoise, kobayashi2020influence, bang2021fakenews}, distributed training~\cite{fairscale2021,rasley2020deepspeed,rajbhandari2020zero}, mixed-precision~\cite{micikevicius2018mixed}, and efficient model~\cite{shen2021efficientattn,dai2021mesm, winata2020lrt, winata2021mme,tan2020efficientnet,wang2020linformer,katharopoulos2020lineartrf,choromanski2021performer}.

\begin{table}[!t]
\centering
\resizebox{0.85\textwidth}{!}{
    \begin{tabular}{lcccc}
    \toprule
    \multirow{2}{*}{\textbf{Method}} & \textbf{Require} & \textbf{Training} & \textbf{Fine-Tuning} & \textbf{Inference} \\
     & \textbf{Prior Model} & \textbf{Efficient} &  \textbf{Efficient} & \textbf{Efficient} \\
    \midrule
    Distillation & \cmark & \xmark & \xmark / \cmark & \cmark \\
    Pruning & \cmark & \xmark & \xmark / \cmark & \cmark \\
    Quantization & \cmark & \xmark & \xmark & \cmark \\
    Fast Adaptation & \xmark / \cmark & \xmark & \cmark & \xmark \\
    Data Cleansing & \xmark & \cmark & \cmark & \xmark \\
    Distributed Training & \xmark & \cmark & \cmark & \xmark \\
    Mixed-Precision & \xmark & \cmark & \cmark & \cmark \\
    Efficient Model & \xmark & \cmark & \cmark & \cmark \\  \bottomrule
    \end{tabular}
}
\caption{Comparison of different efficiency methods}
\label{tab:computing-efficiency}
\end{table}

As shown in Table \ref{tab:computing-efficiency}, depending on when it is applied, an efficiency method can provide different effects to the computational cost on different modeling phases. Distillation, pruning, and quantization can reduce the computational cost drastically during the inference phase, but these methods require having a prior model which makes the overall training cost higher. For distillation and pruning, the fine-tuning process could also become more efficient as the distillation and pruning can be applied beforehand. Mixed-precision and efficient models decrease the computational cost during both training, fine-tuning, and inference. Nevertheless applying mixed-precision during inference might produce a slight inconsistent prediction as a result of rounding error. Unlike the other approaches, fast adaptation, data cleansing, and distributed training do not reduce the computational cost of the model, but these approaches can reduce the actual training time during the training and/or fine-tuning in different ways. Fast adaptation methods allow the model to learn a representation that can quickly adapt to a new task, hence requiring only a fraction of data in the fine-tuning phase. Data cleansing makes training and fine-tuning phases more efficient by reducing the number of samples to be trained by the model. Recent development in distributed training \cite{fairscale2021, rasley2020deepspeed, rajbhandari2020zero} allows better resource allocation and distribution strategy which ends up with significant memory reduction and faster training time.

In this thesis, with the spirit to alleviate the enormous cost of developing transformer-based models, we introduce Greenformers. Greenformers is a collection of efficient methods for improving the efficiency of the transformer model in terms of computation cost, memory cost, and/or the number of parameters. We focus our Greenformers on improving the transformer model efficiency with low-rank approximation methods as low-approximation can not only greatly improve both computational and memory efficiency, but also reducing the number of the model parameters significantly. Specifically, we explore two different two low-rank model variants to increase the efficiency of a transformer-based model: 1) low-rank factorized linear projection transformer called low-rank transformer (LRT)~\cite{winata2020lrt} and 2) low-rank factorized self-attention transformer called Linformer~\cite{wang2020linformer}. For LRT, we conduct our experiment on speech recognition tasks where we utilize low-rank approximation to factorize the model parameters. With this approach, the model size and computational cost can be decreased significantly, yielding a more efficient speech recognition model. For Linformer, we conduct our experiment on long-sequence genomics data. Specifically, we experiment on Alzheimer's disease risk prediction in the Chinese cohort. In this task, given a long genomics sequence, our goal is to predict the risk of the subject of getting Alzheimer's disease. We formulate this task as a sequence classification task with an input sequence length of $\sim$30,000 long. We evaluate our approach by comparing the result with the existing FDA-approved risk-scoring biomarkers. Additionally, we conduct deeper analysis to analyze the efficiency of our low-rank transformer variants compared to the original transformer model and provide insights on which low-rank variant works best given a certain input characteristic.

\section{Thesis Outline}

The contents of this thesis are focused on the application of efficient modeling techniques for transformer models via low-rank approximation. The rest of the thesis is divided into four chapters and organized as follows:

\begin{itemize}
    \item Chapter 2 (Preliminaries and Related Work) introduces the background and important preliminaries and related works on transformer model, efficiency methods, and low-rank matrix approximation.
    \item  Chapter 3 (Efficient Transformer Model via Low-Rank Approximation) presents two low-rank transformer variants that reduce the memory usage and computational cost of the original transformer model.
    \item  Chapter 4 (Low-Rank Transformer for Speech Recognition) shows the effectiveness of low-rank transformer models in speech recognition tasks.
    \item  Chapter 5 (Linformer for Alzheimer's Disease Risk Prediction) shows the applicability of the efficient Linformer model on long-genome understanding tasks for predicting Alzheimer's disease in the Chinese cohort.
    \item Chapter 6 (Conclusion) summarizes this thesis and the significance of the low-rank approximation in transformer models and discusses the possible future research directions.
\end{itemize}

\newpage

%% file: chapter/sec-2-related.tex
\chapter{Preliminaries and Related Work}
\label{sec-preliminaries}

\section{Transformer Model}

\begin{figure}[!ht]
  \centering
  \includegraphics[width=.75\linewidth]{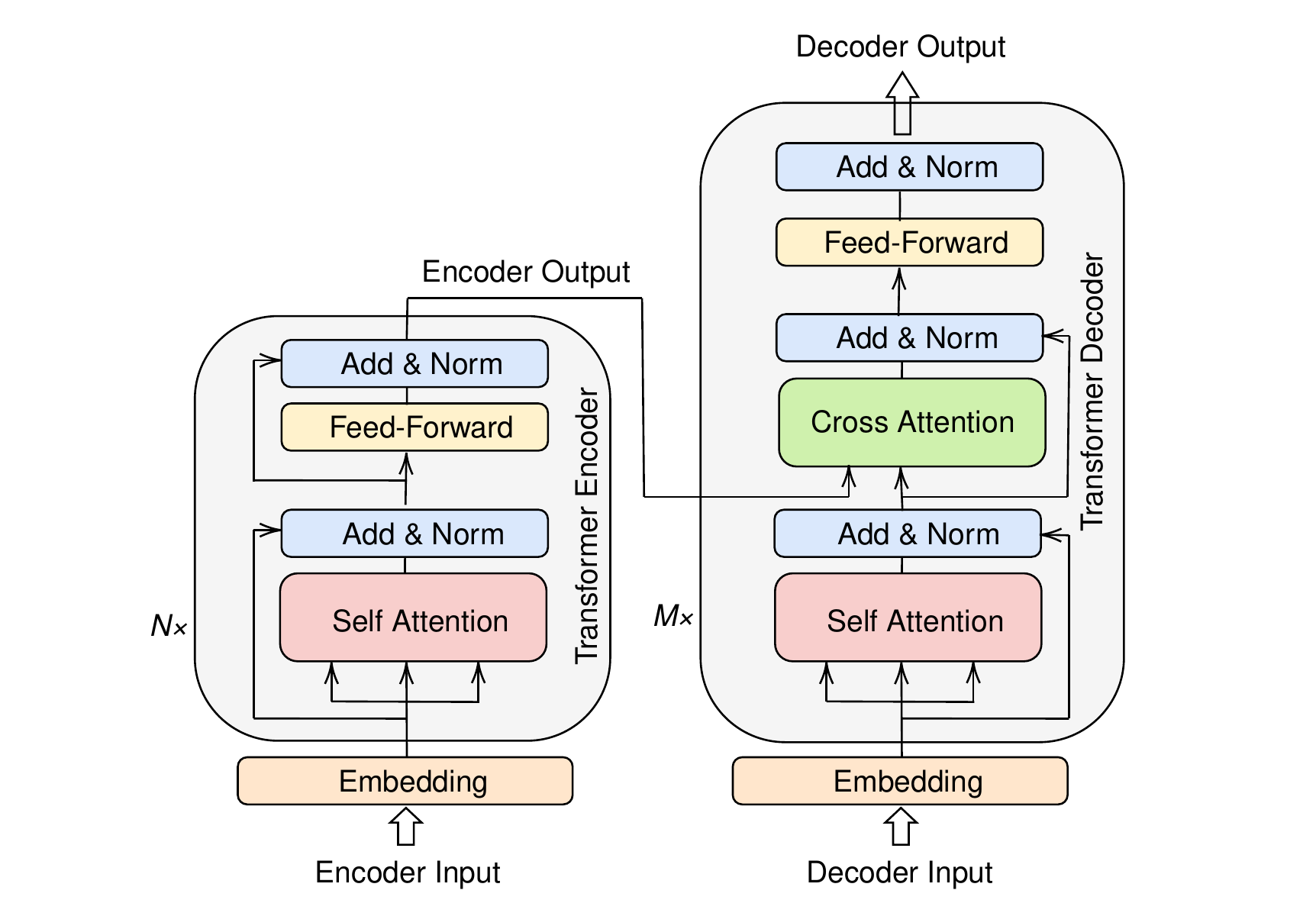}  
  \caption[Transformer model encoder-decoder architecture]{Transformer model encoder-decoder architecture.}
  \label{fig:transformer}
\end{figure}

Transformer is a neural network architecture based on the attention mechanism. Compared to the recurrent neural network (RNN) \cite{graves2016rnn}, the transformer model can process a sequence in parallel which enables the model to take full advantage of modern fast computing devices such as TPUs and GPUs. A transformer model consists of a stack of transformer layers where each layer consists of a self-attention and a position-wise feed-forward network. To avoid gradient vanishing two residual connections are added, one after the self-attention and another one after the feed-forward network. Normalization layer is also added after the residual connection to stabilize the hidden state which allows the model to converge faster~\cite{ba2016layernorm}. The aforementioned transformer layer is known as a transformer encoder layer. There is another kind of transformer layer, i.e, transformer decoder layer, which is very similar to transformer encoder layer with an additional cross-attention in between the self-attention and the feed-forward network. The depiction of transformer encoder layer, transformer decoder layer, and the two types of layer interact is shown in Figure~\ref{fig:transformer}.


\subsection{Transformer Components}

\paragraph{Scaled Dot-Product Attention}
The attention-mechanism in the Transformer is computed with scaled dot-product attention~\cite{vaswani2017transformer,winata2021multilingual}. The scaled dot-product attention accepts a query sequence $Q \in \mathbb{R}^{N \times d_k}$, a key sequence $K \in \mathbb{R}^{N \times d_k}$, and a value sequence $V \in \mathbb{R}^{N \times d_v}$ as its input, and produce an output $O \in \mathbb{R}^{N \times d_v}$. Scaled dot-product attention is done by first finding the similarity of $Q$ and $K$ with a dot-product operation scaled with a factor of $\frac{1}{\sqrt{d_k}}$ to reduce the magnitude, and then apply a softmax operation to get the probability distribution over different locations. The probability distribution is then used as the attention weight of $V$ to get the output sequence $O$. Scaled-dot product attention can be formulated as:

\begin{equation}
Attention(Q,K,V) = softmax(\frac{QK^T}{\sqrt{d_k}})V
\end{equation}

\begin{figure}[!t]
  \centering
  \includegraphics[width=.65\linewidth]{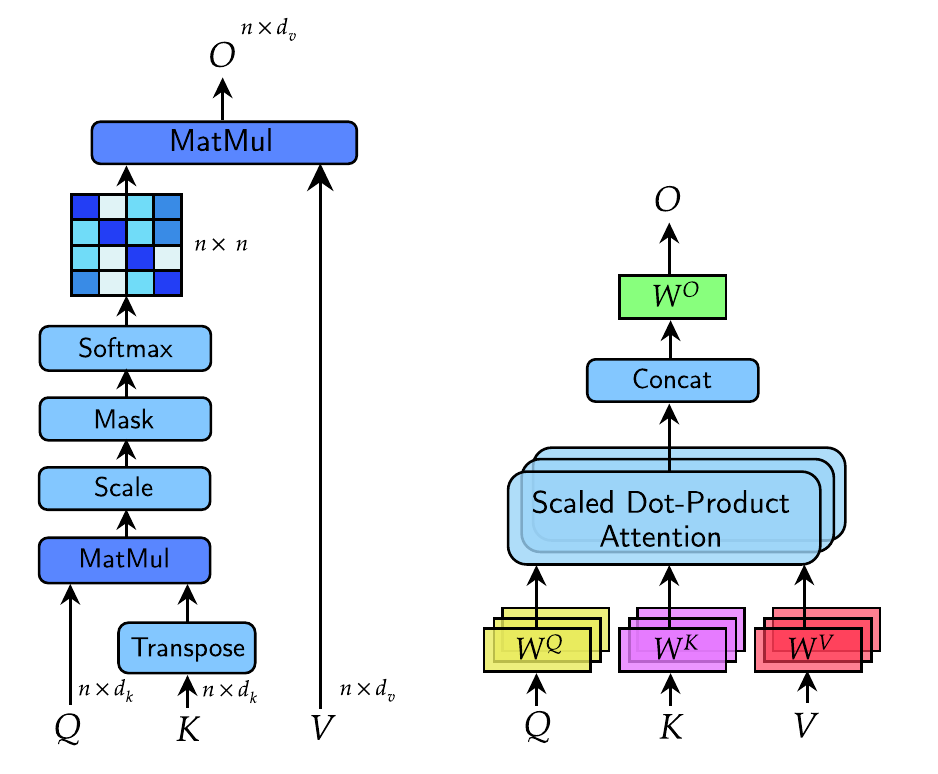}  
  \caption[Illustration of scaled dot-product attention and multi-head attention]{Illustration of scaled dot-product \textbf{Left} attention and multi-head attention \textbf{Right}}
  \label{fig:attention}
\end{figure}

\paragraph{Multi-Head Attention} The attention in Transformer is also multi-headed. Multi-head attention split the $d_k$ and $d_v$ in $Q$, $K$, and $V$ into multiple heads $h$ with equal size. For each head, the scaled dot-product attention is applied. The results are then concatenated and projected to get the output $O$. Unlike single-head attention, Multi-head attention allows the model to jointly attend to information from different representation subspaces at different positions~\cite{vaswani2017transformer}. The depiction of scaled dot-product attention and multi-head attention is shown in Figure~\ref{fig:attention}.

\paragraph{Position-wise Feed Forward Network} Position-wise feed-forward network is computed after the self-attention in the transformer encoder layer and the cross-attention in the transformer decoder layer. Position-wise feed-forward network consists of two linear layers that are applied to each position and with an activation function in between. The original transformer model uses ReLU~\cite{nair2010relu} as the activation function, while more recent pre-trained transformer model such as BERT~\cite{devlin2019bert}, GPT-2~\cite{radford2019gpt2}, and GPT-3~\cite{brown2020gpt3} use GELU~\cite{hendrycks2016gelu} as their activation function which proven to yield a better evaluation performance. Position-wise feed-forward network can be formulated as:

\begin{equation}
FFN(x) = Act(xW_1 +b_1)W_2 +b_2
\end{equation}

where $x$ denotes the input vector, $Act$ denotes the activation function, $W_1$ and $b_1$ denote the weight and bias of the first linear layer, and $W_2$ and $b_2$ denote the weight and bias of the second linear layer.

\subsection{Transformer Layers}
\paragraph{Transformer Encoder and Transformer Decoder} There are two types of transformer layers: 1) Transformer encoder and 2) Transformer decoder. The transformer encoder layer process the input sequence $X^{enc}$ in a non-autoregressive way and produce an output sequence $Y^{enc}$. This allows the transformer encoder layer to be computed in parallel over different sequence positions during both the training and inference. While the Transformer decoder layer process the input sequence $X^{dec}$ in an autoregressive way which makes the inference step should run sequentially as it produces output for one position for each time step. While the training process can be done in parallel by performing autoregressive masking when the attention is computed.

\paragraph{Self-Attention and Cross-Attention} As shown in Figure~\ref{fig:transformer}, the transformer encoder layer only has a self-attention while the transformer decoder layer consists of two different kinds of attention, i.e, self-attention and cross-attention. On the self-attention mechanism, the $Q$, $K$, and $V$ sequences are generated by a learned projections weight $W^Q$, $W^K$, and $W^V$ from the input sequence, respectively. While in the cross-attention, the query sequence $Q$ is generated from the decoder input $X^{dec}$ and the key and value sequences, $K$ and $V$, are generated from the encoder output $Y^{enc}$.

\section{Efficient Transformer}

\begin{figure}[!ht]
  \centering
  \includegraphics[width=1\linewidth]{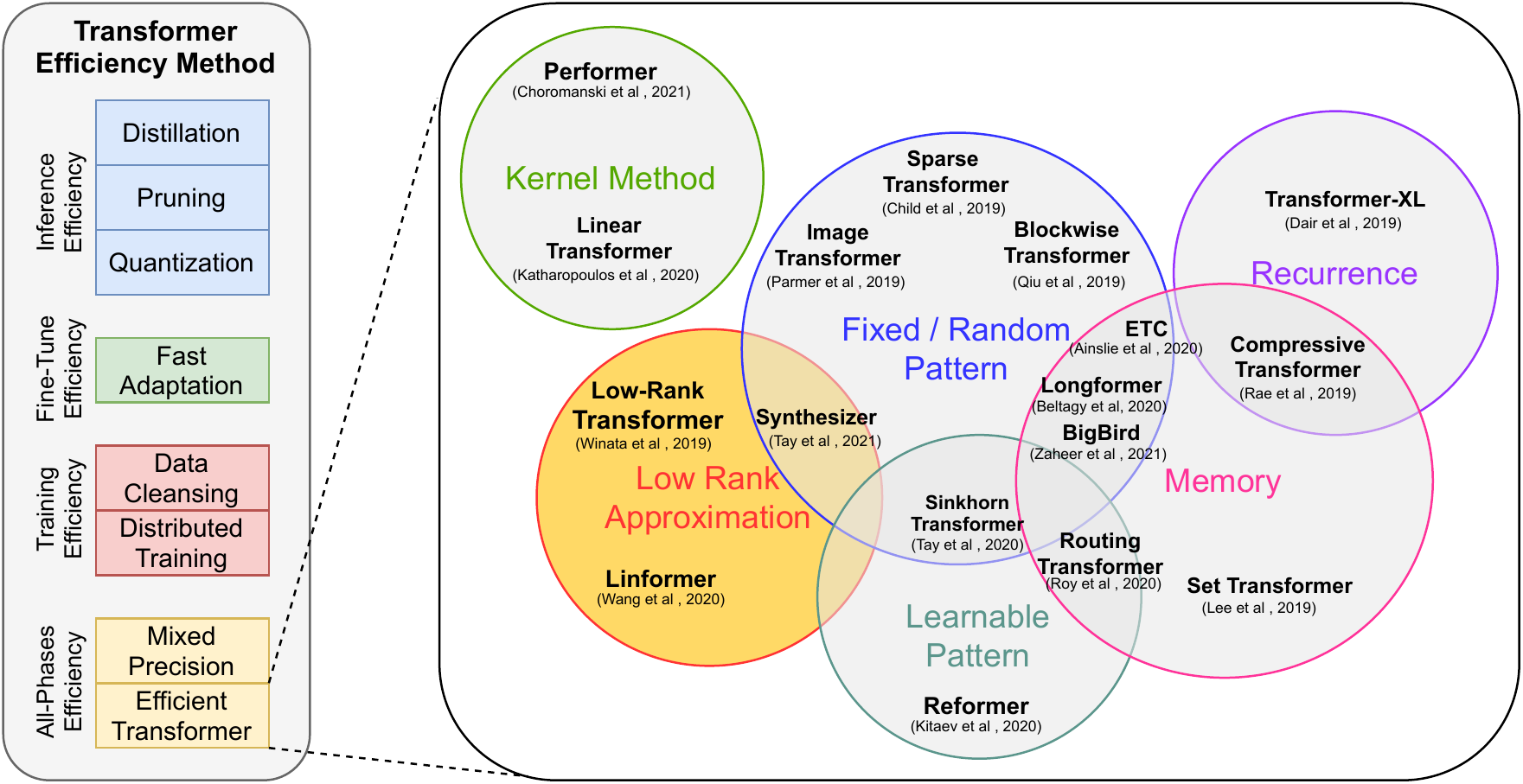}  
  \caption[Efficiency Methods for Transformer]{Efficiency Methods for Transformer. In this thesis we focus on architecture-specific low-rank approximation method for transformer model.}
  \label{fig:transformer}
\end{figure}

Transformer model has shown a great performance in many different fields~\cite{dai2021mesm,madotto2020learning, winata2021multilingualcodeswitch,winata2021multilingual,armar2018imagetrf}. Aside from its great progression, the improvement usually comes with an increase in the size of the model which yields a much larger model and more expensive computation and memory requirements~\cite{devlin2019bert,radford2019gpt2,shoeybi2020megatronlm,fedus2021switch,brown2020gpt3}. Multiple efficiency methods have been developed to improve the efficiency of a deep learning model. Some efficiency methods are architecture-independent, while some others are more specific. 
Several efficiency methods can also be used in conjunction with other types of efficiency methods. In the following section, we will briefly describe each of the general efficiency methods and then provide a deeper explanation of architecture-specific transformer efficiency methods.

\subsection{General Efficiency Methods}
As shown in Figure~\ref{tab:efficiency}, in general, an efficiency method can be divided into four categories based on where the efficiency takes place: 1) inference efficiency, 2) fine-tuning efficiency, 3) training efficiency, and 4) All-phases efficiency.

\paragraph{Inference efficiency}

\begin{figure*}[ht!]
    \centering
    \resizebox{1.0\textwidth}{!}{  
        \includegraphics{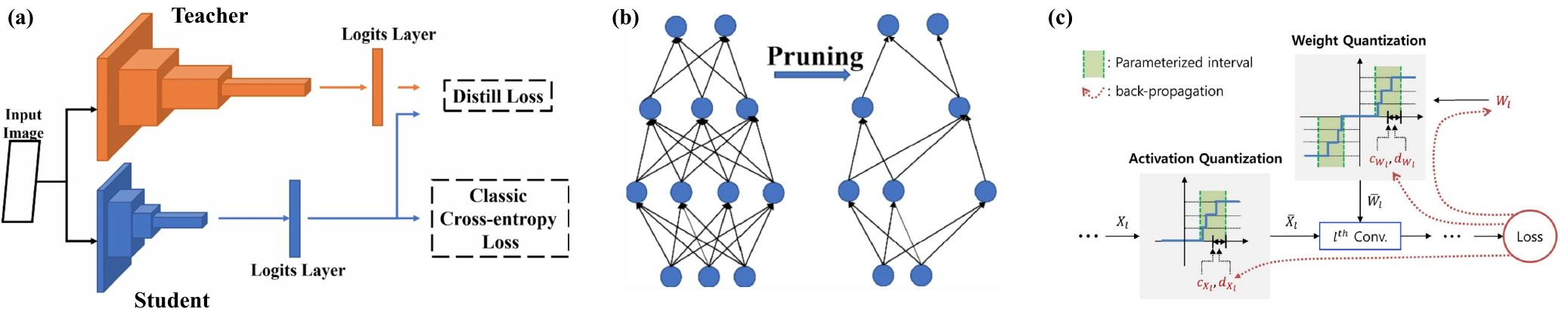}
    }
    \caption[Methods for inference efficiency]{Methods for inference efficiency. (a) Distillation~\cite{chen2021prunedistil}. (b) Pruning~\cite{chen2021prunedistil}, (c) Quantization~\cite{jung2019quantize}}
    \label{fig:inference_efficiency}
\end{figure*}

Inference efficiency methods such as distillation~\cite{hinton2015distil,sanh2019distilbert,sun2020mobilebert,chen2021prunedistil}, pruning~\cite{hao2016pruning,xu2021networkpruning,hasibi1993sodpruning,reed1993pruning,han2015learningprune,han2015learningprune,chen2021prunedistil}, and quantization~\cite{jacob2018quantization,zhang020ternarybert,bai2020binarybert,jung2019quantize} can be used for improving the efficiency during the inference phase by reducing the model size during the fine-tuning process. Recent approaches ~\cite{sun2020mobilebert,xu2021networkpruning} introduce a task-agnostic distillation and pruning which can further improve the efficiency during the fine-tuning phase. Distillation reduces the model size by generating a smaller student model which is taught by the pre-trained teacher model. Pruning reduces the model size by removing unimportant connections according to a criterion such as based on its magnitude~\cite{reed1993pruning}. Quantization decreases the model size by quantizing the 32-bit floating-point parameters into a lower bit-depth such as 8-bit integer quantization~\cite{jacob2018quantization}, 3-bit quantization in Ternary BERT~\cite{zhang020ternarybert}, and 2-bit quantization in Binary BERT~\cite{bai2020binarybert}. The illustration for distillation, pruning, and quantization is shown in Figure~\ref{fig:inference_efficiency}.

\paragraph{Fine-tuning efficiency}
Although the goal of fast adaptation or few-shot learning methods~\cite{finn2017maml,winata2020fa-casr,winata2020mtl} is to improve model generalization of the model, these approaches also help to improve the efficiency on the fine-tuning phase as it only allows the model to be trained with only a tiny amount of data during the fine-tuning. Fast adaptation is done by learning a representation that can generalize well over different tasks by optimizing the model on multiple tasks with a meta-learning approach~\cite{finn2017maml}. Nevertheless, this method requires building a prior model which incurs some additional cost to build the model beforehand.

\begin{figure*}[t!]
    \centering
    \resizebox{1.0\textwidth}{!}{  
        \includegraphics{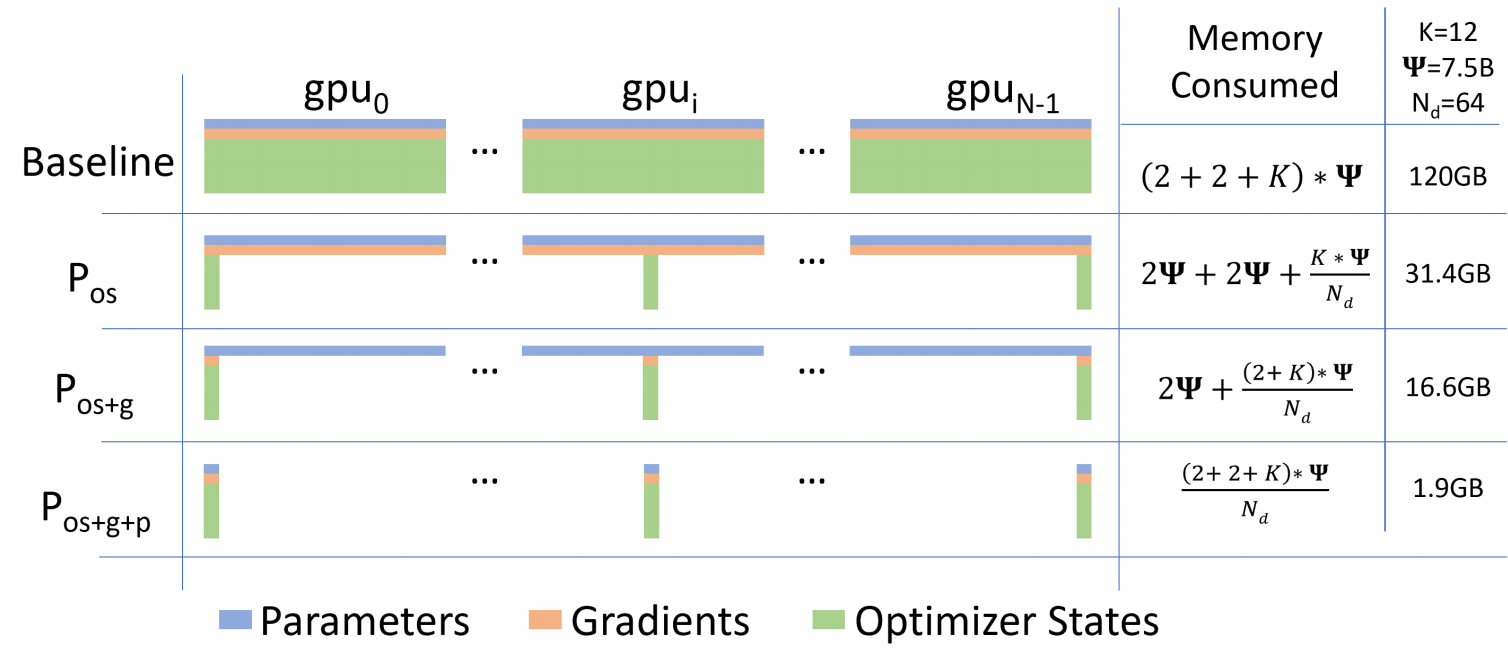}
    }
    \caption[Comparison of different DeepSpeed ZeRO allocation strategies]{Comparison of different DeepSpeed ZeRO allocation strategy with the distributed training baseline~\cite{rajbhandari2020zero}}
    \label{fig:distributed-efficient}
\end{figure*}

\paragraph{Training efficiency}
Training efficiency methods improve the model efficiency in both pre-training and fine-tuning phases. There are two different approaches to achieve training efficiency: 1)
data cleansing and 2) distributed training. Data cleansing approaches~\cite{lee2019denoise, kobayashi2020influence, bang2021fakenews} improve the efficiency during the training and fine-tuning phase by removing data point that is considered as unimportant based-on a certain criterion. While, recent distributed training approaches \cite{fairscale2021, rasley2020deepspeed, rajbhandari2020zero} allow better resource allocation and distribution strategy which significantly reduces the memory usage and enables faster training and fine-tuning. Figure~\ref{fig:distributed-efficient} shows the distributed training allocation with different DeepSpeed ZeRO~\cite{rajbhandari2020zero} configurations compared to the normally distributed training allocation approach.

\paragraph{All-Phases efficiency} 
There are two methods that can provide efficiency across all phases, i.e,  mixed-precision and efficient model. mixed-precision~\cite{micikevicius2018mixed} is mostly used to decrease the computational cost mainly during both training, fine-tuning by reducing the bit-depth of the model parameter similar to the quantization approach. But unlike quantization which reduces the bit-depth from 32-bit floating point to lower bit integer, mixed-precision only reduces the precision of floating-point from 32-bit to 16-bit and only changes the bit-depth on certain layer types. Although mixed-precision is mainly used only on training and fine-tuning, It can also be applied on the inference phase, although it might yield an erroneous prediction due to the effect of rounding error. While, model efficiency can describe any architectural model efficiency methods such as sparse computation, low-rank approximation, kernel methods, and many others. A more specific description of the transformer model efficiency approach is elucidated in Section~\ref{sec-efficient-transformer}.

\subsection{Efficient Transformer}
\label{sec-efficient-transformer}

In the recent years, many research works have tried to build an efficient variant of the transformer model. Extending from Tay et al. (2020)~\cite{tay2020efficient}, we categorized different model-based efficiency approaches for transformer model into six categories as shown in Figure~\ref{tab:efficiency}, i.e, kernel method, fixed/random pattern, learnable pattern, recurrence, memory, and low-rank approximation. 

\paragraph{Kernel Method} 
Kernel methods~\cite{li2019speechtransformer,katharopoulos2020lineartrf} reformulate the scaled dot-product attention mechanism by applying kernel tricks which allows the model to avoid a costly $N \times N$ softmax operation. Kernel methods rewrite the scaled dot-product attention in the following way:

\begin{align}
    &Attention(Q,K,V) = softmax(\frac{QK^T}{\sqrt{d_k}})V\\
    &Attention(Q,K,V)_k = \frac{\sum_{i=1}^{N}{sim(Q_k,K_i^T)V_i}}{\sum _{j=1}^{N}{sim(Q_k,K_j)^T\sqrt{d_k}}} \\
    &Attention(Q,K,V)_k = \frac{\sum_{i=1}^{N}{\phi(Q_k)\phi(K_i^T)V_i}}{\sum _{k=1}^{N}{\phi(Q_k)\phi(K_j^T)\sqrt{d_k}}} \label{eq:kernel} \\
    &Attention(Q,K,V)_k = \frac{\phi(Q_k)\sum_{i=1}^{N}{\phi(K_i^T)V_i}}{\phi(Q_k)\sqrt{d_k\sum_{j=1}^{N}{\phi(K_j^T)}}} \\
    &Attention(Q,K,V) = \frac{\phi(Q)(\phi(K^T)V)}{\phi(Q)\sqrt{d_k}\sum_{j=1}^{N}{\phi(K_j^T)}}
\end{align}

Where $Q \in \mathbb{R}^{n \times d_k}$, $K \in \mathbb{R}^{n \times d_k}$, $V \in \mathbb{R}^{n \times d_v}$ denote the query, key, and value sequences,  respectively. $K_i \in \mathbb{R}^{d_k}$ and $V_i \in \mathbb{R}^{d_v}$ denotes the key and value at position $i$, $sim(.)$ denotes the similarity function which in this case represented as an exponent of the dot-product of the two vectors, and $\phi(.)$ is the feature map of the function $sim$. The kernel trick in Equation~\ref{eq:kernel} can be applied with the kernel $sim(.)$ and the feature map $\phi(.)$ as $sim(.)$ satisfy the properties of a valid kernel function which has to be symmetric and positive semi-definite.

\paragraph{Fixed/Random Pattern} Fixed/random pattern reduces the time complexity of the attention mechanism by manipulating the attention matrix into a sparse matrix to limit the field of view of the attention to fixed, predefined patterns such as local windows and block patterns with fixed strides which are easily parallelized with GPU and TPU devices. Longformer~\cite{beltagy2020longformer} applies a strided local window pattern. Blockwise transformer~\cite{qiu2020blockwise} and image transformer~\cite{armar2018imagetrf} apply a fix block pattern. ETC~\cite{ainslie2020etc} combines local window pattern with additional global attention on several tokens. Sinkhorn Transformer~\cite{tay2020sinkhorn} uses the block pattern approach to generate a local group. BigBird~\cite{zaheer2020bigbird} applies a combination of local window patterns, random patterns, and global patterns on several tokens. The illustration of the random, window, and global attention patterns are shown in Figure~\ref{fig:fixed-pattern}.

\begin{figure*}[ht]
    \centering
    \resizebox{0.8\textwidth}{!}{  
        \includegraphics{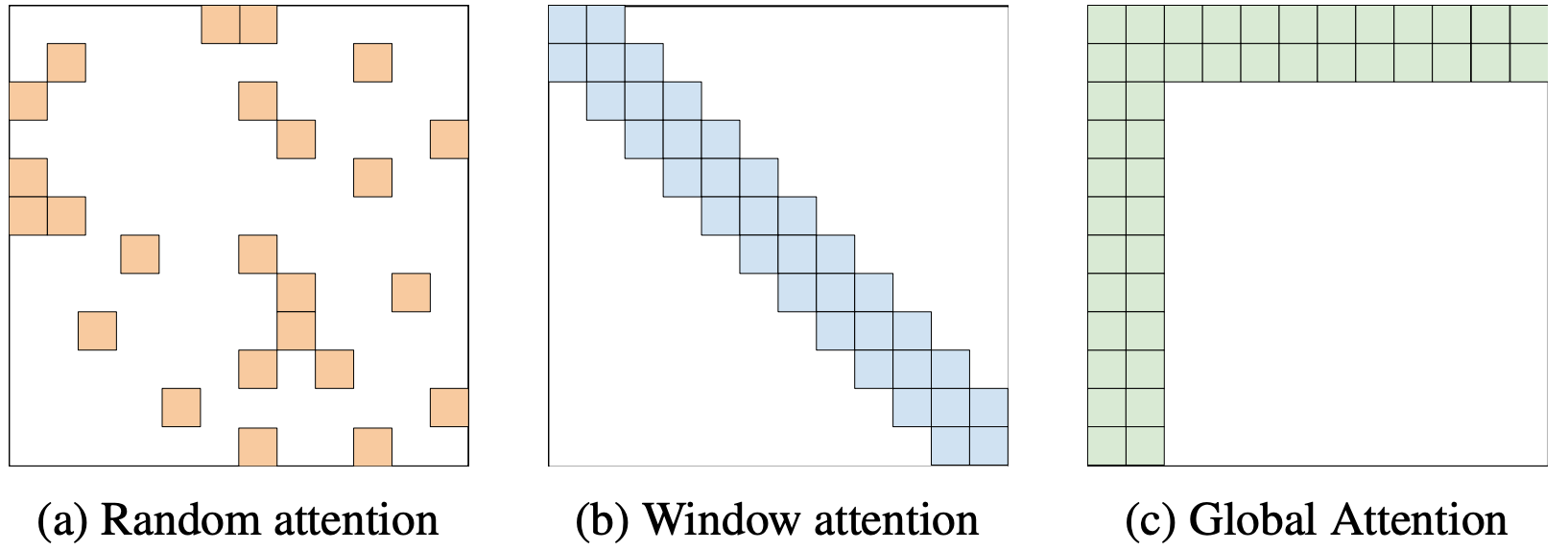}
    }
    \caption[Illustration of random, fixed, and global attention patterns]{Illustration of random (a), fixed (b), and global (c) attention patterns~\cite{zaheer2020bigbird}}
    \label{fig:fixed-pattern}
\end{figure*}

\paragraph{Learnable Pattern} Similar to fixed/random patterns, the learnable pattern tries to find the sparse attention matrix to reduce the time complexity. Sinkhorn Transformer~\cite{tay20sinkhorn} generates a learnable pattern from a generated local group to sort and filter out some of the groups to reduce the computation cost. Reformer~\cite{kitaev2020reformer} performs a hash-based similarity function called locality sensitivity hashing (LSH) to cluster tokens into several groups and compute the attention locally within each group. Similarly, Routing Transformer~\cite{roy2020routingtrf} cluster tokens into several groups by performing k-means clustering. The depiction of the learnable pattern approach is shown in Figure~\ref{fig:learnable-pattern}.

\begin{figure*}[ht]
    \centering
    \resizebox{1.0\textwidth}{!}{  
        \includegraphics{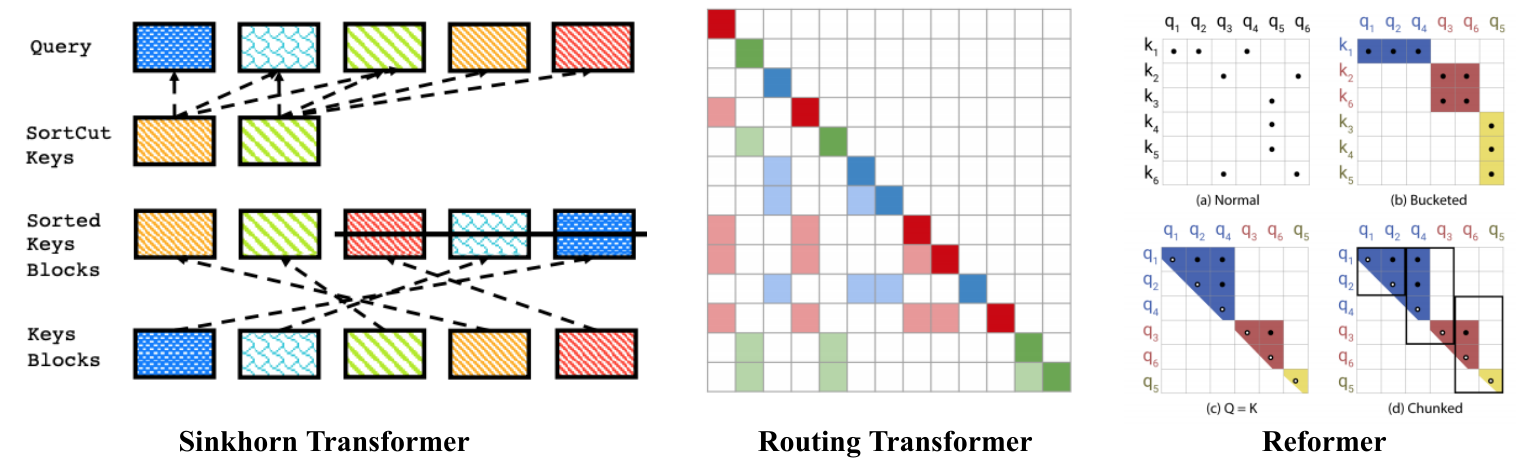}
    }
    \caption[Illustration of learnable pattern approaches]{Illustration of learnable pattern approaches. \textbf{(Left)} Sinkhorn Transformer~\cite{tay2020sinkhorn}, \textbf{(Center)} Routing Transformer~\cite{roy2020routingtrf}, and \textbf{(Right)} Reformer~\cite{kitaev2020reformer}}
    \label{fig:fixed-pattern}
\end{figure*}

\begin{figure*}[t!]
    \centering
    \resizebox{1.0\textwidth}{!}{  
        \includegraphics{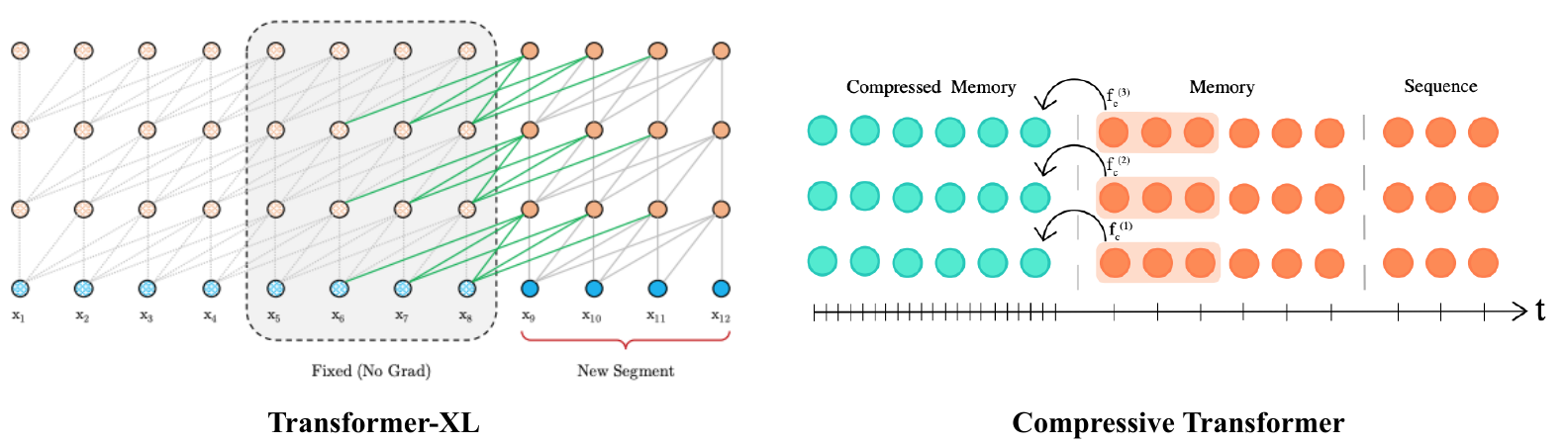}
    }
    \caption[Illustration of recurrence approaches]{Illustration of recurrence approaches. \textbf{(Left)} Transformer-XL~\cite{dai2019transformerxl} and \textbf{(Right)} Compressive Transfoemr~\cite{rae2020compressivetrs}}
    \label{fig:recurrence}
\end{figure*}

\paragraph{Recurrence} The recurrence method is in some sense similar to the combination of block pattern with local window pattern, as this method computes a segment-level recurrence to connect multiple blocks of sequence. Transformer-XL~\cite{dai2019transformerxl} provides a segment-level recurrence mechanism that allows the current segment to attend to the previous segment. Compressive Transformer~\cite{rae2020compressivetrs} extends Transformer-XL capability to attend to long-distance past segments by encoding the past segment into a fine-grained memory segment. The illustration of Transformer-XL and Compressive Transformer are shown in Figure~\ref{fig:recurrence}.

\paragraph{Memory} A memory approach leverages a side memory module that can access multiple tokens at once. A global token is common for the memory approach which can be seen as a form of memory that can access the entire sequence. The global token is incorporated in Set Tranformer~\cite{lee2019settrf}, ETC~\cite{ainslie2020etc}, Longformer~\cite{beltagy2020longformer}, and Bigbird~\cite{zaheer2020bigbird}. Compressive Transformer~\cite{rae2020compressivetrs} uses a form of memory module to encode the past segments. While the k-means centroids in the Routing Transformer can also be seen as a parameterized memory module that can access the entire sequence.

\paragraph{Low-Rank Approximation}
Low-rank approaches reduce the computational cost and memory usage by leveraging low-rank approximation on the parameters or activations of the model. Low-rank transformer (LRT)~\cite{winata2020lrt} reduces the computational cost of a transformer model by performing low-rank approximation on the weight of the linear layer inside the transformer model. Although LRT does not reduce the space and time complexity of the model, it improves the efficiency by significantly shrink down the model parameters. Linformer~\cite{wang2020linformer} reduces the space and time complexity of the attention mechanism from O($n^2$) to O($n$) with low-rank approximation to the $N \times N$ attention matrix. Linformer projects the sequence length of the key and value sequence to a lower-dimensional space. More detail on LRT and Linformer model is described in Chapter~\ref{ch:efficient_trf_low_rank}.

\section{Low-Rank Approximation and Matrix Factorization}

We denote a real-valued matrix $M \in \mathbb{R}^{m \times n}$ with rank $r, r \leq min(m,n)$. A low-rank approximation of $M$ is denoted as $\hat{M}=UV^T$ where $U \in \mathbb{R}^{m \times k}$, $V \in \mathbb{R}^{k \times n}$, and $k << min(m, n)$. Given the matrix $M$, such factorization can be estimated by solving a minimization problem where the cost function is the measure of fitness of between the matrix $M$ and the product of the low-rank matrices $\hat{M}$~\cite{golub1996matrix}. Distance function such as Frobenius norm $\left\lVert X \right\rVert _F = \sqrt{\sum_i \sum_j X_{ij}}$ is often use to solve the minimization problem. We can define the minimization problem as:

\begin{align}
\hat{M} &= \underset{\hat{M}}{\operatorname{argmin}} \left\lVert M-\hat{M} \right\rVert _F \\
(U,V) &= \underset{U,V}{\operatorname{argmin}} \left\lVert M-UV^T \right\rVert _F
\end{align}

The quadratic minimization problem can be solved through different methods such as singular value decomposition (SVD)~\cite{golub1970svd} or non-negative matrix factorization (NMF)~\cite{lee2001nmf}. Additionally, recent works in matrix factorization~\cite{kuchaiev2017factorization,lan2020albert,gao2018efficientsl,mirza2020efficientol} apply weight factorization to the model parameter before the model is trained and yield a comparable or even better result with fewer parameters and computational cost.

\subsection{Singular Value Decomposition}
SVD is an iterative approach to SVD decomposes a given a matrix $M \in \mathbb{R}^{m \times n}$ with rank $r$ into $M = U \Sigma V^T$ where $U  \in \mathbb{R}^{m \times m}$ is called as left-singular vectors, $V \in \mathbb{R}^{n \times n}$ is called as right-singular vectors, and $\Sigma \in \mathbb{R}^{m \times n}$ is a diagonal matrix consisting the singular values of the matrix $M$ on its first $r$ diagonal and 0 otherwise. In this form, SVD is useful for many applications in linear algebra including calculating pseudo inverse, solving least-square system, and determining the rank, range, and null space of the matrix.

\begin{figure*}[ht!]
    \centering
    \resizebox{0.75\textwidth}{!}{  
        \includegraphics{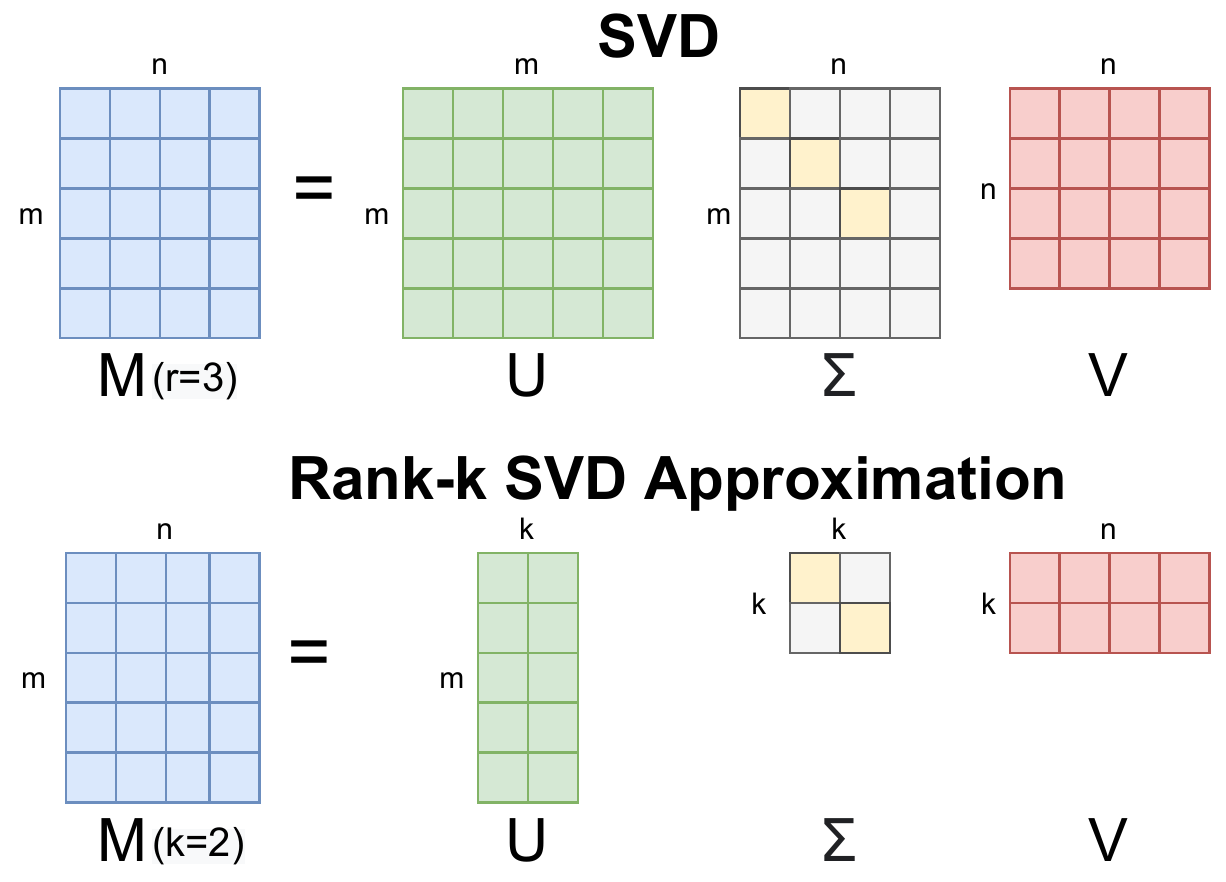}
    }
    \caption[Illustration of Singular Value Decomposition]{Illustration of Singular Value Decomposition}
    \label{fig:svd}
\end{figure*}

With the low-rank approximation, we can instead perform SVD with a pre-defined rank $k$ which denotes the number of $k$-highest singular values to consider and produce a much smaller $U$, $V$, and $\Sigma$ matrices. Based on Eckart-Young theorem, the best rank-$k$ SVD approximation to a matrix $M$ in Frobenius norm is attained by $B = u_i \sigma _i v_i$ and its error is $\left\lVert M-B \right\rVert _F = \sqrt{\sum_{i=k+1}^r \sigma _i^2}$ where $u_i$ is the column vector of matrix $U$, $v_i$ is column vector of matrix $V$, and $\sigma _i$ is diagonal entry of $\Sigma$. To get two matrices as defined before, we can simply apply the dot-product of $U\Sigma$ as the first matrix and use the $V$ as the second matrix. The rank-$k$ SVD approximation can also be used for denoising data as it removes the eigenvectors with smaller eigenvalues which can be considered as noise ~\cite{guo2016svddenoising}. The depiction of SVD and rank-$k$ SVD approximation is shown in Figure~\ref{fig:svd}.

\subsection{Non-negative Matrix Factorization}
NMF is another factorization method which factorize a matrix $V \in \mathbb{R}^{m \times n}$ into $V = WH$ where $W  \in \mathbb{R}^{m \times p}$, $H \in \mathbb{R}^{p \times n}$, and $p << min(m,n)$. Unlike SVD, NMF imposes non-negative constraint to all three matrices~\cite{wang2013nmf}. There are several solvers to find $W$ and $H$ for NMF and the most common method is called multiplicative update rule~\cite{lee2001nmf}. In multiplicative update rule, $W$ and $H$ are initialized with non-negative values and then iteratively performs element-wise update to $H$ and $W$ with the following equations: 

\begin{align}
H_{[i,j]} \longleftarrow H_{[i,j]} \frac{(W^TV)_{[i,j]}}{(WW^TH)_{[i,j]}} \\
W_{[i,j]} \longleftarrow W_{[i,j]} \frac{(VH^T)_{[i,j]}}{(WHH^T)_{[i,j]}}
\end{align}

\begin{figure*}[t!]
    \centering
    \resizebox{0.55\textwidth}{!}{  
        \includegraphics{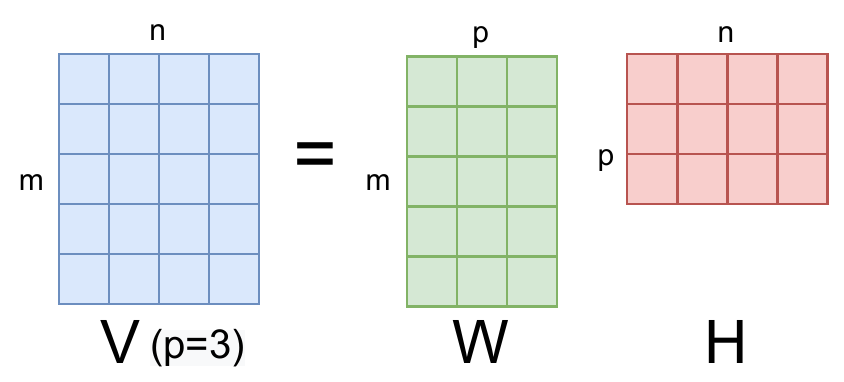}
    }
    \caption[Illustration of Non-negative Matrix Factorization]{Illustration of Non-negative Matrix Factorization}
    \label{fig:nmf}
\end{figure*}

The iterative update runs until it is stable or a certain pre-defined stopping criterion such as maximum iteration count is reached. A depiction of NMF factorization is shown in Figure~\ref{fig:nmf}.

\begin{figure*}[ht!]
    \centering
    \resizebox{0.95\textwidth}{!}{  
        \includegraphics{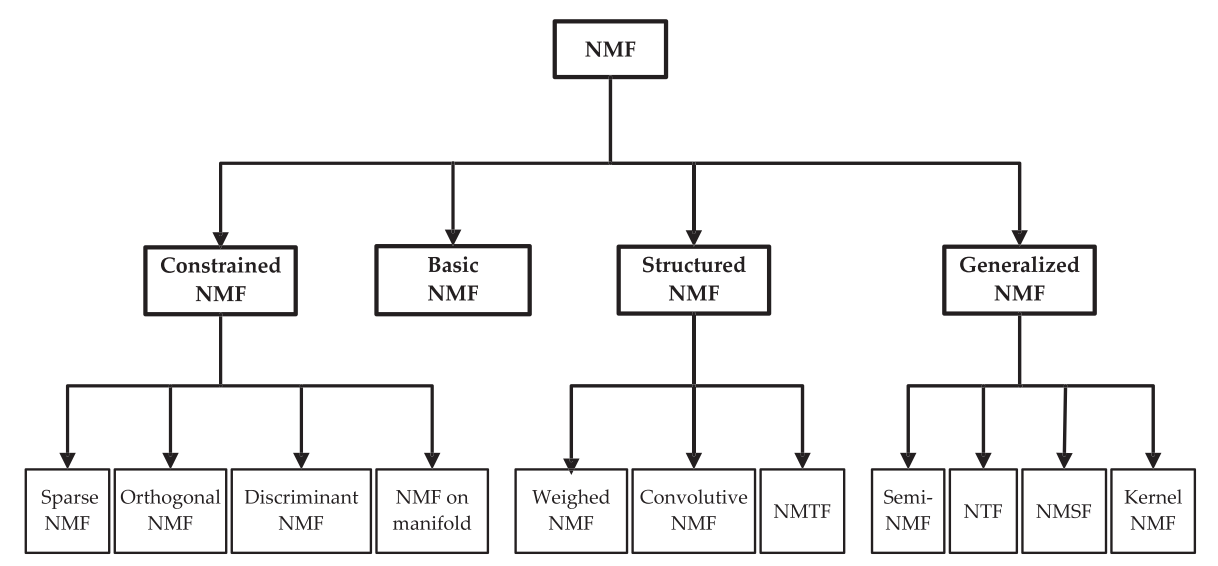}
    }
    \caption[Categorization of Non-negative Matrix Factorization Algorithms]{Categorization of Non-negative Matrix Factorization Algorithms~\cite{wang2013nmf}}
    \label{fig:nmf_categorization}
\end{figure*}

NMF algorithms can be divided into 4 categories: Basic NMF (BNMF), Constrained NMF (CNMF), Structured NMF (SNMF), and Generalized NMF (GNMF). CNMF imposes some additional constraints as regularization to the BNMF. SNMF enforces other characteristics or structures in the solution to the NMF learning problem of BNMF. While, GNMF generalizes BNMF by
breaching the intrinsic non-negativity constraint to some extent, or changes the data types, or alters the factorization pattern. Each NMF category is further divided into several subclasses as shown in Figure~\ref{fig:nmf_categorization}. 

\subsection{Pre-Training Low-Rank Matrix Factorization}

Both SVD and NMF require the information of the matrix $M$ to be known for the methods to take place. This means that both SVD and NMF can only be applied to factorize the weight matrix of the model once the training process is completed. This limits the applicability of the low-rank method to only increase the efficiency of the inference phase. Other works in low-rank matrix factorization explore the possibility to perform low-rank factorization to the weight matrix of the model before training the model. Specifically, given a transformation layer of the form $y = f(xW)$, where $x$ is an $d_{in}$-dimensional input, $y$ is an $d_out$-dimensional output, and $W \in \mathbb{R}^{d_{in} \times d_{out}}$ is the weight matrix of the function $f$, we can decrease the number of the parameters of function $f$ by expressing $W$ as the product of two smaller matrices $W = UV$, where $U \in \mathbb{R}^{d_{in} \times k}$, $V \in \mathbb{R}^{k \times d_{out}}$, and $k << min(d_{in}, d_{out})$. By choosing a small $k$, we can achieve a substantial reduction in the number of parameters, memory usage, and computation cost. We call this method as in-training low-rank factorization method.

With the in-training low-rank factorization, we can simply replace $W$ with two smaller matrices $U$ and $V$ and compute derivatives for $U$ and $V$ instead of $W$ during the training. This approach has been explored in the previous work ~\cite{denil2013predictparam} with a multi-layer perceptron network and their experimental results suggest that this approach is very unstable and lead to a higher error rate. Nevertheless, more recent works~\cite{kuchaiev2017factorization,winata2020lrt,lan2020albert,gao2018efficientsl, mirza2020efficientol} apply similar methods to different model architecture to reduce the model parameters and reduce its computational cost. These works suggest that training more advance neural network model architectures, such as CNN and RNN, with randomly initialized low-rank factorized weight matrix can result in a factorized model that works as good as the non-factorized counterpart while gaining a significant reduction in terms of the number of parameters, computational cost, and memory usage.

\newpage

%% file: chapter/sec-2.5-low-rank-factorized-transformer.tex
\chapter{Greenformers: Efficient Transformer Model via Low-Rank Approximation}
\label{ch:efficient_trf_low_rank}

In this chapter, we explore two Greenformers models that apply low-rank approximation to the transformer model: 1) low-rank approximation on the linear layer of the transformer model, and 2) low-rank approximation on the self-attention mechanism of the transformer model. We called the first variant as Low-Rank Transformer (LRT)~\cite{winata2020lrt}, while the second variant as Linformer~\cite{wang2020linformer}. Both variants can help to improve the efficiency of the transformer model significantly and we conduct a study on these two variants to get a better understanding of the drawbacks of each model and find the best condition to utilize each model.

We conduct a thorough analysis to compare the efficiency of the original transformer model, LRT, and Linformer. We compare the three variants in terms of memory usage and computational cost on different characteristics of the input sequence. We further analyze the effect of low-rank factor $r$ on the efficiency improvement of LRT and Linformer compared to the baseline transformer model. Lastly, we conclude our analysis by presenting the benefits and limitations of each low-rank transformer model and provide a recommendation of the best setting to use each model variant.

\section{Methodology}

\subsection{Low-Rank Transformer: Efficient Transformer with Factorized Linear Projection}

\begin{figure}[!t]
  \centering
  \includegraphics[width=.6\linewidth]{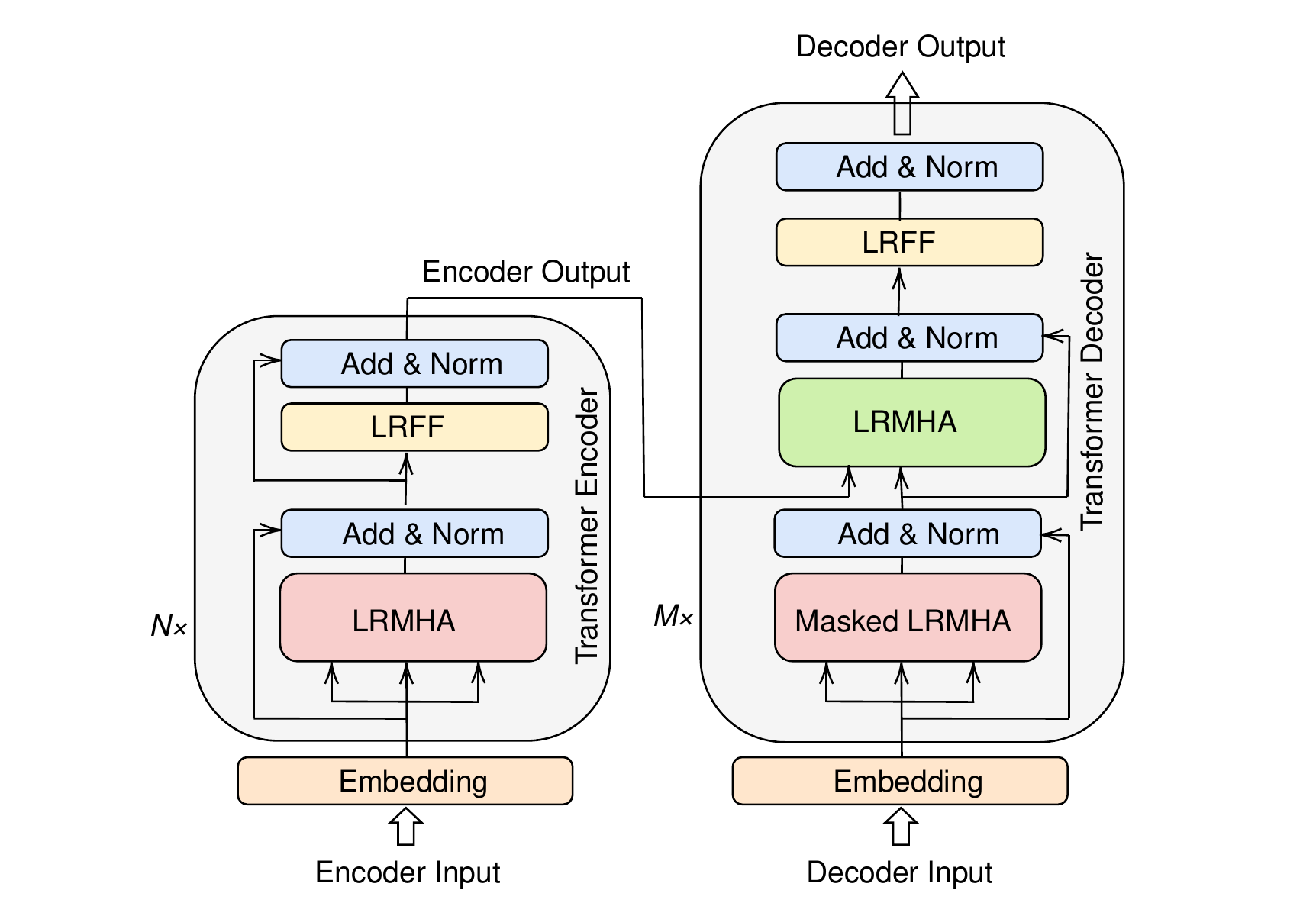}  
  \caption[Low-Rank Transformer Architecture]{Low-Rank Transformer Architecture. Low-Rank Transformer consists of $N$ layers of low-rank transformer encoder and $M$ layers of low-rank transformer decoder.~\cite{winata2020lrt}}
  \label{fig:lrt}
\end{figure}

\begin{figure*}[!t]
\begin{minipage}{.20\textwidth}
  \centering
  \includegraphics[width=.80\linewidth]{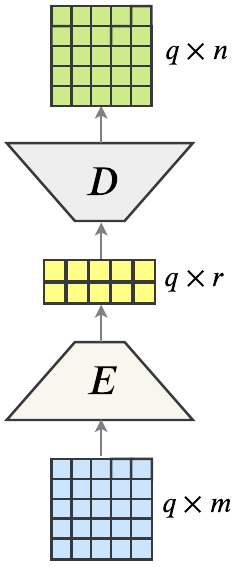}
\end{minipage}
\begin{minipage}{.45\textwidth}
  \centering
  \includegraphics[width=1.00\linewidth]{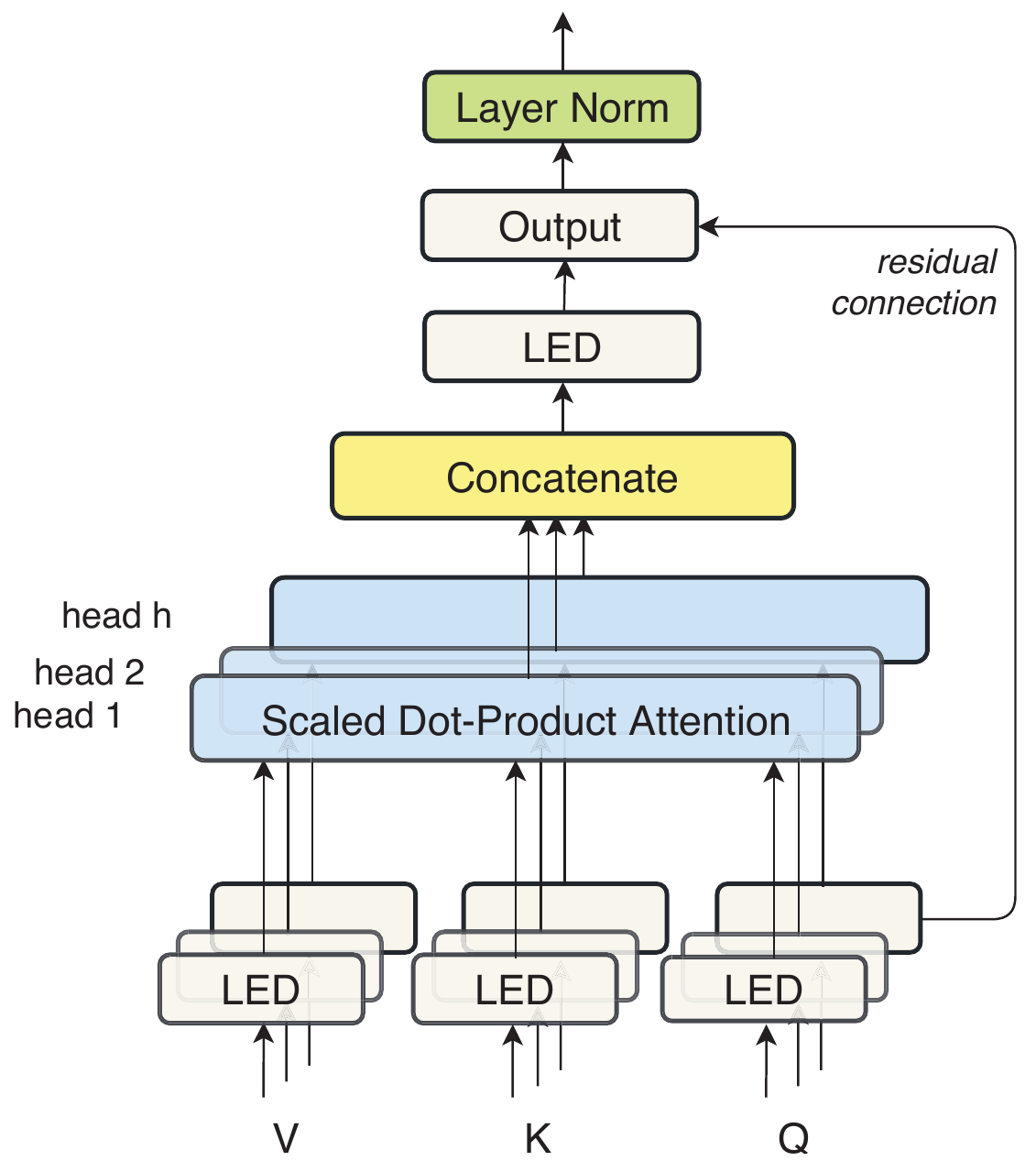}  
\end{minipage}
\hspace{20pt}
\begin{minipage}{.29\textwidth}
  \centering
  \includegraphics[width=.85\linewidth]{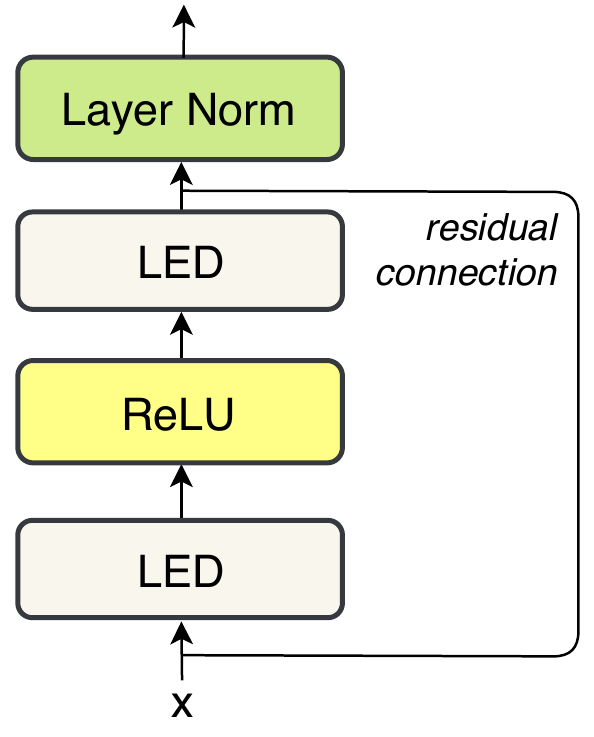} 
\end{minipage}
\caption[Low-Rank Transformer Unit]{Low-Rank Transformer Unit. \textbf{Left:} Linear Encoder-Decoder (LED), \textbf{Center:} Low-Rank Multi-head Attention (LRMHA), and \textbf{Right:} Low-Rank Feed-forward (LRFF).~\cite{winata2020lrt}}
\label{fig:lr-units}
\end{figure*}

To reduce the computational cost of the transformer model~\cite{vaswani2017transformer}, we extend the idea of the in-training low-rank factorization method~\cite{kuchaiev2017factorization,gao2018efficientsl,denil2013predictparam} and introduce the low-rank variant of the transformer model
called LRT. Specifically, we apply low-rank matrix factorization to the linear layers in a transformer model. We call the factorized linear layer as linear encoder-decoder (LED) unit. The matrix factorization is applied by approximating the matrix $\mathbf{W} \in \mathbb{R}^{m \times n}$ in the linear layer using two smaller matrices, $\mathbf{E}  \in \mathbb{R}^{m \times r}$ and $\mathbf{D}  \in \mathbb{R}^{r \times n}$ such that $\mathbf{W} \approx \mathbf{E} \times \mathbf{D}$. The matrix $\mathbf{W}$ requires $mn$ parameters and $mn$ FLOPS, while  $\mathbf{E}$ and $\mathbf{D}$ require $rm + rn=r(m+n)$ parameters and $rm + rn=r(m+n)$ flops. If we choose the rank to be very low $r<<m,n$, the number of parameters and FLOPS in $\mathbf{E}$ and $\mathbf{D}$ are much smaller compared to $\mathbf{W}$. The design of our LED unit is shown in Figure~\ref{fig:lr-units} (Left).

As shown in Figure~\ref{fig:lrt}, our proposed LRT model consists of $N$ layers of the LRT encoder to refine the low-level input features into higher-level features, and $M$ layers of the LRT decoder to generate the output sequence. Each LRT encoder layer consists of low-rank multi-head attention (LRMHA) layer followed by a low-rank feed-forward (LRFF) layer. While each LRT decoder layer consists of a masked LRMHA layer that uses causal masking to prevent attention to the future time-step, followed by an LRMHA layer to perform a cross-attention mechanism with the output from the LRT encoder, and followed by an LRFF layer.

\subsubsection{Low-Rank Multi-Head Attention}

To reduce the computational in the multi-head attention layer, we introduce LRMHA layer. We apply low-rank approximation to LRMHA by factorizing the query projection layers $W_i^Q \in \mathbb{R}^{d_{model} \times d_k}$, key projection layers $W_i^K \in \mathbb{R}^{d_{model} \times d_k}$, value projection layers $W_i^V \in \mathbb{R}^{d_{model} \times d_v}$, and the output projection layer $W^O \in \mathbb{R}^{d_{v} \times d_{model}}$. As shown in Figure~\ref{fig:lr-units} (Center), our LRMHA layer accepts a query sequence $Q$, a key sequence $K$, and a value sequence $V$. Similar to the original transformer model, we add a residual connection from the input query sequence $Q$ with the result from the output projection. Specifically, we formulate our LRMHA layer as follow:

\begin{align}
&\textnormal{Attention}(Q,K,V)=\textnormal{Softmax}(\frac{QK^T}{\sqrt{d_k}}V),\\
&hd_i = \textnormal{Attention}(QE_i^Q D_i^Q, KE_i^K D_i^K, VE_i^V D_i^V),\\
&f(Q,K,V) = \textnormal{LayerNorm}(\textnormal{Concat}(h_1,\cdots,h_H)E^O D^O+Q),
\end{align}

where $f$ denotes the LRMHA function, $h_i$ denotes the head of $i$, $H$ denotes the number of heads, $\textnormal{LayerNorm}$ denotes layer normalization function, $\textnormal{Concat}$ denotes concatenation operation between multiple heads, and $E_i^Q \in \mathbb{R}^{d_{model} \times d_r}$; $D_i^Q \in \mathbb{R}^{d_{r} \times d_k}$; $E_i^K \in \mathbb{R}^{d_{model} \times d_r}$; $D_i^K \in \mathbb{R}^{d_{r} \times d_k}$; $E_i^V \in \mathbb{R}^{d_{model} \times d_r}$; $D_i^V \in \mathbb{R}^{d_{r} \times d_v}$; $E^O \in \mathbb{R}^{d_{v} \times d_r}$; $D^O \in \mathbb{R}^{d_{r} \times d_{model}}$ are the low-rank encoder-decoder matrices for query projection $W_i^Q$, key projection $W_i^K$, value projection $W_i^V$ and output projection $W^O$. $d_{model}$, $d_{r}$, $d_{k}$, and $d_{v}$ denote dimensions of hidden size, rank, key, and value, respectively.

\subsubsection{Low-Rank Feed-Forward}

As shown in Figure~\ref{fig:lr-units} (Right), LRFF consists of two LED units and a ReLU activation function is applied in between the two LED units. Similar to the original transformer model, A residual connection is added in LRFF by connecting the input sequence $x$ with the output from the second LED unit to alleviate the gradient vanishing issue. Specifically, we define our LRFF unit as follow:

\begin{equation}
    g(x) = \textnormal{LayerNorm}(  \textnormal{max}(0, x E_1 D_1)E_2 D_2 + x),
\end{equation}

where $g$ denotes the low-rank feed-forward (LRFF) function and  $E_1 \in \mathbb{R}^{d_{model} \times d_r}$; $D_1 \in \mathbb{R}^{d_{r} \times d_{inner}}$; $E_2 \in \mathbb{R}^{d_{inner} \times d_r}$; $D_2 \in \mathbb{R}^{d_{r} \times d_{model}}$ are the low-rank encoder-decoder matrices for the first and the second LED units respectively.

\subsection{Linformer: Efficient Transformer with Factorized Attention Mechanism}
\label{sec-modeling}

\begin{figure}[h!]
\centering
\resizebox{1.0\linewidth}{!}{
\tikzset{every picture/.style={line width=0.75pt}} 

\begin{tikzpicture}[x=0.75pt,y=0.75pt,yscale=-1,xscale=1]

\draw  [fill={rgb, 255:red, 245; green, 245; blue, 245 }  ,fill opacity=1 ] (46,229.6) .. controls (46,200) and (70,176) .. (99.6,176) -- (260.4,176) .. controls (290,176) and (314,200) .. (314,229.6) -- (314,406.92) .. controls (314,436.52) and (290,460.52) .. (260.4,460.52) -- (99.6,460.52) .. controls (70,460.52) and (46,436.52) .. (46,406.92) -- cycle ;
\draw    (63.57,531.01) -- (308.75,531.01) ;
\draw    (63.57,546.83) -- (308.75,546.83) ;
\draw    (72.02,530.7) -- (72.02,546.21) ;
\draw    (84.68,530.7) -- (84.68,546.21) ;
\draw    (97.35,530.7) -- (97.35,546.21) ;
\draw    (110.02,530.7) -- (110.02,546.21) ;
\draw    (122.68,530.7) -- (122.68,546.21) ;
\draw    (135.35,530.73) -- (135.35,546.24) ;
\draw    (148.02,530.73) -- (148.02,546.24) ;
\draw    (160.69,530.7) -- (160.69,546.21) ;
\draw    (173.35,530.7) -- (173.35,546.21) ;
\draw    (186.02,530.7) -- (186.02,546.21) ;
\draw    (198.69,530.7) -- (198.69,546.21) ;
\draw    (211.36,530.7) -- (211.36,546.21) ;
\draw    (224.02,530.73) -- (224.02,546.24) ;
\draw    (236.69,530.73) -- (236.69,546.24) ;
\draw    (249.36,531.32) -- (249.36,546.83) ;
\draw    (262.02,531.32) -- (262.02,546.83) ;
\draw    (274.69,531.32) -- (274.69,546.83) ;
\draw    (287.36,531.35) -- (287.36,546.86) ;
\draw    (300.03,531.35) -- (300.03,546.86) ;
\draw  [fill={rgb, 255:red, 255; green, 230; blue, 204 }  ,fill opacity=1 ] (71.2,477) .. controls (71.2,473.69) and (73.89,471) .. (77.2,471) -- (293.2,471) .. controls (296.51,471) and (299.2,473.69) .. (299.2,477) -- (299.2,495) .. controls (299.2,498.31) and (296.51,501) .. (293.2,501) -- (77.2,501) .. controls (73.89,501) and (71.2,498.31) .. (71.2,495) -- cycle ;
\draw   (196.4,512.58) -- (191.05,512.58) -- (191.05,526.65) -- (180.36,526.65) -- (180.36,512.58) -- (175.02,512.58) -- (185.71,503.2) -- cycle ;
\draw    (186.35,470.73) -- (186.01,412.6) ;
\draw [shift={(186,410.6)}, rotate = 449.66] [color={rgb, 255:red, 0; green, 0; blue, 0 }  ][line width=0.75]    (10.93,-3.29) .. controls (6.95,-1.4) and (3.31,-0.3) .. (0,0) .. controls (3.31,0.3) and (6.95,1.4) .. (10.93,3.29)   ;
\draw    (235,436) -- (235,412) ;
\draw [shift={(235,410)}, rotate = 450] [color={rgb, 255:red, 0; green, 0; blue, 0 }  ][line width=0.75]    (10.93,-3.29) .. controls (6.95,-1.4) and (3.31,-0.3) .. (0,0) .. controls (3.31,0.3) and (6.95,1.4) .. (10.93,3.29)   ;
\draw    (135,435.5) -- (135,412.5) ;
\draw [shift={(135,410.5)}, rotate = 450] [color={rgb, 255:red, 0; green, 0; blue, 0 }  ][line width=0.75]    (10.93,-3.29) .. controls (6.95,-1.4) and (3.31,-0.3) .. (0,0) .. controls (3.31,0.3) and (6.95,1.4) .. (10.93,3.29)   ;
\draw  [fill={rgb, 255:red, 248; green, 206; blue, 204 }  ,fill opacity=1 ] (100,365.23) .. controls (100,359.05) and (105.01,354.04) .. (111.19,354.04) -- (270.21,354.04) .. controls (276.39,354.04) and (281.4,359.05) .. (281.4,365.23) -- (281.4,398.81) .. controls (281.4,404.99) and (276.39,410) .. (270.21,410) -- (111.19,410) .. controls (105.01,410) and (100,404.99) .. (100,398.81) -- cycle ;
\draw  [fill={rgb, 255:red, 218; green, 232; blue, 252 }  ,fill opacity=1 ] (101,312) .. controls (101,308.69) and (103.69,306) .. (107,306) -- (276.4,306) .. controls (279.71,306) and (282.4,308.69) .. (282.4,312) -- (282.4,330) .. controls (282.4,333.31) and (279.71,336) .. (276.4,336) -- (107,336) .. controls (103.69,336) and (101,333.31) .. (101,330) -- cycle ;
\draw  [fill={rgb, 255:red, 218; green, 232; blue, 252 }  ,fill opacity=1 ] (101,204) .. controls (101,200.69) and (103.69,198) .. (107,198) -- (276.4,198) .. controls (279.71,198) and (282.4,200.69) .. (282.4,204) -- (282.4,222) .. controls (282.4,225.31) and (279.71,228) .. (276.4,228) -- (107,228) .. controls (103.69,228) and (101,225.31) .. (101,222) -- cycle ;
\draw  [fill={rgb, 255:red, 255; green, 242; blue, 204 }  ,fill opacity=1 ] (101,252) .. controls (101,248.69) and (103.69,246) .. (107,246) -- (276.4,246) .. controls (279.71,246) and (282.4,248.69) .. (282.4,252) -- (282.4,270) .. controls (282.4,273.31) and (279.71,276) .. (276.4,276) -- (107,276) .. controls (103.69,276) and (101,273.31) .. (101,270) -- cycle ;
\draw    (135,435.5) -- (235.67,435.83) ;
\draw    (185,335.67) -- (185.06,343.34) -- (185,354.33) ;
\draw    (186,306) -- (186,278) ;
\draw [shift={(186,276)}, rotate = 450] [color={rgb, 255:red, 0; green, 0; blue, 0 }  ][line width=0.75]    (10.93,-3.29) .. controls (6.95,-1.4) and (3.31,-0.3) .. (0,0) .. controls (3.31,0.3) and (6.95,1.4) .. (10.93,3.29)   ;
\draw    (82,321.5) -- (99,321.5) ;
\draw [shift={(101,321.5)}, rotate = 180] [color={rgb, 255:red, 0; green, 0; blue, 0 }  ][line width=0.75]    (10.93,-3.29) .. controls (6.95,-1.4) and (3.31,-0.3) .. (0,0) .. controls (3.31,0.3) and (6.95,1.4) .. (10.93,3.29)   ;
\draw    (82,321.5) -- (82,449.5) ;
\draw    (82,448.87) -- (186.4,449) ;
\draw    (185,227.67) -- (185.06,235.34) -- (185,246.33) ;
\draw    (82,213.5) -- (99,213.5) ;
\draw [shift={(101,213.5)}, rotate = 180] [color={rgb, 255:red, 0; green, 0; blue, 0 }  ][line width=0.75]    (10.93,-3.29) .. controls (6.95,-1.4) and (3.31,-0.3) .. (0,0) .. controls (3.31,0.3) and (6.95,1.4) .. (10.93,3.29)   ;
\draw    (82,213.5) -- (82,295.5) ;
\draw    (82,294.87) -- (186.4,295) ;
\draw    (186,198) -- (186,158) ;
\draw [shift={(186,156)}, rotate = 450] [color={rgb, 255:red, 0; green, 0; blue, 0 }  ][line width=0.75]    (10.93,-3.29) .. controls (6.95,-1.4) and (3.31,-0.3) .. (0,0) .. controls (3.31,0.3) and (6.95,1.4) .. (10.93,3.29)   ;
\draw  [fill={rgb, 255:red, 213; green, 232; blue, 212 }  ,fill opacity=1 ] (71.2,132) .. controls (71.2,128.69) and (73.89,126) .. (77.2,126) -- (293.2,126) .. controls (296.51,126) and (299.2,128.69) .. (299.2,132) -- (299.2,150) .. controls (299.2,153.31) and (296.51,156) .. (293.2,156) -- (77.2,156) .. controls (73.89,156) and (71.2,153.31) .. (71.2,150) -- cycle ;
\draw   (196.4,105.58) -- (191.05,105.58) -- (191.05,119.65) -- (180.36,119.65) -- (180.36,105.58) -- (175.02,105.58) -- (185.71,96.2) -- cycle ;
\draw  [dash pattern={on 4.5pt off 4.5pt}]  (360.43,81.86) -- (361,568) ;
\draw  [dash pattern={on 4.5pt off 4.5pt}]  (723.43,79.86) -- (724,568) ;
\draw    (550.35,510.73) -- (550.02,474.6) ;
\draw [shift={(550,472.6)}, rotate = 449.47] [color={rgb, 255:red, 0; green, 0; blue, 0 }  ][line width=0.75]    (10.93,-3.29) .. controls (6.95,-1.4) and (3.31,-0.3) .. (0,0) .. controls (3.31,0.3) and (6.95,1.4) .. (10.93,3.29)   ;
\draw    (430,490.5) -- (430,434.5) ;
\draw [shift={(430,432.5)}, rotate = 450] [color={rgb, 255:red, 0; green, 0; blue, 0 }  ][line width=0.75]    (10.93,-3.29) .. controls (6.95,-1.4) and (3.31,-0.3) .. (0,0) .. controls (3.31,0.3) and (6.95,1.4) .. (10.93,3.29)   ;
\draw    (430,490.5) -- (649.67,490.83) ;
\draw  [fill={rgb, 255:red, 218; green, 232; blue, 252 }  ,fill opacity=1 ] (374,404.94) .. controls (374,401.11) and (377.11,398) .. (380.94,398) -- (478.32,398) .. controls (482.15,398) and (485.26,401.11) .. (485.26,404.94) -- (485.26,425.76) .. controls (485.26,429.59) and (482.15,432.7) .. (478.32,432.7) -- (380.94,432.7) .. controls (377.11,432.7) and (374,429.59) .. (374,425.76) -- cycle ;
\draw  [fill={rgb, 255:red, 218; green, 232; blue, 252 }  ,fill opacity=1 ] (596,394.54) .. controls (596,389.27) and (600.27,385) .. (605.54,385) -- (697.72,385) .. controls (702.99,385) and (707.26,389.27) .. (707.26,394.54) -- (707.26,423.16) .. controls (707.26,428.42) and (702.99,432.7) .. (697.72,432.7) -- (605.54,432.7) .. controls (600.27,432.7) and (596,428.42) .. (596,423.16) -- cycle ;
\draw  [fill={rgb, 255:red, 255; green, 230; blue, 204 }  ,fill opacity=1 ] (495,521.54) .. controls (495,516.27) and (499.27,512) .. (504.54,512) -- (596.72,512) .. controls (601.99,512) and (606.26,516.27) .. (606.26,521.54) -- (606.26,550.16) .. controls (606.26,555.42) and (601.99,559.7) .. (596.72,559.7) -- (504.54,559.7) .. controls (499.27,559.7) and (495,555.42) .. (495,550.16) -- cycle ;
\draw    (650,490.5) -- (650,434.5) ;
\draw [shift={(650,432.5)}, rotate = 450] [color={rgb, 255:red, 0; green, 0; blue, 0 }  ][line width=0.75]    (10.93,-3.29) .. controls (6.95,-1.4) and (3.31,-0.3) .. (0,0) .. controls (3.31,0.3) and (6.95,1.4) .. (10.93,3.29)   ;
\draw  [fill={rgb, 255:red, 218; green, 232; blue, 252 }  ,fill opacity=1 ] (560.04,361.72) .. controls (563.87,361.72) and (566.98,364.82) .. (566.98,368.66) -- (566.98,466.04) .. controls (566.98,469.87) and (563.87,472.98) .. (560.04,472.98) -- (539.22,472.98) .. controls (535.39,472.98) and (532.28,469.87) .. (532.28,466.04) -- (532.28,368.66) .. controls (532.28,364.82) and (535.39,361.72) .. (539.22,361.72) -- cycle ;
\draw    (549.57,339.43) -- (549.5,361.5) ;
\draw    (429.57,339.43) -- (429.5,397.5) ;
\draw  [fill={rgb, 255:red, 255; green, 242; blue, 204 }  ,fill opacity=1 ] (476.5,340.6) .. controls (476.5,333.97) and (481.65,328.6) .. (488,328.6) .. controls (494.35,328.6) and (499.5,333.97) .. (499.5,340.6) .. controls (499.5,347.23) and (494.35,352.6) .. (488,352.6) .. controls (481.65,352.6) and (476.5,347.23) .. (476.5,340.6) -- cycle (465,340.6) -- (476.5,340.6) (511,340.6) -- (499.5,340.6) ;
\draw    (429.5,340.6) -- (474.5,340.6) ;
\draw [shift={(476.5,340.6)}, rotate = 180] [color={rgb, 255:red, 0; green, 0; blue, 0 }  ][line width=0.75]    (10.93,-3.29) .. controls (6.95,-1.4) and (3.31,-0.3) .. (0,0) .. controls (3.31,0.3) and (6.95,1.4) .. (10.93,3.29)   ;
\draw    (549.5,340.6) -- (501.5,340.6) ;
\draw [shift={(499.5,340.6)}, rotate = 360] [color={rgb, 255:red, 0; green, 0; blue, 0 }  ][line width=0.75]    (10.93,-3.29) .. controls (6.95,-1.4) and (3.31,-0.3) .. (0,0) .. controls (3.31,0.3) and (6.95,1.4) .. (10.93,3.29)   ;
\draw    (488,328) -- (488,311) ;
\draw [shift={(488,309)}, rotate = 450] [color={rgb, 255:red, 0; green, 0; blue, 0 }  ][line width=0.75]    (10.93,-3.29) .. controls (6.95,-1.4) and (3.31,-0.3) .. (0,0) .. controls (3.31,0.3) and (6.95,1.4) .. (10.93,3.29)   ;
\draw  [fill={rgb, 255:red, 213; green, 232; blue, 212 }  ,fill opacity=1 ] (433,231.01) .. controls (433,220.14) and (441.81,211.33) .. (452.67,211.33) -- (524.59,211.33) .. controls (535.45,211.33) and (544.26,220.14) .. (544.26,231.01) -- (544.26,290.02) .. controls (544.26,300.89) and (535.45,309.7) .. (524.59,309.7) -- (452.67,309.7) .. controls (441.81,309.7) and (433,300.89) .. (433,290.02) -- cycle ;
\draw    (649.57,196.43) -- (649.5,385.5) ;
\draw  [fill={rgb, 255:red, 255; green, 242; blue, 204 }  ,fill opacity=1 ] (546.5,197.6) .. controls (546.5,190.97) and (551.65,185.6) .. (558,185.6) .. controls (564.35,185.6) and (569.5,190.97) .. (569.5,197.6) .. controls (569.5,204.23) and (564.35,209.6) .. (558,209.6) .. controls (551.65,209.6) and (546.5,204.23) .. (546.5,197.6) -- cycle (535,197.6) -- (546.5,197.6) (581,197.6) -- (569.5,197.6) ;
\draw    (488.5,197.6) -- (544.5,197.6) ;
\draw [shift={(546.5,197.6)}, rotate = 180] [color={rgb, 255:red, 0; green, 0; blue, 0 }  ][line width=0.75]    (10.93,-3.29) .. controls (6.95,-1.4) and (3.31,-0.3) .. (0,0) .. controls (3.31,0.3) and (6.95,1.4) .. (10.93,3.29)   ;
\draw    (649.5,197.6) -- (571.5,197.6) ;
\draw [shift={(569.5,197.6)}, rotate = 360] [color={rgb, 255:red, 0; green, 0; blue, 0 }  ][line width=0.75]    (10.93,-3.29) .. controls (6.95,-1.4) and (3.31,-0.3) .. (0,0) .. controls (3.31,0.3) and (6.95,1.4) .. (10.93,3.29)   ;
\draw    (558,185) -- (558,167) ;
\draw [shift={(558,165)}, rotate = 450] [color={rgb, 255:red, 0; green, 0; blue, 0 }  ][line width=0.75]    (10.93,-3.29) .. controls (6.95,-1.4) and (3.31,-0.3) .. (0,0) .. controls (3.31,0.3) and (6.95,1.4) .. (10.93,3.29)   ;
\draw  [fill={rgb, 255:red, 213; green, 232; blue, 212 }  ,fill opacity=1 ] (503,126.54) .. controls (503,121.27) and (507.27,117) .. (512.54,117) -- (604.72,117) .. controls (609.99,117) and (614.26,121.27) .. (614.26,126.54) -- (614.26,155.16) .. controls (614.26,160.42) and (609.99,164.7) .. (604.72,164.7) -- (512.54,164.7) .. controls (507.27,164.7) and (503,160.42) .. (503,155.16) -- cycle ;
\draw    (488.5,197.6) -- (488.43,211.67) ;
\draw    (920.35,510.73) -- (920.02,474.6) ;
\draw [shift={(920,472.6)}, rotate = 449.47] [color={rgb, 255:red, 0; green, 0; blue, 0 }  ][line width=0.75]    (10.93,-3.29) .. controls (6.95,-1.4) and (3.31,-0.3) .. (0,0) .. controls (3.31,0.3) and (6.95,1.4) .. (10.93,3.29)   ;
\draw    (800,490.5) -- (800,434.5) ;
\draw [shift={(800,432.5)}, rotate = 450] [color={rgb, 255:red, 0; green, 0; blue, 0 }  ][line width=0.75]    (10.93,-3.29) .. controls (6.95,-1.4) and (3.31,-0.3) .. (0,0) .. controls (3.31,0.3) and (6.95,1.4) .. (10.93,3.29)   ;
\draw    (800,490.5) -- (1019.67,490.83) ;
\draw  [fill={rgb, 255:red, 218; green, 232; blue, 252 }  ,fill opacity=1 ] (744,404.94) .. controls (744,401.11) and (747.11,398) .. (750.94,398) -- (848.32,398) .. controls (852.15,398) and (855.26,401.11) .. (855.26,404.94) -- (855.26,425.76) .. controls (855.26,429.59) and (852.15,432.7) .. (848.32,432.7) -- (750.94,432.7) .. controls (747.11,432.7) and (744,429.59) .. (744,425.76) -- cycle ;
\draw  [fill={rgb, 255:red, 218; green, 232; blue, 252 }  ,fill opacity=1 ] (966,394.54) .. controls (966,389.27) and (970.27,385) .. (975.54,385) -- (1067.72,385) .. controls (1072.99,385) and (1077.26,389.27) .. (1077.26,394.54) -- (1077.26,423.16) .. controls (1077.26,428.42) and (1072.99,432.7) .. (1067.72,432.7) -- (975.54,432.7) .. controls (970.27,432.7) and (966,428.42) .. (966,423.16) -- cycle ;
\draw  [fill={rgb, 255:red, 255; green, 230; blue, 204 }  ,fill opacity=1 ] (865,521.54) .. controls (865,516.27) and (869.27,512) .. (874.54,512) -- (966.72,512) .. controls (971.99,512) and (976.26,516.27) .. (976.26,521.54) -- (976.26,550.16) .. controls (976.26,555.42) and (971.99,559.7) .. (966.72,559.7) -- (874.54,559.7) .. controls (869.27,559.7) and (865,555.42) .. (865,550.16) -- cycle ;
\draw    (1020,490.5) -- (1020,434.5) ;
\draw [shift={(1020,432.5)}, rotate = 450] [color={rgb, 255:red, 0; green, 0; blue, 0 }  ][line width=0.75]    (10.93,-3.29) .. controls (6.95,-1.4) and (3.31,-0.3) .. (0,0) .. controls (3.31,0.3) and (6.95,1.4) .. (10.93,3.29)   ;
\draw  [fill={rgb, 255:red, 218; green, 232; blue, 252 }  ,fill opacity=1 ] (930.04,361.72) .. controls (933.87,361.72) and (936.98,364.82) .. (936.98,368.66) -- (936.98,466.04) .. controls (936.98,469.87) and (933.87,472.98) .. (930.04,472.98) -- (909.22,472.98) .. controls (905.39,472.98) and (902.28,469.87) .. (902.28,466.04) -- (902.28,368.66) .. controls (902.28,364.82) and (905.39,361.72) .. (909.22,361.72) -- cycle ;
\draw    (919.57,339.43) -- (919.5,361.5) ;
\draw    (799.57,339.43) -- (799.5,397.5) ;
\draw  [fill={rgb, 255:red, 255; green, 242; blue, 204 }  ,fill opacity=1 ] (846.5,340.6) .. controls (846.5,333.97) and (851.65,328.6) .. (858,328.6) .. controls (864.35,328.6) and (869.5,333.97) .. (869.5,340.6) .. controls (869.5,347.23) and (864.35,352.6) .. (858,352.6) .. controls (851.65,352.6) and (846.5,347.23) .. (846.5,340.6) -- cycle (835,340.6) -- (846.5,340.6) (881,340.6) -- (869.5,340.6) ;
\draw    (799.5,340.6) -- (844.5,340.6) ;
\draw [shift={(846.5,340.6)}, rotate = 180] [color={rgb, 255:red, 0; green, 0; blue, 0 }  ][line width=0.75]    (10.93,-3.29) .. controls (6.95,-1.4) and (3.31,-0.3) .. (0,0) .. controls (3.31,0.3) and (6.95,1.4) .. (10.93,3.29)   ;
\draw    (919.5,340.6) -- (871.5,340.6) ;
\draw [shift={(869.5,340.6)}, rotate = 360] [color={rgb, 255:red, 0; green, 0; blue, 0 }  ][line width=0.75]    (10.93,-3.29) .. controls (6.95,-1.4) and (3.31,-0.3) .. (0,0) .. controls (3.31,0.3) and (6.95,1.4) .. (10.93,3.29)   ;
\draw    (858,328) -- (858,311) ;
\draw [shift={(858,309)}, rotate = 450] [color={rgb, 255:red, 0; green, 0; blue, 0 }  ][line width=0.75]    (10.93,-3.29) .. controls (6.95,-1.4) and (3.31,-0.3) .. (0,0) .. controls (3.31,0.3) and (6.95,1.4) .. (10.93,3.29)   ;
\draw  [fill={rgb, 255:red, 213; green, 232; blue, 212 }  ,fill opacity=1 ] (825,259.34) .. controls (825,252.52) and (830.52,247) .. (837.34,247) -- (877.66,247) .. controls (884.48,247) and (890,252.52) .. (890,259.34) -- (890,296.36) .. controls (890,303.17) and (884.48,308.7) .. (877.66,308.7) -- (837.34,308.7) .. controls (830.52,308.7) and (825,303.17) .. (825,296.36) -- cycle ;
\draw    (1019.57,196.43) -- (1019.5,385.5) ;
\draw  [fill={rgb, 255:red, 255; green, 242; blue, 204 }  ,fill opacity=1 ] (916.5,197.6) .. controls (916.5,190.97) and (921.65,185.6) .. (928,185.6) .. controls (934.35,185.6) and (939.5,190.97) .. (939.5,197.6) .. controls (939.5,204.23) and (934.35,209.6) .. (928,209.6) .. controls (921.65,209.6) and (916.5,204.23) .. (916.5,197.6) -- cycle (905,197.6) -- (916.5,197.6) (951,197.6) -- (939.5,197.6) ;
\draw    (858.5,197.6) -- (914.5,197.6) ;
\draw [shift={(916.5,197.6)}, rotate = 180] [color={rgb, 255:red, 0; green, 0; blue, 0 }  ][line width=0.75]    (10.93,-3.29) .. controls (6.95,-1.4) and (3.31,-0.3) .. (0,0) .. controls (3.31,0.3) and (6.95,1.4) .. (10.93,3.29)   ;
\draw    (1019.5,197.6) -- (941.5,197.6) ;
\draw [shift={(939.5,197.6)}, rotate = 360] [color={rgb, 255:red, 0; green, 0; blue, 0 }  ][line width=0.75]    (10.93,-3.29) .. controls (6.95,-1.4) and (3.31,-0.3) .. (0,0) .. controls (3.31,0.3) and (6.95,1.4) .. (10.93,3.29)   ;
\draw    (928,185) -- (928,167) ;
\draw [shift={(928,165)}, rotate = 450] [color={rgb, 255:red, 0; green, 0; blue, 0 }  ][line width=0.75]    (10.93,-3.29) .. controls (6.95,-1.4) and (3.31,-0.3) .. (0,0) .. controls (3.31,0.3) and (6.95,1.4) .. (10.93,3.29)   ;
\draw  [fill={rgb, 255:red, 213; green, 232; blue, 212 }  ,fill opacity=1 ] (873,126.54) .. controls (873,121.27) and (877.27,117) .. (882.54,117) -- (974.72,117) .. controls (979.99,117) and (984.26,121.27) .. (984.26,126.54) -- (984.26,155.16) .. controls (984.26,160.42) and (979.99,164.7) .. (974.72,164.7) -- (882.54,164.7) .. controls (877.27,164.7) and (873,160.42) .. (873,155.16) -- cycle ;
\draw    (858.5,197.6) -- (858.43,247.67) ;
\draw  [draw opacity=0][fill={rgb, 255:red, 255; green, 255; blue, 255 }  ,fill opacity=1 ] (603.53,239.61) .. controls (603.53,233.62) and (608.38,228.76) .. (614.38,228.76) -- (780.68,228.76) .. controls (786.67,228.76) and (791.53,233.62) .. (791.53,239.61) -- (791.53,279.91) .. controls (791.53,285.91) and (786.67,290.76) .. (780.68,290.76) -- (614.38,290.76) .. controls (608.38,290.76) and (603.53,285.91) .. (603.53,279.91) -- cycle ;
\draw    (544.01,273.5) -- (817.01,274.82)(543.99,276.5) -- (816.99,277.82) ;
\draw [shift={(825,276.36)}, rotate = 180.28] [color={rgb, 255:red, 0; green, 0; blue, 0 }  ][line width=0.75]    (10.93,-3.29) .. controls (6.95,-1.4) and (3.31,-0.3) .. (0,0) .. controls (3.31,0.3) and (6.95,1.4) .. (10.93,3.29)   ;
\draw  [color={rgb, 255:red, 75; green, 172; blue, 162 }  ,draw opacity=1 ] (28,52) -- (1085.83,52) -- (1085.83,579.47) -- (28,579.47) -- cycle ;

\draw (73.69,534.04) node [anchor=north west][inner sep=0.75pt]  [font=\scriptsize] [align=left] {A};
\draw (86.36,534.04) node [anchor=north west][inner sep=0.75pt]  [font=\scriptsize] [align=left] {A};
\draw (99.03,534.07) node [anchor=north west][inner sep=0.75pt]  [font=\scriptsize] [align=left] {G};
\draw (112.54,534.04) node [anchor=north west][inner sep=0.75pt]  [font=\scriptsize] [align=left] {T};
\draw (124.36,534.07) node [anchor=north west][inner sep=0.75pt]  [font=\scriptsize] [align=left] {C};
\draw (137.03,534.07) node [anchor=north west][inner sep=0.75pt]  [font=\scriptsize] [align=left] {C};
\draw (150.54,534.07) node [anchor=north west][inner sep=0.75pt]  [font=\scriptsize] [align=left] {T};
\draw (162.36,534.04) node [anchor=north west][inner sep=0.75pt]  [font=\scriptsize] [align=left] {C};
\draw (175.88,534.04) node [anchor=north west][inner sep=0.75pt]  [font=\scriptsize] [align=left] {A};
\draw (187.84,534.07) node [anchor=north west][inner sep=0.75pt]  [font=\scriptsize] [align=left] {A};
\draw (200.37,534.04) node [anchor=north west][inner sep=0.75pt]  [font=\scriptsize] [align=left] {T};
\draw (213.03,534.04) node [anchor=north west][inner sep=0.75pt]  [font=\scriptsize] [align=left] {G};
\draw (226.54,534.07) node [anchor=north west][inner sep=0.75pt]  [font=\scriptsize] [align=left] {A};
\draw (238.37,534.07) node [anchor=north west][inner sep=0.75pt]  [font=\scriptsize] [align=left] {C};
\draw (251.88,534.07) node [anchor=north west][inner sep=0.75pt]  [font=\scriptsize] [align=left] {T};
\draw (263.98,534.07) node [anchor=north west][inner sep=0.75pt]  [font=\scriptsize] [align=left] {T};
\draw (277.21,534.07) node [anchor=north west][inner sep=0.75pt]  [font=\scriptsize] [align=left] {A};
\draw (288.19,534.07) node [anchor=north west][inner sep=0.75pt]  [font=\scriptsize] [align=left] {G};
\draw (54.52,526.7) node [anchor=north west][inner sep=0.75pt]   [align=left] {...};
\draw (305.05,526.7) node [anchor=north west][inner sep=0.75pt]   [align=left] {...};
\draw (152,477.6) node [anchor=north west][inner sep=0.75pt]   [align=left] {Embedding};
\draw (150,375.6) node [anchor=north west][inner sep=0.75pt]   [align=left] {Self Attention};
\draw (156,312.6) node [anchor=north west][inner sep=0.75pt]   [align=left] {Add \& Norm};
\draw (155,204.6) node [anchor=north west][inner sep=0.75pt]   [align=left] {Add \& Norm};
\draw (149,252.6) node [anchor=north west][inner sep=0.75pt]   [align=left] {Feed-Forward};
\draw (96,132.6) node [anchor=north west][inner sep=0.75pt]   [align=left] {Contextualized Embedding};
\draw (324,340) node [anchor=north west][inner sep=0.75pt]   [align=left] {\textit{N×}};
\draw (496,494.6) node [anchor=north west][inner sep=0.75pt]   [align=left] {Input};
\draw (395,404.6) node [anchor=north west][inner sep=0.75pt]   [align=left] {$\displaystyle \mathit{Q:n\times d_{k}}$};
\draw (518.93,344.31) node [anchor=north west][inner sep=0.75pt]   [align=left] {$ $};
\draw (520,526.6) node [anchor=north west][inner sep=0.75pt]   [align=left] {$\displaystyle \mathit{X:n\times d}$};
\draw (376,380.6) node [anchor=north west][inner sep=0.75pt]   [align=left] {Query};
\draw (518,343.6) node [anchor=north west][inner sep=0.75pt]   [align=left] {Key};
\draw (534,392.6) node [anchor=north west][inner sep=0.75pt]   [align=left] {$\displaystyle \mathit{K^{T} :}$};
\draw (534,412.6) node [anchor=north west][inner sep=0.75pt]  [font=\scriptsize] [align=left] {$\displaystyle \mathit{d_{k} \times n}$};
\draw (598,366.6) node [anchor=north west][inner sep=0.75pt]   [align=left] {Value};
\draw (619,399.6) node [anchor=north west][inner sep=0.75pt]   [align=left] {$\displaystyle \mathit{V:n\times d_{v}}$};
\draw (477,327.6) node [anchor=north west][inner sep=0.75pt]  [font=\large] [align=left] {$\displaystyle \times $};
\draw (390,170.6) node [anchor=north west][inner sep=0.75pt]   [align=left] {Pairwise\\Similarities};
\draw (450,252.6) node [anchor=north west][inner sep=0.75pt]   [align=left] {$\displaystyle \mathit{\rho (S):n\times n}$};
\draw (547,185.6) node [anchor=north west][inner sep=0.75pt]  [font=\large] [align=left] {$\displaystyle \times $};
\draw (504,99.6) node [anchor=north west][inner sep=0.75pt]   [align=left] {Output};
\draw (528,131.6) node [anchor=north west][inner sep=0.75pt]   [align=left] {$\displaystyle \mathit{Y:n\times d_{v}}$};
\draw (866,494.6) node [anchor=north west][inner sep=0.75pt]   [align=left] {Input};
\draw (765,404.6) node [anchor=north west][inner sep=0.75pt]   [align=left] {$\displaystyle \mathit{Q:n\times d_{k}}$};
\draw (888.93,344.31) node [anchor=north west][inner sep=0.75pt]   [align=left] {$ $};
\draw (890,526.6) node [anchor=north west][inner sep=0.75pt]   [align=left] {$\displaystyle \mathit{X:n\times d}$};
\draw (746,380.6) node [anchor=north west][inner sep=0.75pt]   [align=left] {Query};
\draw (888,343.6) node [anchor=north west][inner sep=0.75pt]   [align=left] {Key};
\draw (904,392.6) node [anchor=north west][inner sep=0.75pt]   [align=left] {$\displaystyle \mathit{\hat{K}^{T} :}$};
\draw (904,412.6) node [anchor=north west][inner sep=0.75pt]  [font=\scriptsize] [align=left] {$\displaystyle \mathit{d_{k} \times k}$};
\draw (968,366.6) node [anchor=north west][inner sep=0.75pt]   [align=left] {Value};
\draw (989,399.6) node [anchor=north west][inner sep=0.75pt]   [align=left] {$\displaystyle \mathit{\hat{V}:k\times d_{v}}$};
\draw (847,328.6) node [anchor=north west][inner sep=0.75pt]  [font=\large] [align=left] {$\displaystyle \times $};
\draw (774,181.6) node [anchor=north west][inner sep=0.75pt]   [align=left] {Low-Rank\\Pairwise\\Similarities};
\draw (832,258.6) node [anchor=north west][inner sep=0.75pt]   [align=left] {$\displaystyle  \begin{array}{{>{\displaystyle}l}}
\mathit{\rho (S)}\\
\mathit{n \times k}
\end{array}$};
\draw (917,185.6) node [anchor=north west][inner sep=0.75pt]  [font=\large] [align=left] {$\displaystyle \times $};
\draw (874,99.6) node [anchor=north west][inner sep=0.75pt]   [align=left] {Output};
\draw (898,131.6) node [anchor=north west][inner sep=0.75pt]   [align=left] {$\displaystyle \mathit{Y:n\times d_{v}}$};
\draw (43,75.6) node [anchor=north west][inner sep=0.75pt]   [align=left] {{\large Transformer Model}};
\draw (373,75.6) node [anchor=north west][inner sep=0.75pt]   [align=left] {Transformer Attention};
\draw (733,75.6) node [anchor=north west][inner sep=0.75pt]   [align=left] {Linformer Attention};
\draw (600,235) node [anchor=north west][inner sep=0.75pt]   [align=center] {Reduce Complexity\\ O($N^2$) $\rightarrow$ O($N$)};
\end{tikzpicture}
}
\caption[Comparison of the attention mechanism in original transformer model and Linformer model]{Comparison of the attention mechanism in original transformer model and Linformer model}
\label{fig:linformer-model}
\end{figure}
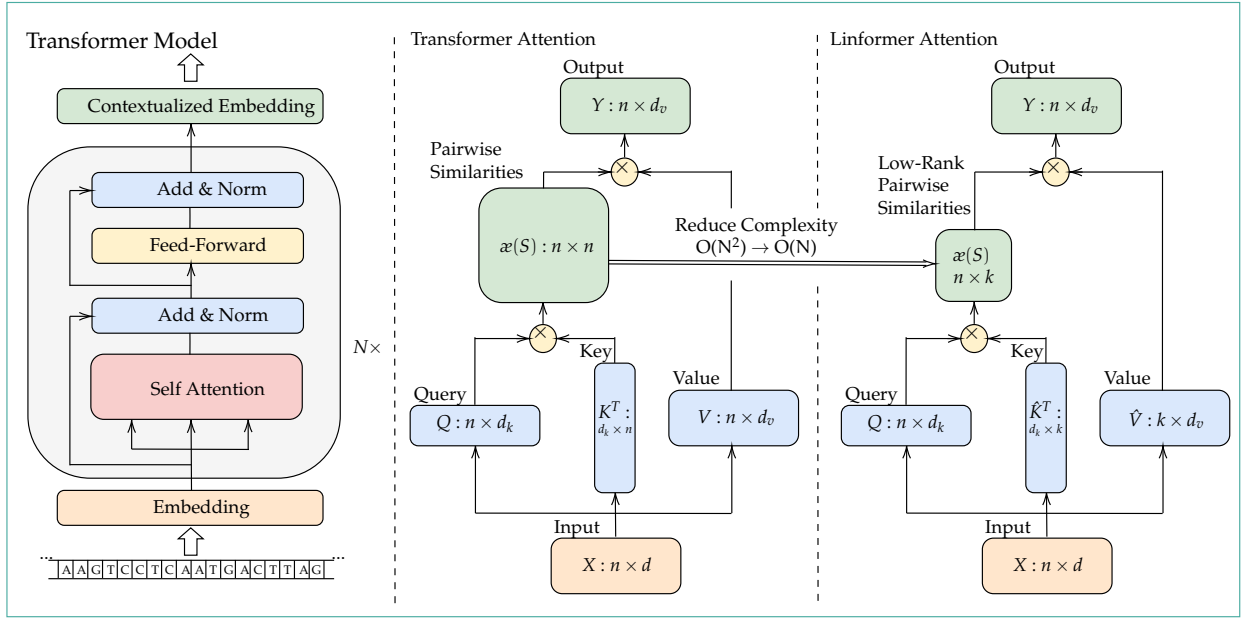

Linformer~\cite{wang2020linformer} performs a low-rank approximation on the self-attention mechanism of the transformer model. With this approach, the space and time complexity is reduced from  O($n^2$) to  O($n$), thus enabling the model to process an even longer sequences compared to the quadratic-attention mechanism.

As shown in Figure \ref{fig:linformer-model}, Linformer reduces the complexity of attention mechanism by performing low-rank approximation to the self-attention layer of a transformer model. Mathematically, A self-attention layer accepts an input sequence $X \in \mathbb{R}^{n \times d}$ an output another sequence $X \in \mathbb{R}^{n \times d_v}$. In the original transfomer model, the input sequence $X$ is first projected into three different sequences: query $Q \in \mathbb{R}^{n \times d_k}$, key $K \in \mathbb{R}^{n \times d_k}$, and value $V \in \mathbb{R}^{n \times d_v}$ by using query projection $W^Q \in \mathbb{R}^{d \times d_k}$, key projection $W^K \in \mathbb{R}^{d \times d_k}$, and value projection $W^V \in \mathbb{R}^{d \times d_v}$, respectively. A scaled-dot product attention is then applied from the three sequences to produce the sequence $Y$. The formula for calculating the self-attention mechanism in Transformer is as follow:

\begin{align}
&Q = XW^Q, K = XW^K, V = XW^V\\
&Y=\textnormal{Softmax}(\frac{QK^T}{\sqrt{d_k}}V)
\end{align}

While in Linformer model, after $X$ is projected to $Q$, $K$, and $V$, another projection from $K \in \mathbb{R}^{n \times d_k}$ and $V \in \mathbb{R}^{n \times d_v}$ with $W^{\hat{K}} \in \mathbb{R}^{k \times n}$ and $W^{\hat{V}} \in \mathbb{R}^{k \times n}$, respectively. The projection produces a low-rank key sequence $\hat{K} \in \mathbb{R}^{k \times d_k}$ and a low-rank value sequence $\hat{V} \in \mathbb{R}^{k \times d_v}$. The complete formula for calculating the self-attention mechanism in Linformer is as follow: 

\begin{align}
&Q = XW^Q, K = XW^K, V = XW^V\\
&\hat{K} = W^{\hat{K}}K, \hat{V} = W^{\hat{V}}V\\
&Y=\textnormal{Softmax}(\frac{Q\hat{K}^T}{\sqrt{d_k}}\hat{V})
\end{align}

The additional projection steps reduce the space and time complexity of computing $Y$ from O($n^2$) to O($nk$) which can also be written as O($n$) as $k$ is a constant. If we choose the rank $k$ to be very low $k<<n$, the Linformer model could get a huge reduction in terms of both computation time and memory usage.

\section{Experimental Setup}

We capture all metrics that impact the efficiency for both training and inference. Specifically, we report 5 efficiency metrics, i.e, 1) the time required to perform forward and backward passes to measure the time efficiency in the training step, 2) the memory usage required to perform forward pass with all of the cached activations for performing backward pass to measure the memory efficiency on the training step, 3) the time required to perform a forward pass without dropout and caching any activations to measure the time efficiency during the inference step, 4) the memory usage required to perform forward pass without caching any activations to measure the memory efficiency on the inference step, and 5) the number of parameters over different low-rank factor $r$ to measure the storage efficiency of the model.

We compare the efficiency of our methods with the standard Transformer model. We benchmark the speed and memory usage on a single 12GB NVIDIA TITAN X (Pascal) GPU card. For the input sequence, we randomly generate data up to a certain sequence length $n$ and calculate each metric averaged across 30 runs each. Similar to~\cite{wang2020linformer}, when measuring the time efficiency, we select batch size based on the maximum batch size that can fit the GPU memory for the standard transformer model. We fix the hyperparameter for all models to ensure a fair comparison between each model. Specifically, for each model variant, we use 2-layer encoder layers, with hidden dimension of 768, feed-forward size of 3072, and number of heads of 12.

Additionally, we compare the effectiveness between LRT, Linformer, and Transformer models on the MNIST dataset to provide empirical evidence of the effectiveness of the two efficient transformer variations. Following ~\cite{tay2021long}, we flatten image data in MNIST from $28 \times 28$ pixels into sequences with a length of 784. We append additional an \texttt{[CLS]} token as the sequence prefix. We employ a Transformer model with 4 transformer encoder layers, model dimension of 256, 8 heads, and feed-forward size of 1024. For LRT and Linformer, we utilize low-rank factorization with rank $r=64$.

\section{Result and Discussion}

\subsection{Efficiency comparison to Transformer Model}

\begin{figure*}[h!]
    \centering
    Low-Rank Transformer (LRT) \vspace{2pt}
    \resizebox{0.95\textwidth}{!}{  
        \includegraphics{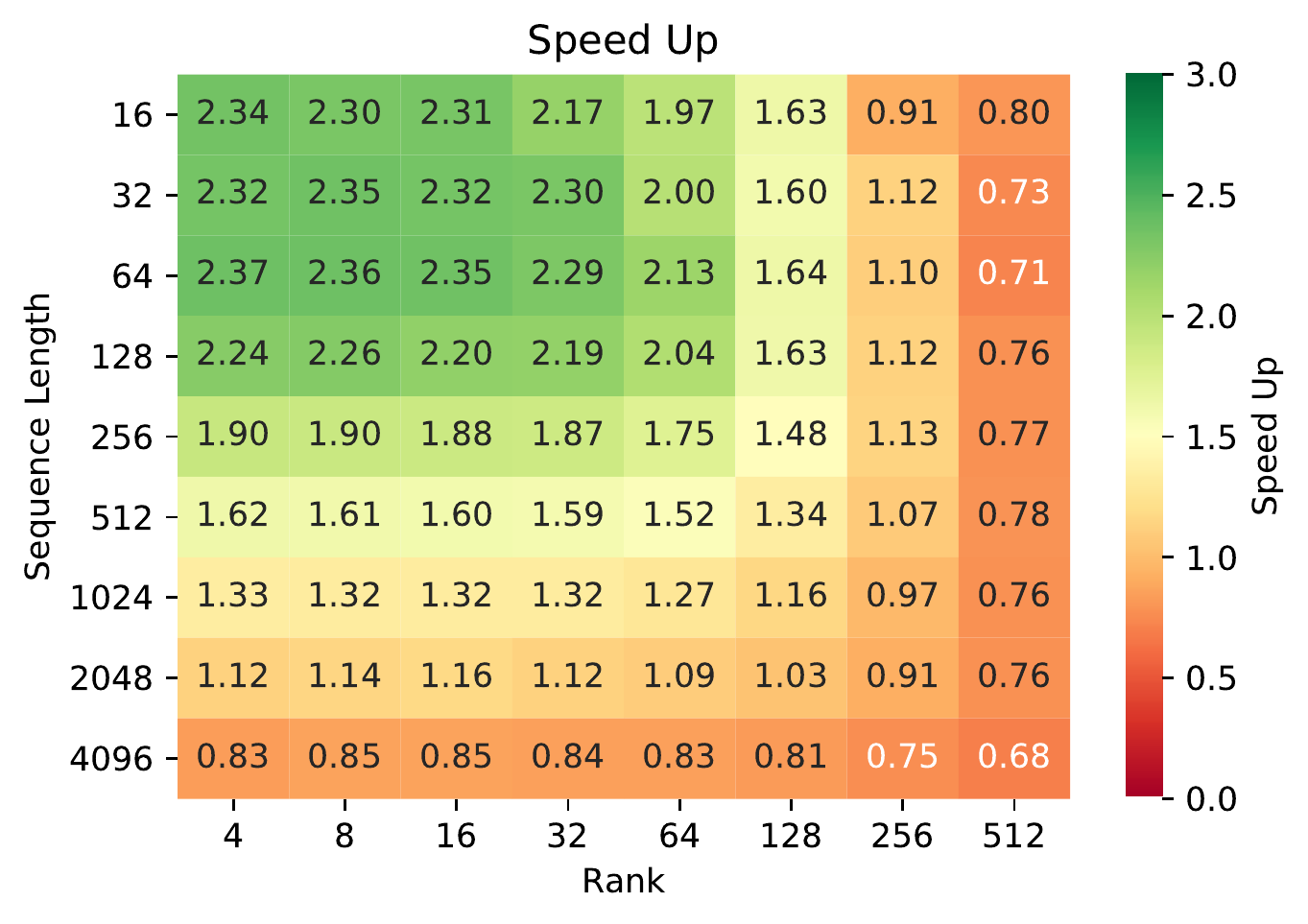}
        \includegraphics{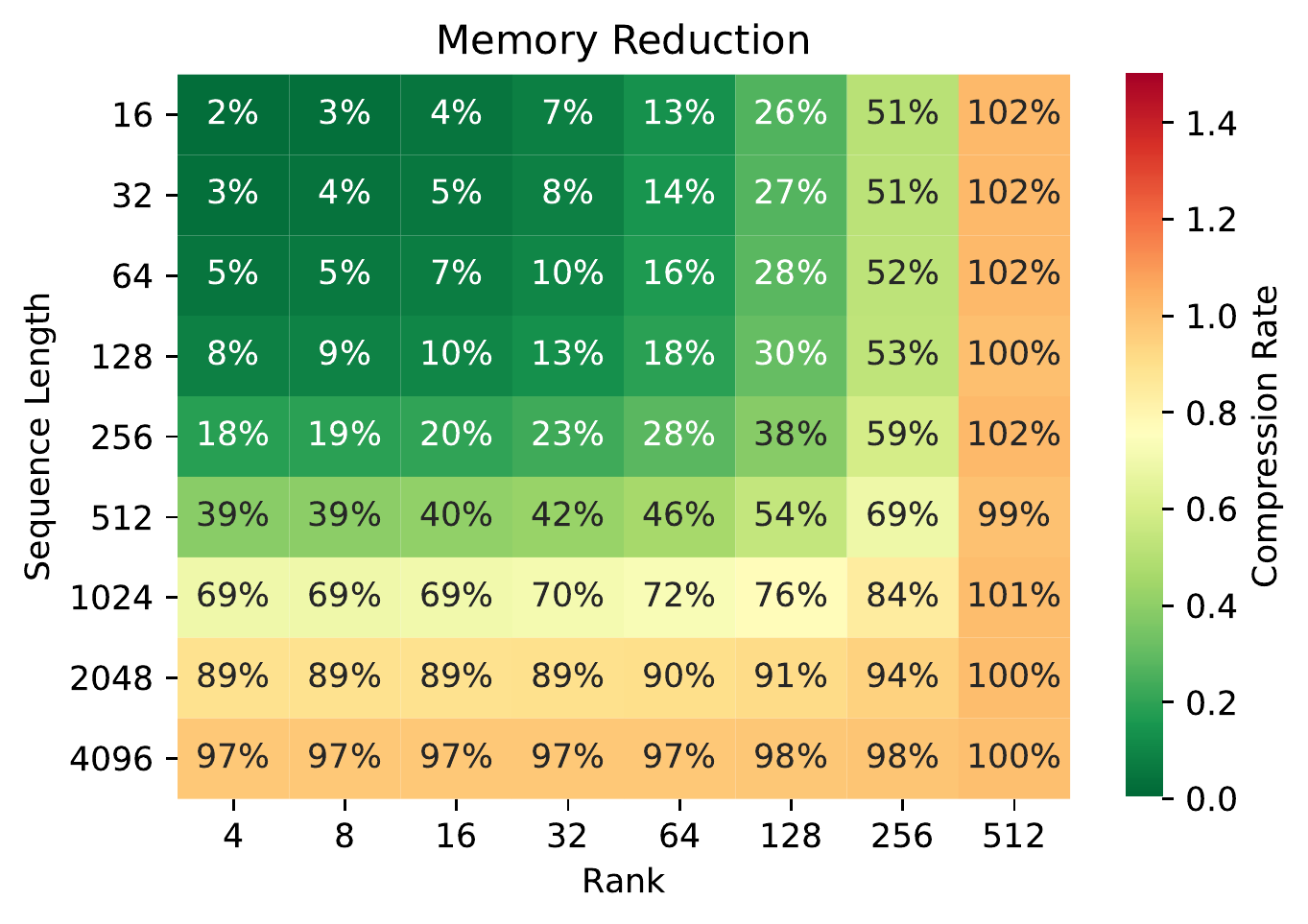}
    }
    \vspace{4pt}
    Linformer Model \vspace{2pt}
    \resizebox{0.95\textwidth}{!}{  
        \includegraphics{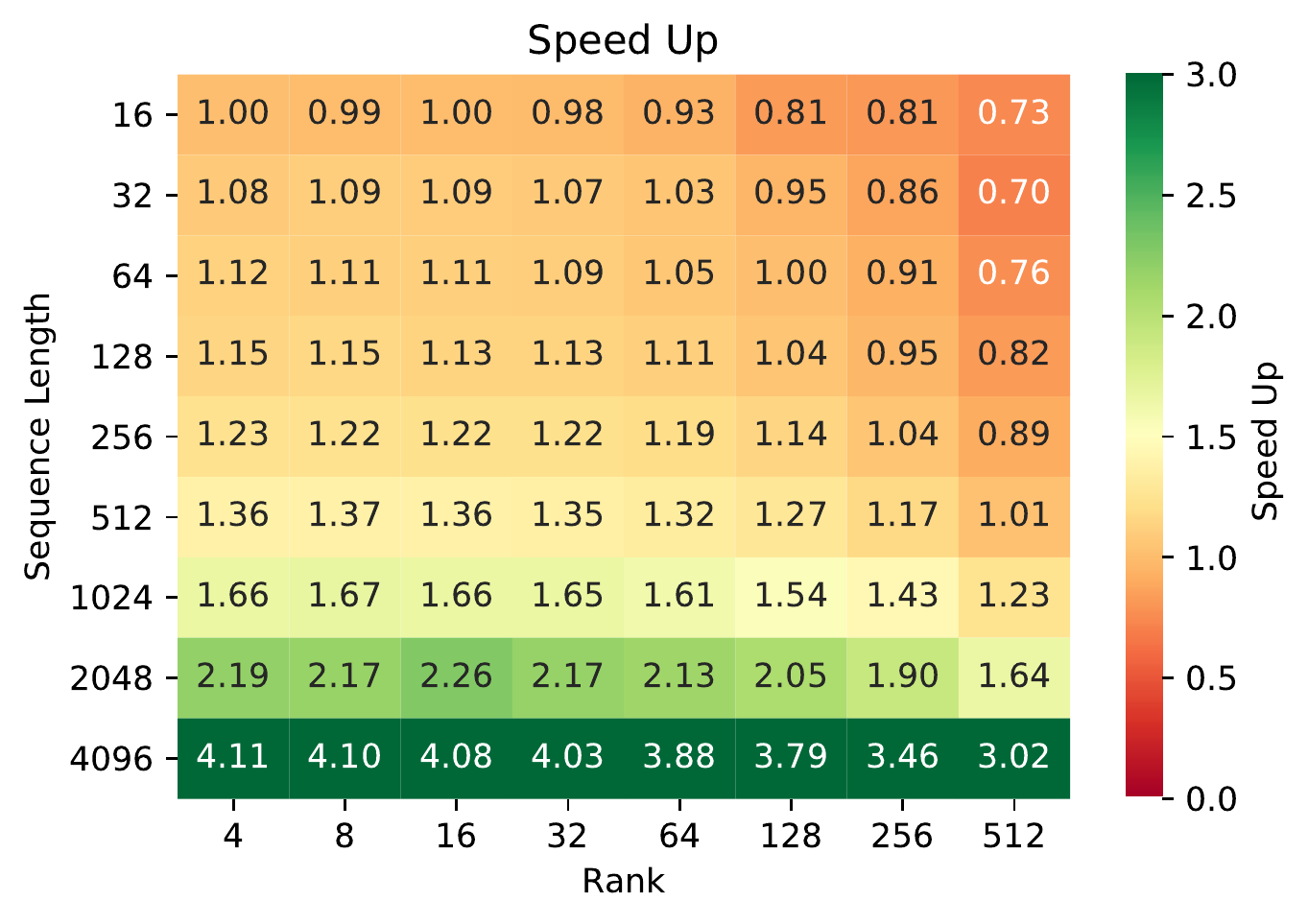}
        \includegraphics{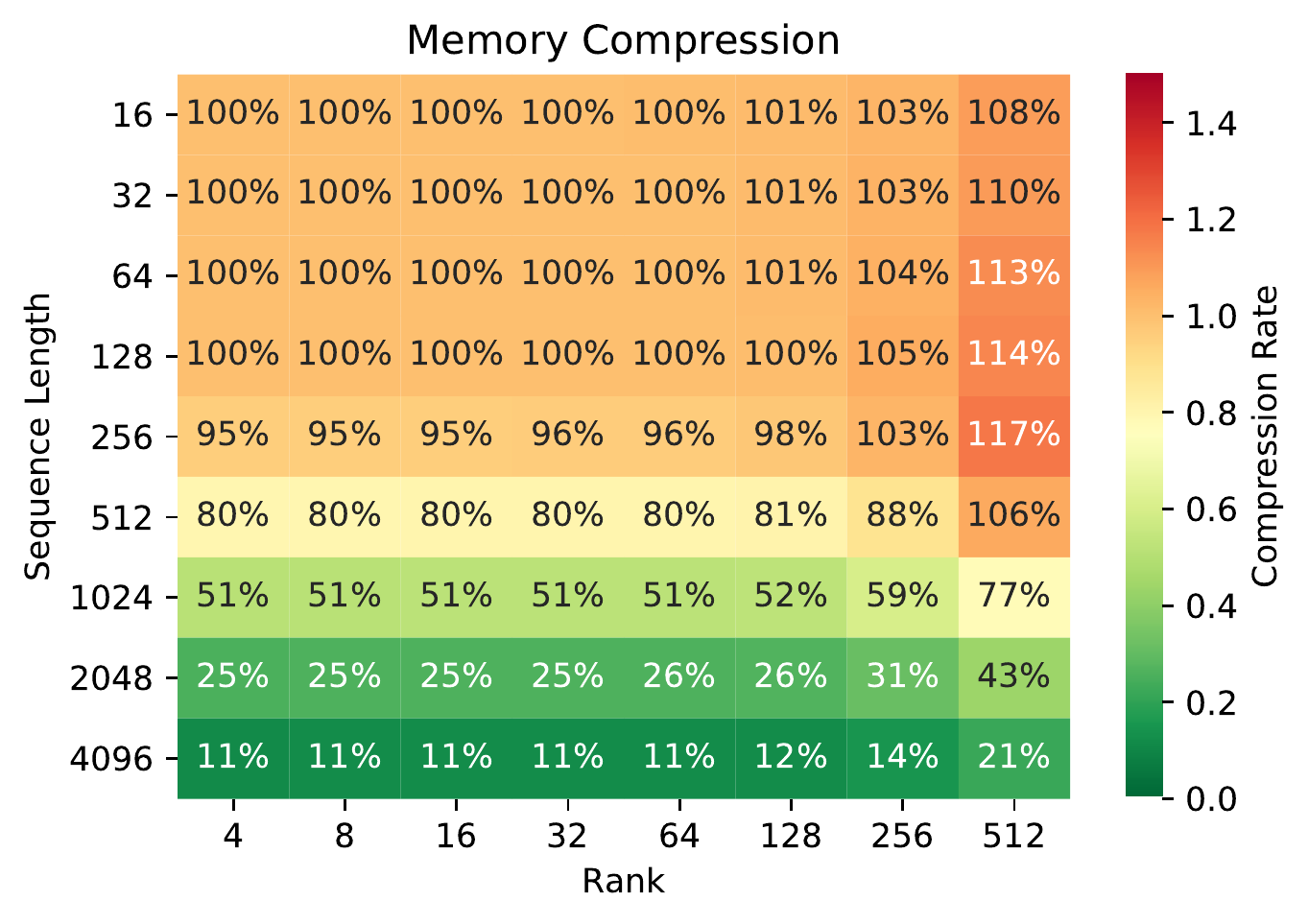}
    }
    \caption[Inference speed up and memory efficiency of LRT and Linformer compared to the Transformer baseline]{Inference speed up and memory efficiency of LRT \textbf{(Top)} and Linformer \textbf{(Bottom)} compared to the Transformer baseline. \textbf{(Left)} Computation speed up against the Transformer model, $\geq 1$ indicates faster or equal speed and slower otherwise. \textbf{(Right)} Memory usage compared to the Transformer model in percentage. Lower value indicates better memory compression.}
    \label{fig:inference_efficiency_vs_transformer}
\end{figure*}

We compare the computation and memory efficiency of our models to the Transformer baseline. As shown in Figure~\ref{fig:inference_efficiency_vs_transformer}, LRT shows significant improvement in terms of computation cost when the sequence length is $\leq 512$ and rank $\leq 128$. In terms of memory usage, LRT reduces $\sim$50\% of the memory usage when the sequence length is $\leq 512$ and rank $\leq 128$. While for the Linformer model, we observe a significant speedup and memory reduction when the sequence length is $\geq 256$ and achieve even higher speed up when the sequence length is larger due to its linear complexity. We also observe a similar trend for both LRT and Linformer models when simulating the training step with both forward and backward passes. This indicates that LRT is potential as an alternative of Transformer model when the sequence length is short ($\leq 512$) while Linformer is potential as an alternative of Transformer model when the sequence length is $\geq 512$.

\subsection{Short and Long Sequence Efficiency}

\begin{figure*}[h!]
    \centering
    Speed Up Ratio \vspace{2pt}
    \resizebox{0.95\textwidth}{!}{  
        \includegraphics{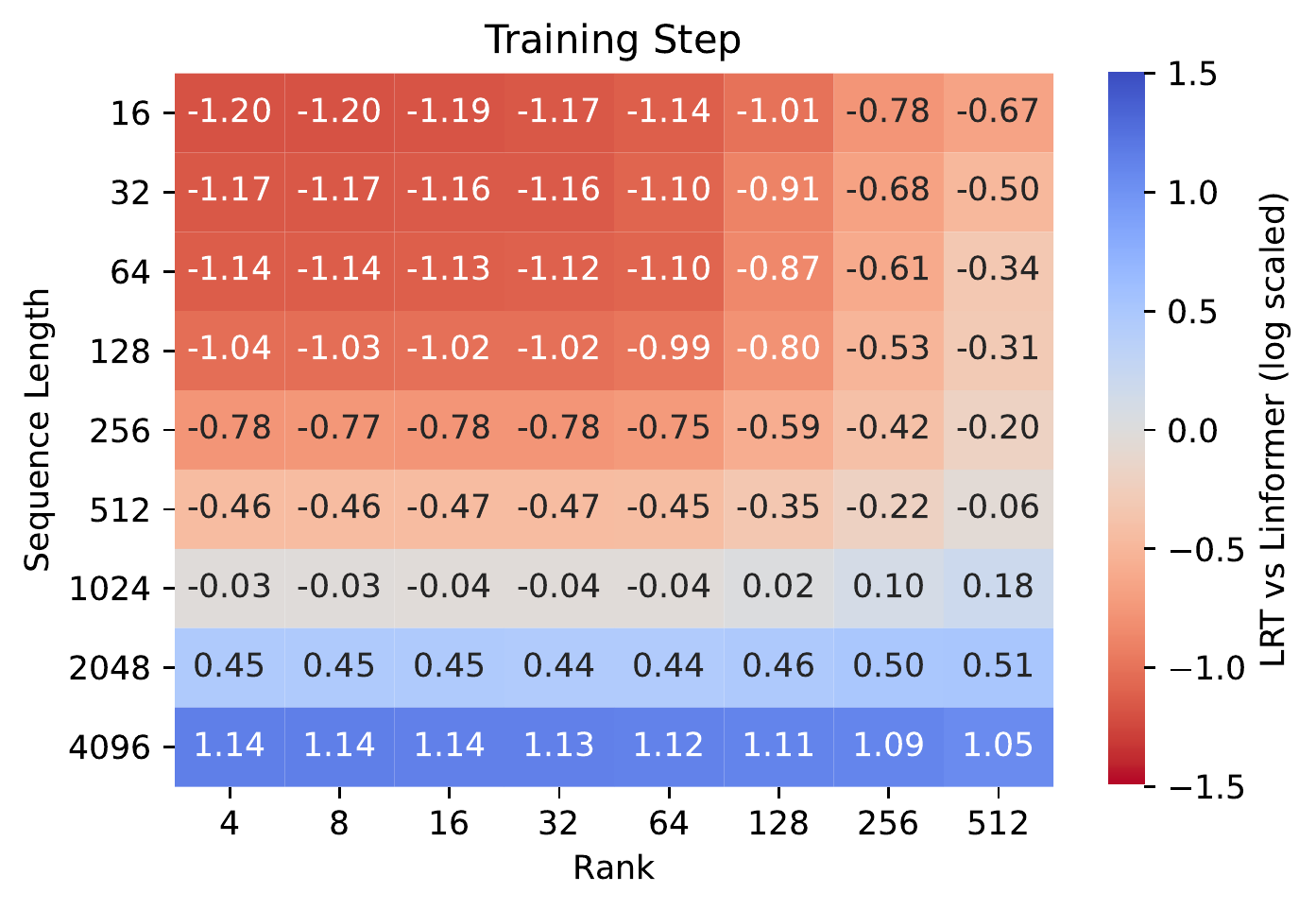}
        \includegraphics{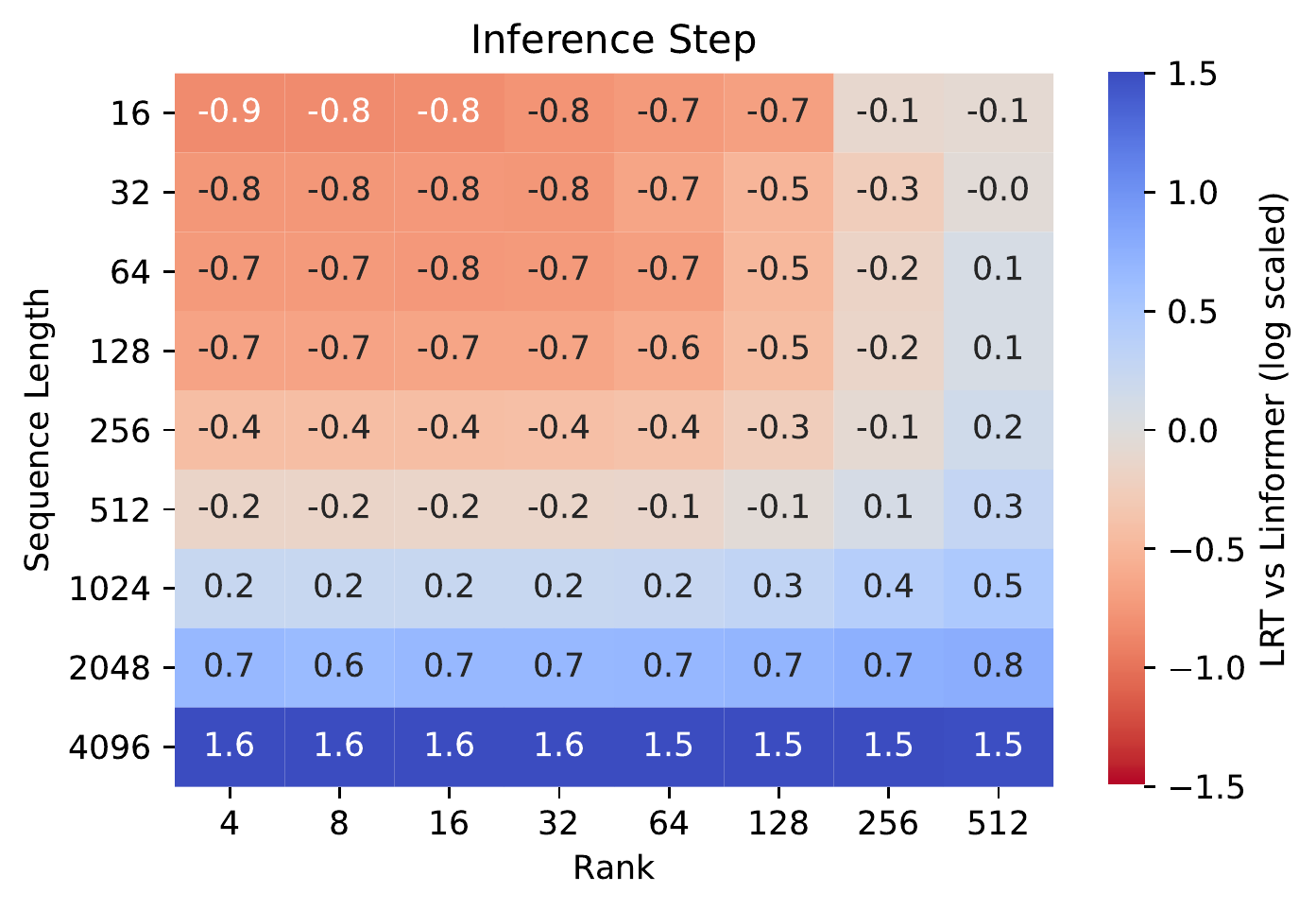}  
    }
    \vspace{4pt}
    Memory Compression Ratio\vspace{2pt}
    \resizebox{0.95\textwidth}{!}{  
        \includegraphics{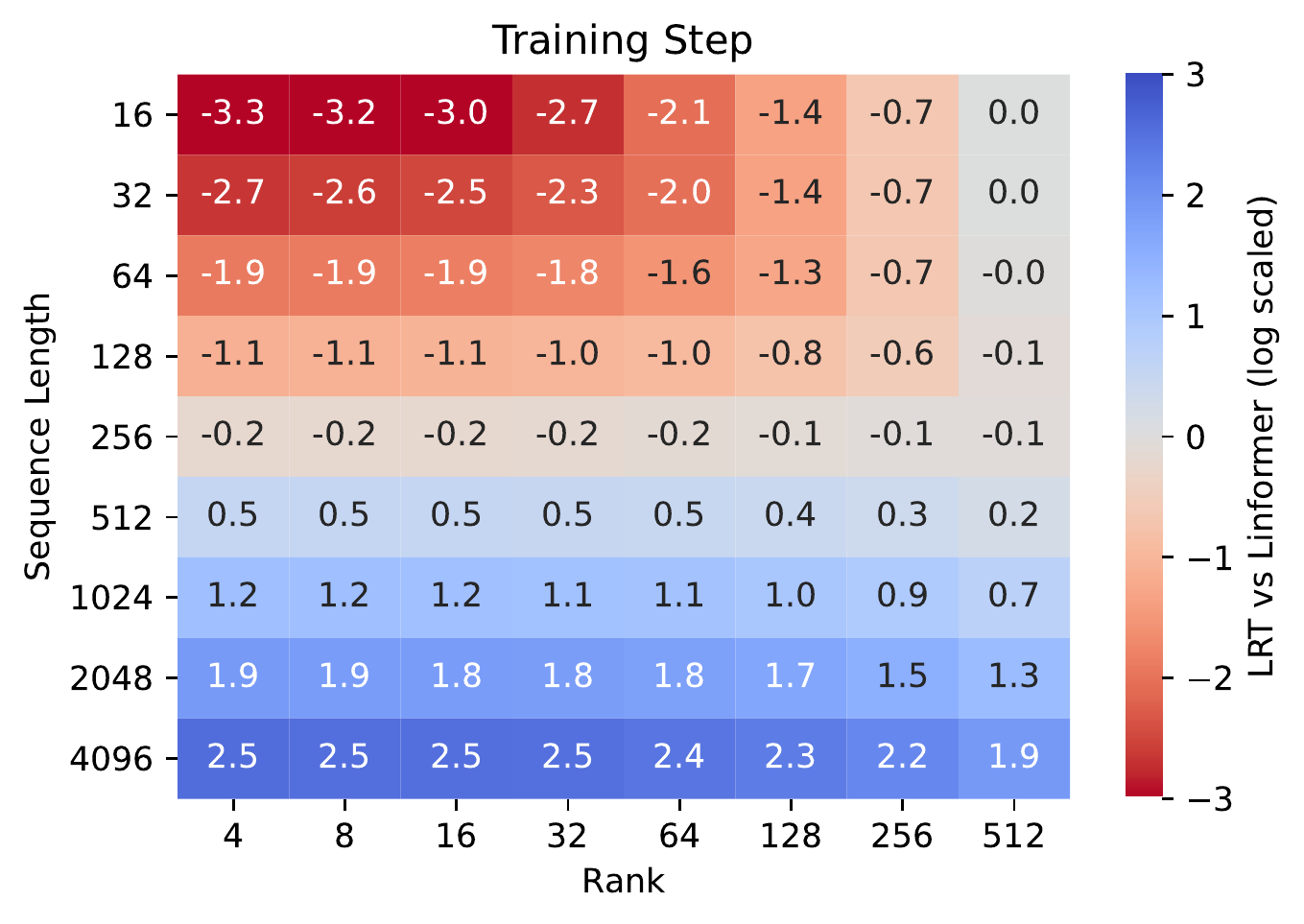}
        \includegraphics{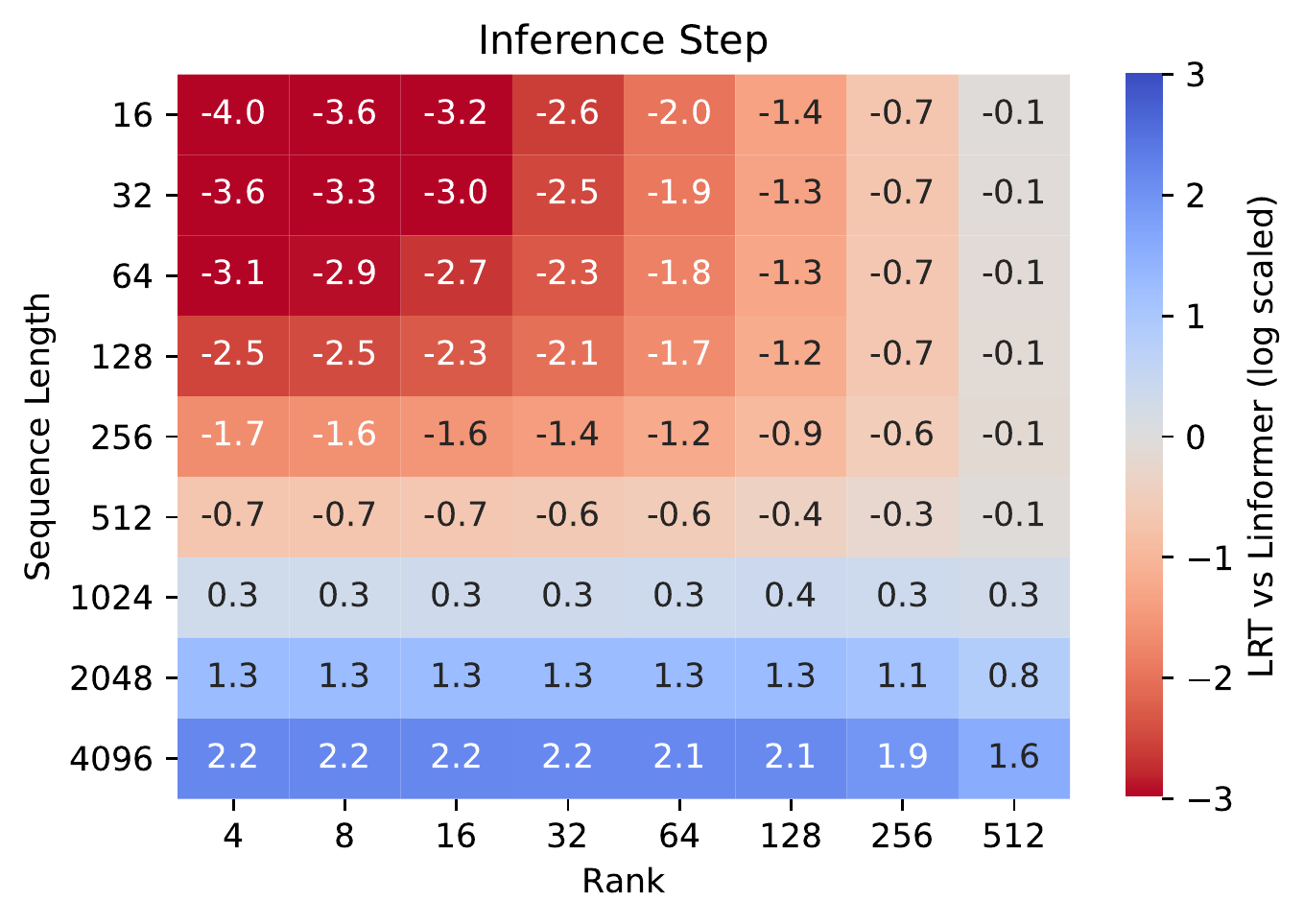}
    }    
    \caption[Speed up and memory compression ratios of LRT and Linformer]{Speed up \textbf{(Top)} and memory compression \textbf{(Bottom)} ratios of LRT and Linformer. \textbf{(Left)} Ratio in the training step. \textbf{(Right)} Ratio in the Inference step. Negative values indicate LRT yields better speed up or memory compression, while positive values indicate Linformer yields better speed up or memory compression.}
    \label{fig:ratio_efficiency}
\end{figure*}

As shown in Figure~\ref{fig:ratio_efficiency}, the LRT model yields better computation and memory efficiency when the sequence length is $\leq 256$, while the Linformer model yields better computation and memory efficiency when the sequence length is $\geq 1024$. For sequence length 512, LRT offers faster during the training step and provides better memory compression on the inference step, while Linformer offers a faster inference step and better memory reduction on the training step. Although this depends on the choice of other hyperparameters such as hidden size and number of heads, the trend indicates that the LRT model can improve the efficiency better when the input sequence is short, while the Linformer model can be an efficient alternative when the sequence length is long. With a larger hidden size configuration, we can expect that the efficiency of the LRT model will increase even further. Additionally, as the sequence length of the Linformer model is pre-defined as one of its parameters, it is also worth mentioning that variation of sequence length in the dataset might also affect the training and inference efficiency of the Linformer model. 

\subsection{Size Efficiency}

\begin{figure*}[h!]
    \centering
    \resizebox{0.65\textwidth}{!}{  
        \includegraphics{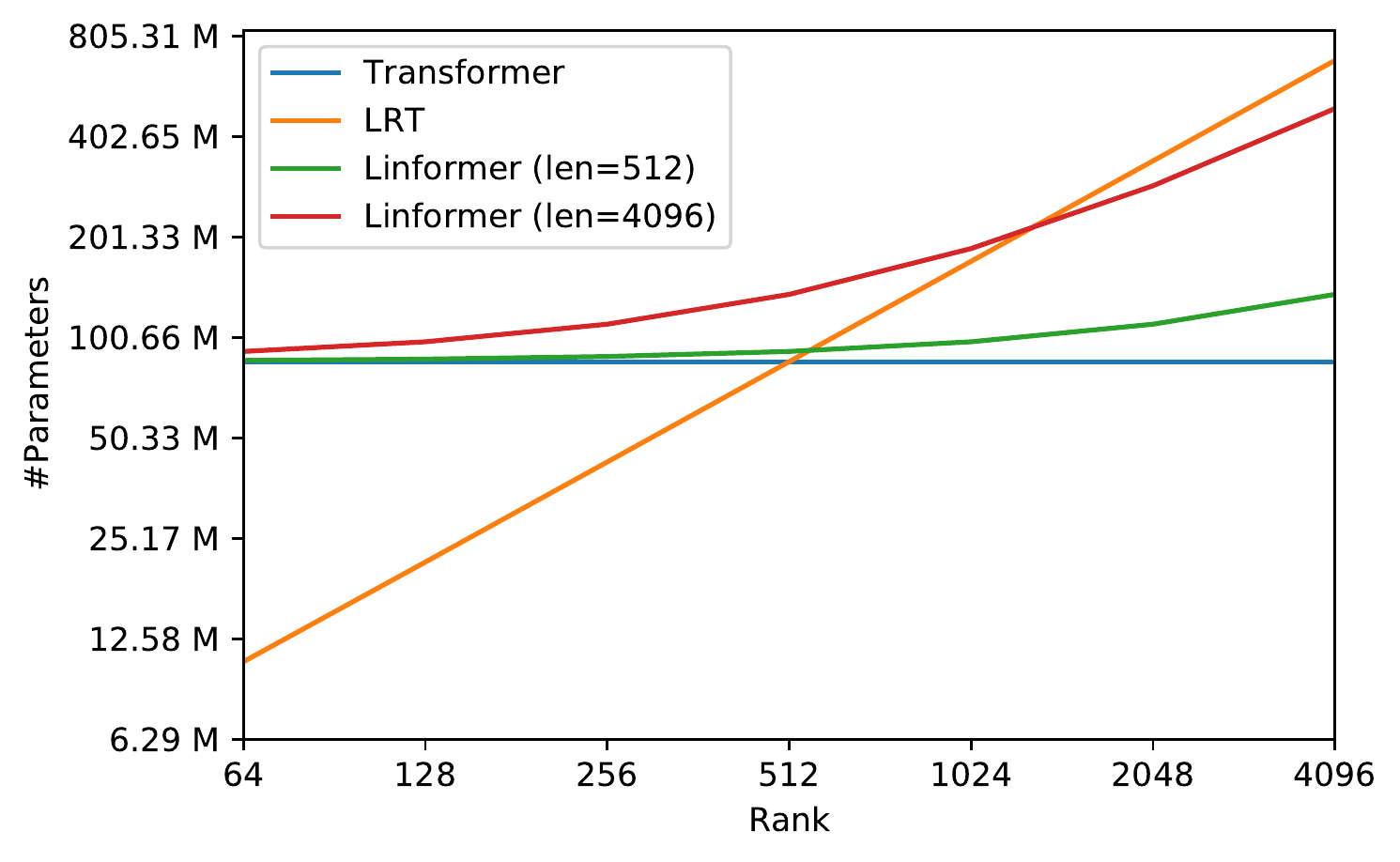}
    }
    \caption[Size comparison of LRT, Linformer, and Transformer model across different Rank]{Size comparison of LRT, Linformer, and Transformer model across different Rank. We visualize both axis in $log_2$ scale.}
    \label{fig:size_efficiency}
\end{figure*}

Other than computation and memory efficiency, the size of a model is also an important factor to consider for the deployment of a model especially in the on-device setting as it affects the amount of storage and networking cost required to deliver the model. As shown in Figure~\ref{fig:size_efficiency}, with the additional projection layer, the Linformer model always increases the model size regardless of its rank. On the other hand, the LRT model provides significant size reduction when the rank is small. For instance, with Rank $r=64$, the model size can be reduced up to $\sim$8$\times$ smaller and the size increases linearly along with the rank. This means the LRT model is a more viable option for on-device computing compared to Linformer and Transformer models.

\subsection{Effectiveness of LRT and Linformer model}

\begin{table}[!th]
\centering
\resizebox{0.65\textwidth}{!}{
    \begin{tabular}{lccc}
    \toprule
    \textbf{Model} & \textbf{\#Parameters} & \textbf{Training Acc.} & \textbf{Test Acc.} \\
    \midrule
    Transformer  & 11.2M & 98.72\% & 96.95\% \\
    LRT & 9.3M & 99.04\% & 97.31\% \\
    Linformer & 11.6M & 98.89\% & 96.82\% \\
    \bottomrule
    \end{tabular}
}
\caption[Evaluation comparison on MNIST dataset]{Evaluation comparison between Transformer, LRT, and Linformer models on MNIST dataset}
\label{tab:result-mnist}
\end{table}

As shown in Table \ref{tab:result-mnist}, both LRT and Linformer models perform as well as the Transformer models with much smaller computing time and memory cost on the MNIST dataset. This suggests that efficiency methods via low-rank approximation, both LRT and Linformer, can significantly improve the efficiency of the model without any loss of evaluation performance. Additionally, the number of parameters for the LRT model can achieve a slightly better score in terms of training and evaluation with much smaller parameters compared to the original Transformer model. This suggests that the bottleneck layer from the low-rank approximation can further improve the generalization of the model which increases the overall score of the low-rank factorized model.

\subsection{Impact of Greenformers}

\begin{figure*}[h!]
    \centering
    Training Efficiency of Low-Rank Transformer (LRT)
    \vspace{2pt}
    \resizebox{0.95\textwidth}{!}{  
        \includegraphics{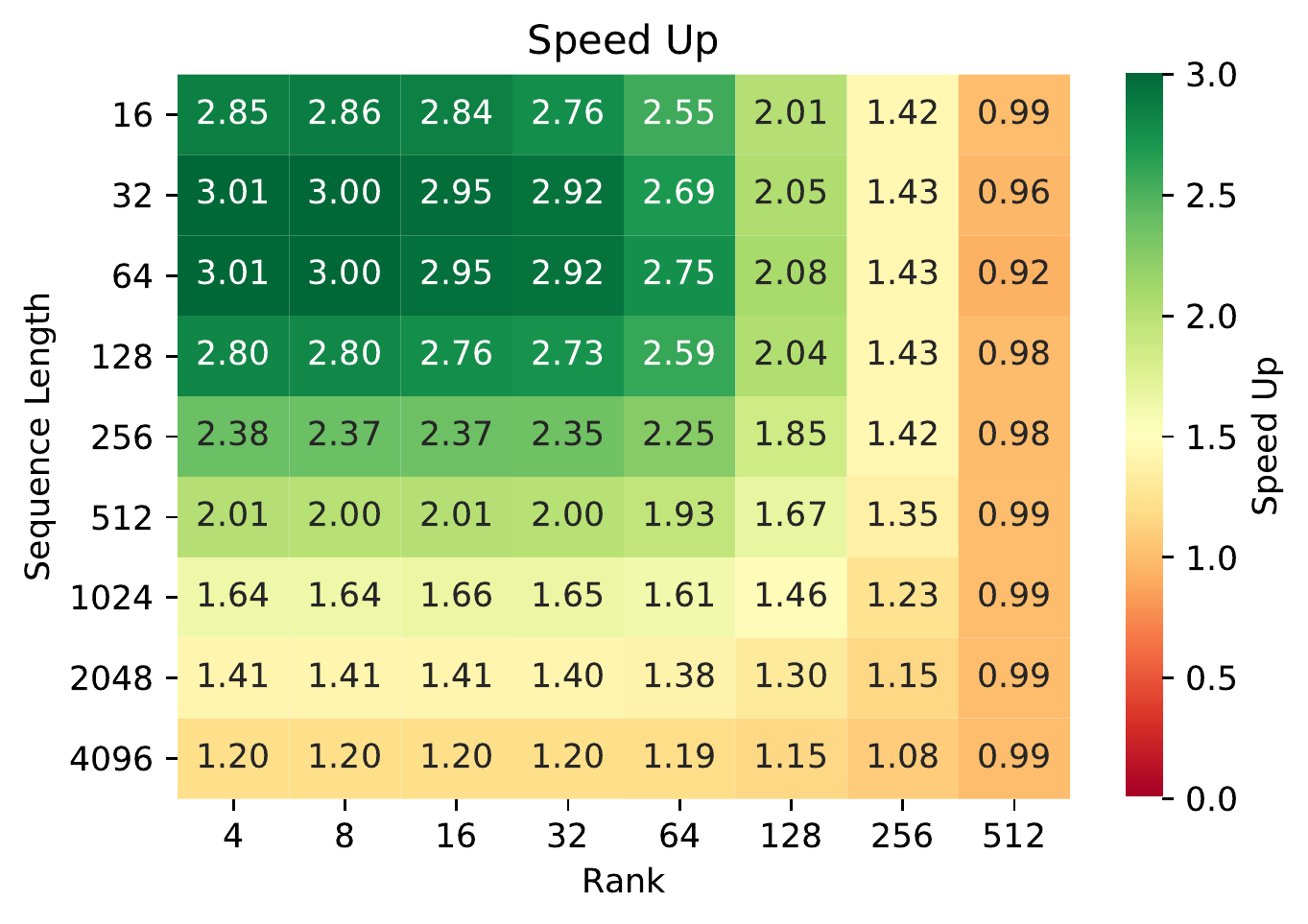}
        \includegraphics{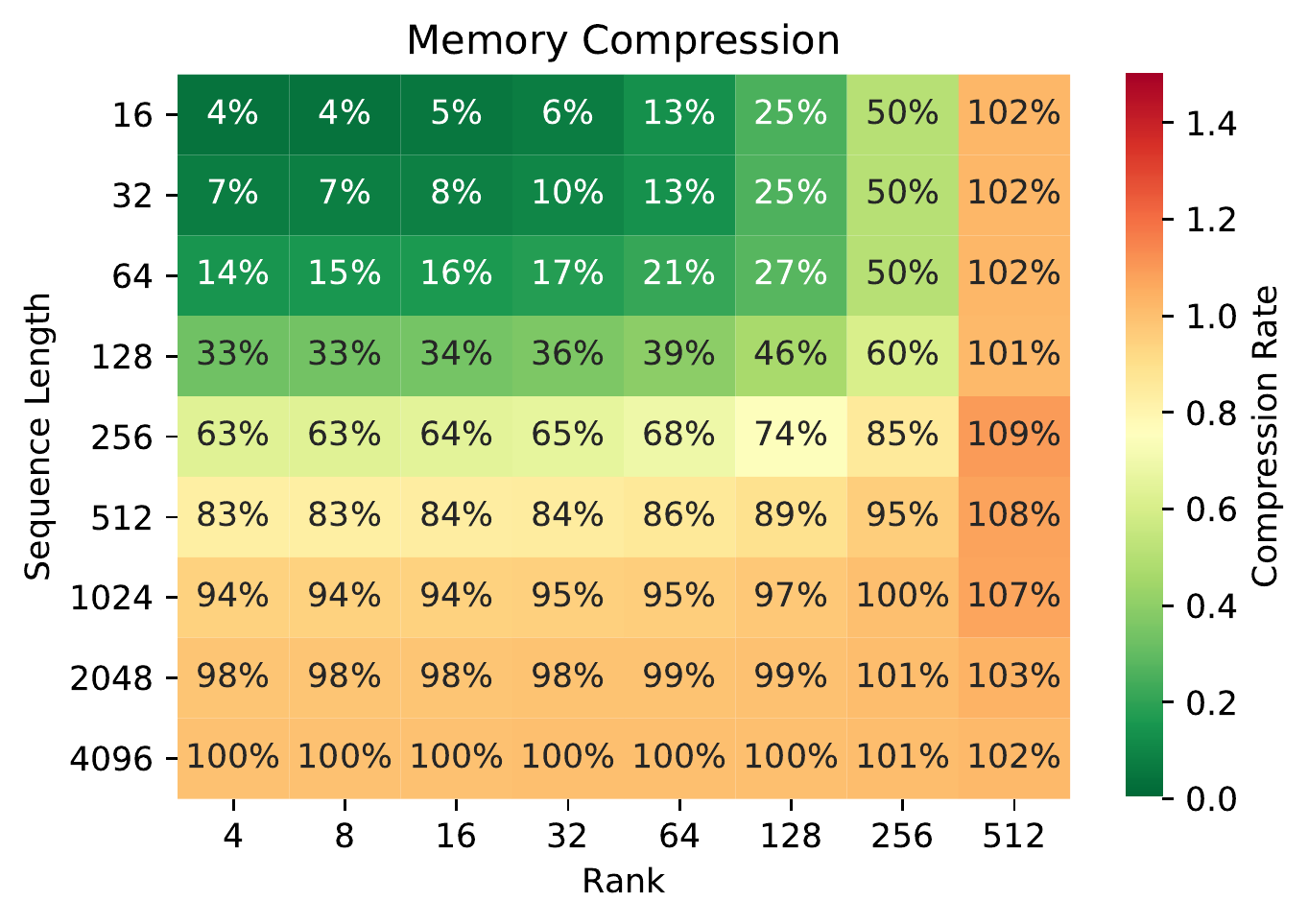}
    }
    \caption[Training speed up and memory efficiency of LRT compared to the Transformer baseline]{Training speed up and memory efficiency of LRT compared to the Transformer baseline. \textbf{(Left)} Computation speed up against the Transformer model, $\geq 1$ indicates faster or equal speed and slower otherwise. \textbf{(Right)} Memory usage compared to the Transformer model in percentage. Lower value indicates better memory compression.}
    \label{fig:training_efficiency_vs_transformer}
\end{figure*}

To further see the direct impact of Greenformers' efficiency methods, we will apply the training efficiency of the LRT model to the existing transformer-based model. As the hyperparameter setting of the transformer model used in the analysis follows BERT$_{BASE}$~\cite{devlin2019bert} model, we can directly apply the training efficiency factor shown in Figure \ref{fig:training_efficiency_vs_transformer} to improve the efficiency of the pre-training cost of the BERT$_{BASE}$ model. BERT model is pre-trained for 1,000,000 steps with a sequence length of 128 for 90\% of the time and a sequence length of 512 for the rest 10\% of the time. We calculate the overall efficiency factor of the LRT model compared to BERT$_{BASE}$ model with the following formula:

\begin{equation}
    Eff_r = \sum_k P_k E_{k,r}
\end{equation}

Where $Eff_r$ denotes the overall efficiency running LRT with rank $r$, $P_k$ denotes the percentage of time running pre-training with $k$ sequence long, and $E_{k,r}$ denotes the efficiency factor of running LRT on a sequence with length $k$ and rank $r$. By applying LRT with rank $r=\{32,64,128,256\}$, the model can achieve overall efficiency of 2.63, 2.50, 1.98, 1.48, respectively, which can provide significant computational, economical, and environmental costs reduction as shown in Table \ref{tab:efficiency_bert}.

\begin{table}[!h]
\centering
\resizebox{1.0\textwidth}{!}{
    \begin{tabular}{lccccc}
    \toprule
    \multirow{2}{*}{\textbf{Model}} & \textbf{Low-Rank} & \textbf{Efficiency} & \textbf{Computational Cost} & \textbf{Economical Cost} & \textbf{CO$_2$ emission} \\
     & \textbf{Factor} & \textbf{Factor} & \textbf{(petaflop/s-day)} & \textbf{(USD)} &  \textbf{(kg)} \\
    \midrule
    BERT$_{BASE}$ & - & 1.00 & 2.24 & \$2,074 - \$6,912 & 652.3 \\
    LRT-BERT$_{BASE}$ & 256 & 1.48 & 1.52 & \$1406 - \$4686 & 442.2 \\
    LRT-BERT$_{BASE}$ & 128 & 1.98 & 1.13 & \$1045 - \$3484 & 328.8 \\
    LRT-BERT$_{BASE}$ & 64 & 2.50 & 0.90 & \$831 - \$2768 & 261.2 \\
    LRT-BERT$_{BASE}$ & 32 & 2.63 & 0.85 & \$789 - \$2629 & 248.1 \\
    \bottomrule
    \end{tabular}
}
\caption[Cost efficiency of applying Low-Rank Transformer to the pre-training phase of BERT$_{BASE}$ model]{Cost efficiency of applying Low-Rank Transformer to the pre-training phase of BERT$_{BASE}$ model}
\label{tab:efficiency_bert}
\end{table}

\section{Conclusion}

In this chapter, we explore two Greenformers models which apply low-rank approximation to the Transformer model, i.e, Low-Rank Transformer (LRT) and Linformer. We conduct a thorough analysis to assess the efficiency of LRT and Linformer variants compared to the Transformer model. Based on our analysis, we figure out that LRT can be a better alternative to the Transformer model when the sequence length is short ($\leq 512$), while Linformer can be a better alternative when the sequence length is long ($\geq 512$). Linformer efficiency depends on the variation of input size and it is much more efficient when the input size is static. In terms of the number of parameters, we show that LRT is more suitable for on-device applications as it can reduce the storage requirements and the network cost significantly. Additionally, we show that both LRT and Linformer models can perform as well as the Transformer model. Lastly, we show that by applying LRT efficiency methods, we can significantly decrease the cost of training a transformer-based model, BERT. These results demonstrate the importance of applying Greenformers which can reduce the computational, economical, and environmental costs of developing the model without sacrificing the performance of the model in the downstream task.

\newpage

%% file: chapter/sec-3-low-rank-transformer.tex
\chapter{Low-Rank Transformer for Automatic Speech Recognition}
\label{sec-lrt}

Traditional HMM-based models have been outperformed by end-to-end automatic speech recognition (ASR) models. End-to-end ASR models allow a simpler end-to-end learning mechanism as it provides only a single model structure for replacing learning acoustic, pronunciation, and language models. Furthermore, end-to-end ASR models require only paired acoustic and text data, without additional prior knowledge, such as phoneme set and word dictionary. End-to-end attention-based recurrent neural network (RNN) models such as LAS~\cite{chan2016listen}, joint CTC-attention model~\cite{kim2017joint}, and attention-based seq2seq model~\cite{tuske2020attention_lstm} have significantly outperformed the traditional approaches by large margins in multiple speech recognition datasets. In recent years, fully-attentional transformer speech recognition models~\cite{dong2018speech,li2019speechtransformer} have further improved the attention-based RNN models in terms of performance and training speed. Transformer models reduce the training time of attention-based RNN models as it enables parallel computation along the time dimension. However, the total computational cost of both model structures remains high, especially in the streaming scenario in which the data is fed periodically, which makes the existing approach impractical for on-device deployment.

In this chapter, we extend the low-rank transformer (LRT)~\cite{winata2020lrt} to develop a compact and more generalized model for speech recognition tasks. We show that our LRT model can significantly reduce the number of parameters and the total computational cost of the model. Additionally, our experiment result suggests that our factorization can help the model to generalize better which enables the model to achieve a slightly higher evaluation score compared to the non-factorized model.

\section{Methodology}

\begin{figure}[ht]
  \centering
  \includegraphics[width=0.9\linewidth]{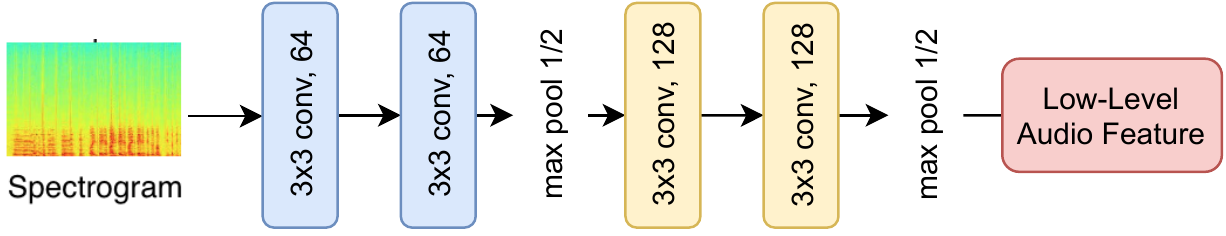}
  \caption[Configuration of our VGGish network]{Configuration of our VGGish network}
  \label{fig:vgg}
\end{figure}

We extend the Low-Rank Transformer (LRT) model~\cite{winata2020lrt} to enable processing audio data for speech recognition tasks. Our LRT variant accepts audio data in form of spectrograms of the power spectral density as its input and produces a sequence of characters as its output. The spectrogram $S \in \mathbb{R}^{f \times t}$ is computed from the magnitude squared of the short-time Fourier-transform (STFT) output, where $f$ denotes the length of the Fourier vector and $t$ denotes the length of time segments. We extend LRT by adding a specific module for processing audio information before feeding the audio information into the LRT encoder layer. Specifically, a small VGGish network~\cite{simonyan2014very} is incorporated to extract low-level audio features from the spectrogram input. Our VGGish network consists of 4 convolution layers with 2 max-pooling layers and produces an output tensor $I \in \mathbb{R}^{\frac{f}{4} \times \frac{t}{4} \times c}$, where c represents the number of output channels and in our configuration $c=128$. We then merge the frequency dimension $\frac{f}{4}$ and the channel dimension $c$ producing from the the tensor $I$ producing a matrix $X \in \mathbb{R}^{t \times \frac{cf}{4}}$ and feed the matrix $X$ into the LRT encoder layers. The illustration of the VGGish network is shown in Figure~\ref{fig:vgg}.

\begin{figure}[t!]
  \centering
  \includegraphics[width=.6\linewidth]{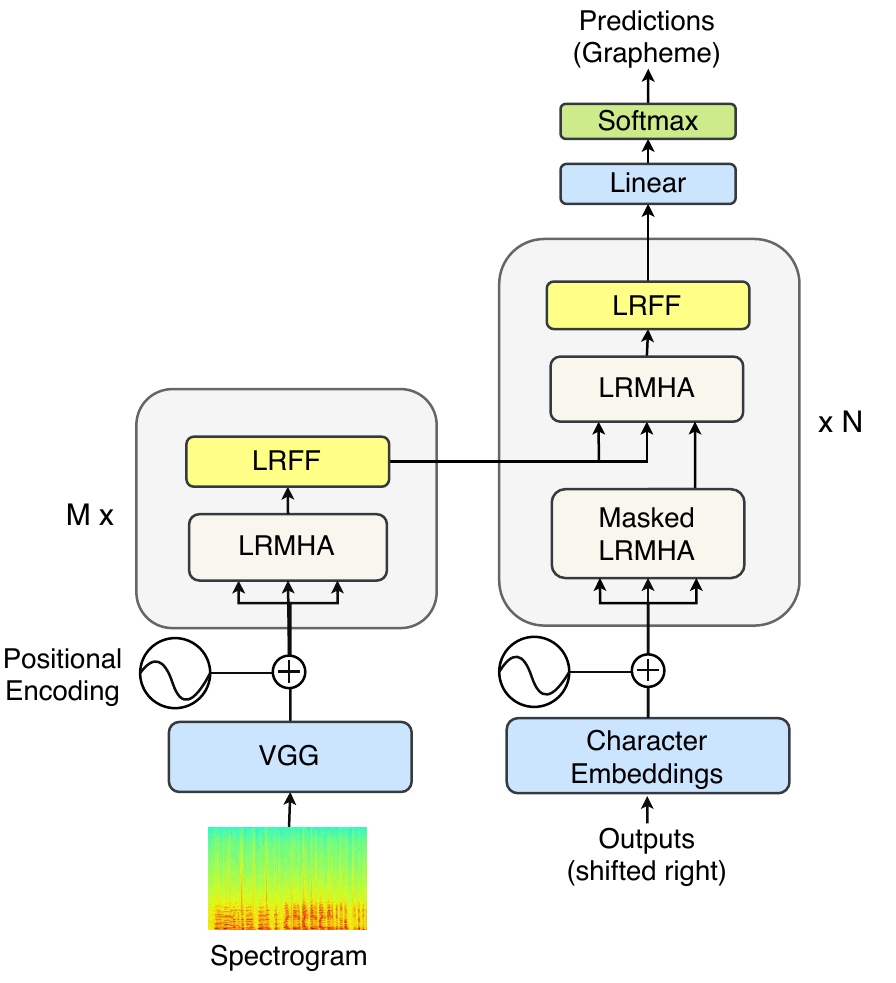}  
  \caption[Low-Rank Transformer Architecture]{Low-Rank Transformer Architecture. Low-Rank Transformer consists of $N$ layers of low-rank transformer encoder and $M$ layers of low-rank transformer decoder.~\cite{winata2020lrt}}
  \label{fig:transformer-asr}
\end{figure}

To generate the character output, we create a vocabulary of characters consisting of all the characters in the dataset and add a projection layer after the output layer of the LRT decoder layers. Specifically, we add an output projection $W \in \mathbb{R}^{d_{model} \times d_{vocab}}$ followed by a softmax layer to calculate the probability of the output grapheme at a specified time-step. The depiction of our speech recognition LRT model is shown in Figure~\ref{fig:transformer-asr}.

\section{Experimental Setup}

\subsection{Dataset}

We conduct our experiments on two speech recognition datasets: 1) a multi-accent Mandarin speech dataset, AiShell-1~\cite{bu2017aishell1}, and 2) a conversational telephone speech recognition dataset, HKUST~\cite{liu2006hkust}. We initialize our vocabulary with all characters in each corpus such that there is no out-of-vocabulary (OOV) and we add three additional special tokens: $<$PAD$>$, $<$SOS$>$, and $<$EOS$>$ for training and inference purposes. We preprocess the raw audio data into spectrograms with a hop length of 20ms, a window size of 10ms, and a length of FFT vector of 320. Table~\ref{tab:asr-dataset} shows the duration statistics of the datasets.

\begin{table*}[ht!]
\centering
\centering
\resizebox{0.4\textwidth}{!}{
\begin{tabular}{l|c|c}
\toprule
\textbf{Dataset} & \textbf{Split} & \textbf{Durations (h)} \\ 
\midrule
\multirow{3}{*}{AiShell-1} & Train & 150 \\ 
 & Valid & 10 \\ 
 & Test & 5 \\ \midrule
\multirow{3}{*}{HKUST} & Train & 152 \\ 
 & Valid & 4.2 \\
 & Test & 5 \\ 
\bottomrule
\end{tabular}
}
\caption[Duration statistics for AiShell-1 and HKUST datasets]{Duration statistics for AiShell-1 and HKUST datasets}
\label{tab:asr-dataset}
\end{table*}

\subsection{Hyperparameters}

To analyze the performance and efficiency trade-off over different rank $r$, we conduct experiment on our LRT model with three different settings for rank $r$, i.e, $r=100$, $r=75$ and $r=50$. For all experimental settings, we use VGG network with 6 convolutional layers, two LRT encoder layers, and four LRT decoder layers, yielding three different models: LRT($r=100$) with ~12M parameters, LRT($r=75$) with ~10M parameters, and LRT($r=50$) with ~8M parameters.

\subsection{Baselines}

For comparison with our LRT models, we develop three transformer models with two transformer encoder layers and four transformer decoder layers. We use different $dim_{model}$ and $dim_{inner}$ settings for the three models, producing Transformer (large) with ~23M parameters, Transformer (medium) with ~12M parameters, and Transformers (small) with ~8M parameters. In terms of model size, our LRT($r=100$) model is comparable to the Transformer (medium) model and LRT($r=50$) model is comparable to the Transformer (small) model. For additional comparison, we provide results gathered from the previous works for each dataset.

\subsection{Training and Evaluation}

In the training phase, we use the cross-entropy loss as the objective function. We optimize all models by using Adam optimizer~\cite{kingma2015adam}. In the evaluation phase, we generate the sequence with beam-search by selecting the best sub-sequence scored using the probability of the sentence $P(Y)$. The probability of the sentence $P(Y)$ is calculated with the following equation:

\begin{equation}
P(Y) = \alpha P_{trans}(Y|X) + \gamma \sqrt{wc(Y)},
\end{equation}

\noindent where $\alpha$ is the parameter to control the decoding probability from the decoder $P_{trans}(Y|X)$, and $\gamma$ is the parameter to control the effect of the word count $wc(Y)$. We use $\alpha = 1$, $\gamma = 0.1$, and a beam size of 8. We evaluate our model by calculating the character error rate (CER). We run our experiment on a single GeForce GTX 1080Ti GPU and Intel Xeon E5-2620 v4 CPU cores. 

\begin{table*}[!t]
\centering
\begin{minipage}{0.47\textwidth}
\centering
\resizebox{1.0\textwidth}{!}{
\begin{tabular}{lcc}
\toprule
\multicolumn{1}{l}{\textbf{Model}} & \multicolumn{1}{l}{\textbf{Params}} & \textbf{CER} \\ \midrule
\multicolumn{3}{c}{\textit{Hybrid approach}} \\ \midrule
\multicolumn{1}{l}{TDNN-HMM~\cite{bu2017aishell1}*} & - & 8.5\% \\ \midrule
\multicolumn{3}{c}{\textit{End-to-end approach}} \\ \midrule
\multicolumn{1}{l}{Attention Model~\cite{li2019end}*} & - & 23.2\% \\ 
\multicolumn{1}{l}{\hspace{7.4mm} + RNNLM~\cite{li2019end}*} & - & 22.0\% \\ 
\multicolumn{1}{l}{CTC~\cite{li2019framewise}*} & $\approx$11.7M & 19.43\% \\
\multicolumn{1}{l}{Framewise-RNN~\cite{li2019framewise}*} & $\approx$17.1M & 19.38\% \\ 
\multicolumn{1}{l}{ACS + RNNLM*~\cite{li2019end}} & $\approx$14.6M & 18.7\% \\ \midrule
\multicolumn{1}{l}{Transformer (large)} & 25.1M & 13.49\% \\
\multicolumn{1}{l}{Transformer (medium)} & 12.7M & 14.47\% \\
\multicolumn{1}{l}{Transformer (small)} & 8.7M & 15.66\%\\ \midrule
\multicolumn{1}{l}{LRT ($r=100$)} & 12.7M & \textbf{13.09\%} \\
\multicolumn{1}{l}{LRT ($r=75$)} & 10.7M & 13.23\% \\
\multicolumn{1}{l}{LRT ($r=50$)} & 8.7M & 13.60\%\\ \bottomrule
\end{tabular}
}
\end{minipage}
\begin{minipage}{0.52\textwidth}
\centering
\resizebox{0.99\textwidth}{!}{
\begin{tabular}{lcc}
\toprule
\multicolumn{1}{l}{\textbf{Model}} & \multicolumn{1}{l}{\textbf{Params}} & \textbf{CER} \\ \midrule
\multicolumn{3}{c}{\textit{Hybrid approach}} \\ \midrule
\multicolumn{1}{l}{DNN-hybrid~\cite{hori2017advances}*} & - & 35.9\% \\ 
\multicolumn{1}{l}{LSTM-hybrid (with perturb.)~\cite{hori2017advances}*} & - & 33.5\% \\ \midrule
\multirow{2}{*}{\begin{tabular}[c]{@{}l@{}}TDNN-hybrid, lattice-free MMI\\ (with perturb.)~\cite{hori2017advances}*\end{tabular}} & \multirow{2}{*}{-} & \multirow{2}{*}{28.2\%} \\ 
& & \\ \midrule
\multicolumn{3}{c}{\textit{End-to-end approach}} \\ \midrule
\multicolumn{1}{l}{Attention Model~\cite{hori2017advances}*} & - & 37.8\% \\
\multicolumn{1}{l}{CTC + LM~\cite{miao2016empirical}*} & $\approx$12.7M & 34.8\% \\
\multicolumn{1}{l}{MTL + joint dec. (one-pass)~\cite{hori2017advances}*} & $\approx$9.6M & 33.9\% \\ 
\multicolumn{1}{l}{\hspace{7.4mm} + RNNLM (joint train)~\cite{hori2017advances}*} & $\approx$16.1M & 32.1\% \\ \midrule
\multicolumn{1}{l}{Transformer (large)} & 22M & 29.21\% \\
\multicolumn{1}{l}{Transformer (medium)} & 11.5M & 29.73\% \\
\multicolumn{1}{l}{Transformer (small)} & 7.8M & 31.30\% \\ \midrule
\multicolumn{1}{l}{LRT ($r=100$)} & 11.5M & \textbf{28.95\%} \\
\multicolumn{1}{l}{LRT ($r=75$)} & 9.7M & 29.08\% \\
\multicolumn{1}{l}{LRT ($r=50$)} & 7.8M & 30.74\%\\ \bottomrule
\end{tabular}
}
\end{minipage}
\caption[Results on AiShell-1 and HKUST test sets]{Results on AiShell-1 (left) and HKUST (right) test sets.* The evaluation score is retrieved from the previous studies. We approximate the number of parameters based on the description in the previous studies.}
\label{results}
\end{table*}

\section{Result and Discussion}

\subsection{Evaluation Performance}

\begin{figure}[!t]
\centering
\includegraphics[width=0.75\linewidth]{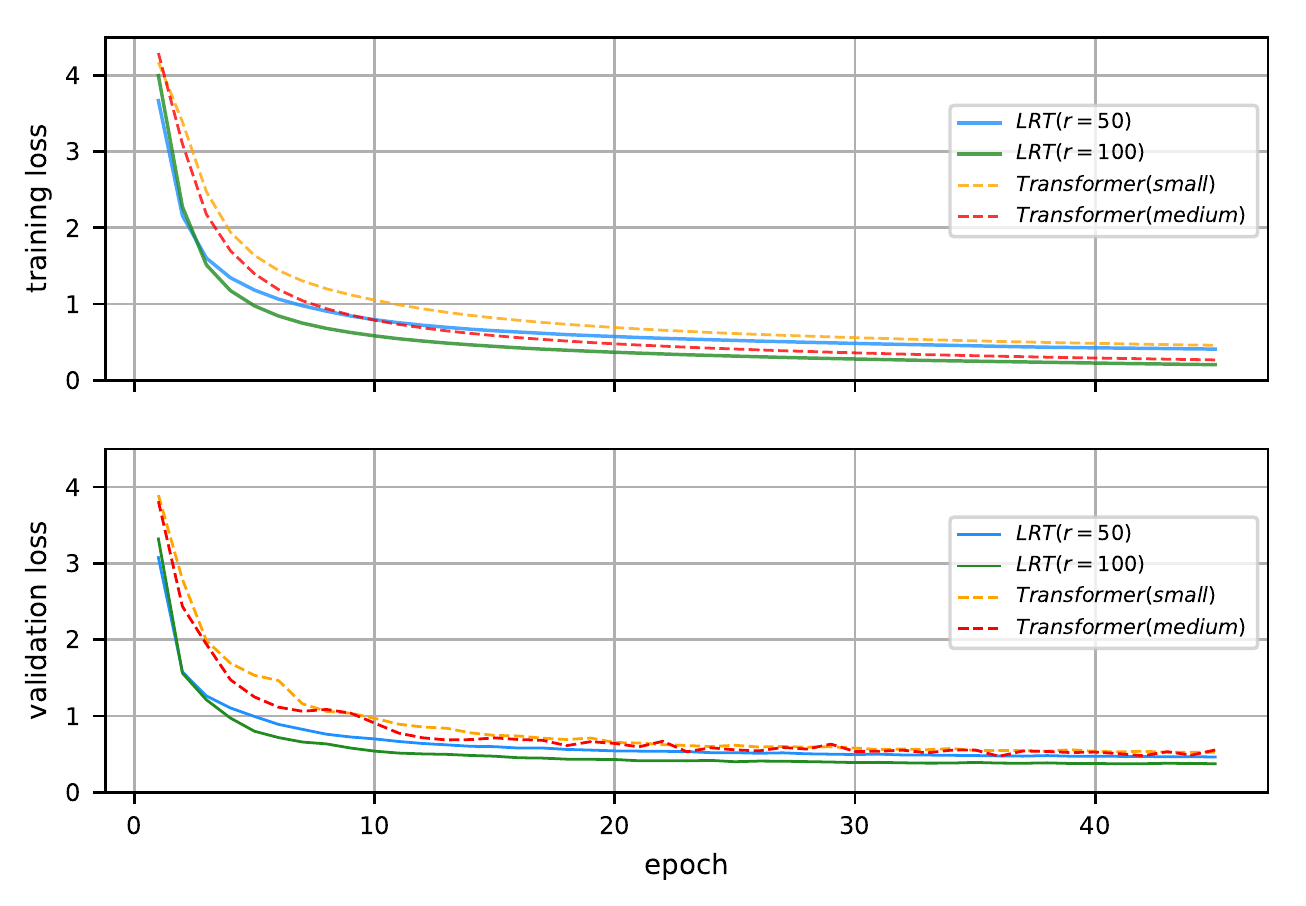}
\caption{Training and validation losses on AiShell-1 data.~\cite{winata2020lrt}}
\label{fig:loss-trend}
\end{figure}

As shown from the experiment results in Table~\ref{results}, LRT models gain some improvement compared to the original transformer model. For instance, LRT ($r=100$) model achieves better score compared to Transformer (large) model with only around $\sim$50\% of its parameters. LRT ($r=100$) model outperforms all original transformer models both in AiShell-1 and HKUST test sets, with a 13.09\% CER and a 28.95\% CER, respectively. Compared to TDNN-HMM, our LRT reduces the gap between the HMM-based hybrid and fully end-to-end approaches without leveraging any data perturbation strategy or external language model. Interestingly, as shown in Figure~\ref{fig:loss-trend}, LRT models achieve lower validation loss compared to the Transformer (large) model, which suggests that LRT models regularize better compared to the original transformer model.

\begin{table}[!t]
\centering
\caption[Compression rate and inference speed-up of LRT vs. Transformer (large)]{Compression rate and inference speed-up of LRT models compared to the Transformer (large) baseline. $\Delta$CER and $|\bar{X}|$ denote the improvement, and the mean length of generated sequences.}
\resizebox{0.85\textwidth}{!}{
\begin{tabular}{lccccccc}
\toprule
\multicolumn{1}{c}{\multirow{2}{*}{\textbf{dataset}}} & \multirow{2}{*}{\textbf{r}} & \multirow{2}{*}{\textbf{$\Delta$CER}} & \multicolumn{1}{c}{\multirow{2}{*}{\textbf{compress.}}} & \multicolumn{2}{c}{\textbf{speed-up}} & \multicolumn{1}{c}{\multirow{2}{*}{\textbf{$|\bar{X}|$}}} \\ \cmidrule{5-6}
\multicolumn{1}{c}{} & & & & \multicolumn{1}{c}{\textbf{GPU}} & \textbf{CPU only} & \\ \midrule
AiShell-1 & baseline & 0 & 0 & 1 & 1 & 23.08 \\
 & 100 & 0.40\% & 49.40\% &  1.17x & 1.15x & 23.15 \\
 & 75 & 0.26\% & 57.37\% & 1.23x & 1.16x & 23.17 \\
 & 50 & -0.11\% & 65.34\% & 1.30x & 1.23x & 23.19 \\ \midrule
HKUST & baseline & 0 & 0 & 1 & 1 & 22.43 \\
 & 100 & 0.26\% & 47.72\% & 1.21x & 1.14x & 22.32 \\
 & 75 & 0.13\% & 55.90\% & 1.26x & 1.15x & 22.15 \\
 & 50 & -1.53\% & 64.54\% & 1.35x & 1.22x & 22.49 \\ \bottomrule
\end{tabular}
}
\label{tab:efficiency}
\end{table}

\subsection{Space and Time Efficiency}

In terms of space efficiency, as shown in Table~\ref{results}, in the AiShell-1 test set, our LRT ($r=50$) model achieves similar performance to the Transformer (large) model with only around $\sim$35\% of the Transformer (large) model parameters. While in the HKUST test set, our LRT ($r=75$) model achieves similar performance to the Transformer (large) with only $\sim$45\% of its parameters. These facts suggest that low-rank approximation can halve the space requirement of transformer models without hurting their performance. In terms of time efficiency, as shown in Table \ref{tab:efficiency}, LRT ($r=75$) models gain generation time speed-up by 1.23x-1.26x in the GPU and 1.15x-1.16x in the CPU compared to the Transformer (large) baseline model. While LRT ($r=50$) model gain generation time speed-up by 1.30x-1.35x in the GPU and 1.22x-1.23x in the CPU. These suggest that low-rank approximation can boost the generation speed of the transformer model by around $\sim$1.25x in the GPU and $\sim$1.15x in the CPU without hurting its performance and even lower rank can be applied to further increase the speed-up ratio although it might degrade the overall performance of the model.

\section{Conclusion}

In this chapter, we show the application of the low-rank transformer (LRT) model on automatic speech recognition tasks. LRT is a memory-efficient and fast neural architecture that compress the network parameters and boosts the speed of the inference time by up to 1.35x in the GPU and 1.23x in the CPU, as well as the speed of the training time for end-to-end speech recognition. Our LRT improves the performance even though the number of parameters is reduced by 50\% compared to the baseline transformer model. Similar to the previous result in the MNIST dataset, LRT could generalize better than uncompressed vanilla transformers and outperforms those from existing baselines on the AiShell-1 and HKUST datasets in an end-to-end setting without using additional external data.

\newpage

%% file: chapter/sec-5-linformer-for-genomics.tex
\chapter{Linformer for Alzheimer's Disease Risk Prediction}
\label{sec-linear-attention}

Recent development in genome modeling shows that deep learning models, including attention-based variant, could achieve pretty impressive performance on multiple haploid regulatory elements prediction tasks in genomics outperforming the other models by a large margin \cite{zhou2015dsea,zaheer2020bigbird,ji2020dnabert,umarov2017cnnprom,umarov2019promidd,min2017denhancer,yin2019dhistone}. Extending this approach for phenotype prediction, such as disease risk prediction, might provide huge advantages from improving the prediction accuracy and creating a new set of analytical tools for better understanding the genomics interaction of a certain phenotype characteristic. In this chapter, we explore the possibility of applying the Linformer model on Alzheimer's disease risk prediction tasks using genomics sequences. Alzheimer's disease is one of the most devastating brain disorders in the elderly. It is estimated that nearly 36 million are affected by Alzheimer's disease globally and the number is expected to reach 115 million by 2050 \cite{huynh2017ad}. Disease risk prediction for Alzheimer's disease can help to prolong the health span of an individual with potential Alzheimer's disease trait and enable research to get a better understanding of the disease progression from a very early stage.

Compared to haploid regulatory elements prediction, genomic modeling for disease risk prediction requires the model to capture a much larger scale of the input sequence. In regulatory elements prediction, regions are usually short and indicated by a certain pattern that is located inside regions. For instance, promoter is about 100-1000 nts~\cite{karni2007promoter} long and is associated with core promoter motifs such as TATA-box and CAAT-box, and GC-box~\cite{blackwood1998enhancer,lundin1994gcbox}. While enhancer is about 50 - 1,500nts long and is associated with GATA-box and E-box~\cite{blackwood1998enhancer,anderson1998gata,ogilvy2007gata}. On the other hand, disease risk prediction requires the model to capture the aberration within at less than a single gene and the average gene length inside the human genome is around 24,000 base pairs long \cite{fuchs2014gene-length}. This sequence will be even longer when we would like to consider neighboring regulatory elements such as promoter, and the remote regulatory elements such as enhancer and silencer, which could be located up to one million base pairs away from the corresponding gene\cite{pennacchio2013enhancer}. Moreover, most animals genomes, including humans, are diploid. This will not only increase the size of the data but also requires additional methods to encode and process the diploid information as the haploid sequences (haplotypes) of a diploid chromosome is not directly observable.

In this chapter, we tackle the two problems mentioned above and develop the first-ever disease risk prediction model for Alzheimer's disease from genome sequence data. To tackle the long input sequence, we propose two solutions: 1) we apply Linformer~\cite{wang2020linformer}, a variant of transformer model with linear attention, to enable processing long sequences, and 2) we extend SentencePiece tokenization to aggregate multiple nucleotides into a single token for different diploid sequence representations. We show that SentencePiece tokenization can reduce the computational cost-effectively by producing a set of tokens constructed from several base pairs. We thus employ the Linformer model with a low-rank self-attention mechanism which allows the model to reduce the memory bottleneck from $O(n^2)$ to $O(n)$ space complexity which significantly increases the capacity of the model to process longer input sequence. These two solutions allow the model to capture $\sim$35$\times$ longer sequence compared to the prior works from $\sim$1,000nts to $\sim$35,0000nts long.

For processing the diploid chromosome, we introduce a diplotype representation, where we encode a diploid chromosome sequence into 66 diplotype tokens comprising combination of 11 tokens which represents 4 nucleotide tokens (\texttt{A}, \texttt{G}, \texttt{T}, and \texttt{C}), unspecified token (\texttt{N}), and insertion-deletion tokens (\texttt{AI}, \texttt{GI}, \texttt{TI}, \texttt{CI}, \texttt{NI}, and \texttt{DEL}). Our experiment result suggests that linear attention-mechanism with diplotype token representation can effectively model long diplotype for disease prediction and can produce a higher score compared to the Food and Drug Administration (FDA) approved feature set for predicting Alzheimer's disease. We further analyze that our model can capture important single nucleotide polymorphisms (SNPs) which are located inside the gene and the regulatory region. Our analysis approach can be further utilized to provide an explainable analysis of the linear attention-based model.

\section{Preliminaries}

\begin{figure}[!t]
    \centering
    \resizebox{1.0\textwidth}{!}{
        \includegraphics[scale=1.0]{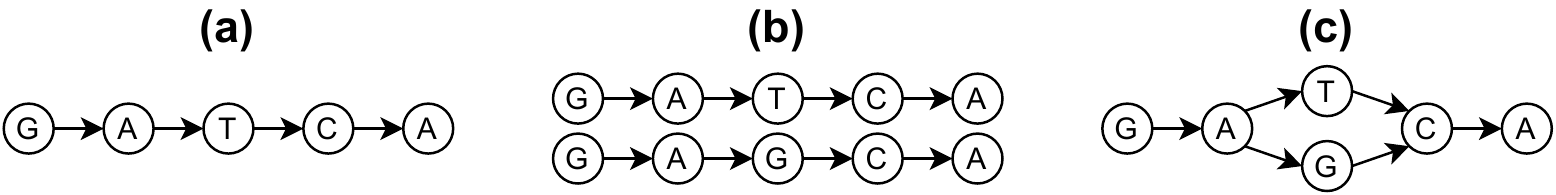}
    }
    \caption[DNA sequence representation for haploid and diploid chromosome]{DNA sequence representation for haploid and diploid chromosome. (a) Haplotype representation of a haploid chromosome (b) Two haplotypes  representation of a diploid chromosome. (c) Diplotype representation of a diploid chromosome.}
    \label{fig:sequence_representation}
\end{figure}

A human genome consists of 23 pairs of chromosomes, each chromosome comprises tens to hundreds of millions of nucleotides long sequence. There are four different kind of nucleotides: adenine (\texttt{A}), guanine (\texttt{G}), thymine (\texttt{T}), and cytosine (\texttt{C}) each connected to its reverse complement, i.e., \texttt{A} with \texttt{T} and \texttt{C} with \texttt{G}, which construct a deoxyribonucleic acid (DNA). Some regions inside a chromosome encode the template for creating a protein sequence. These regions are called genes. As human chromosomes are diploid (paired), there are also two copies of the same gene inside a human body. Each copy of a gene is called an allele. 

In 1911, the terms ``phenotype'' and ``genotype'' are introduced\cite{johannsen1911genotype}. Phenotype refers to the physical or observable characteristics of an individual , while genotype refers to a genetic characteristic of an individual. As the definition of genotype is rather broad and fluid depending on the context, in this work, we use the term diploid genotype (diplotype) as the sequence of nucleotides from a paired chromosome and haploid genotype (haplotype) as the sequence of nucleotides from a single chromosome. Specifically, we define a haplotype as a sequence $H = (h_1, h_2, ..., h_n)$, where $h_i$ denotes the nucleotide corresponding to the position $i$ and a diplotype as a sequence $D = (d_1, d_2, ..., xd_n)$, where $d_i$ denotes the pair-nucleotide corresponding to the position $i$.

To define a paired (diploid) chromosomes, we can represent the diploid chromosome as either: 1) as a set of two haplotypes $S = {H_1, H_2}$ where $H1$ and $H2$ are the two haplotypes, one for each chromosome; or 2) as a diplotype which can be visualized as an assembly graph $G = (B, E)$, where $B$ is a set of nucleotide, and $E$ denotes the edge connecting each nucleotide. 
Intuitively, for a diploid chromosome, two haplotypes representation would be the most biologically plausible way to representing a diploid chromosome sequence as it matches the physical representation of the diploid chromosome. Unfortunately, due to the nature of primer design in sequencing \cite{pham2021aspcr-hlac,darawi2013aspcr-ad}, these haplotypes cannot be directly observed with a standard sequencing technique.
Figure~\ref{fig:sequence_representation} shows the example of each sequence representation.

\section{Methodology}

Aside from using the Linformer model to improve the transformer's efficiency, the choice of the sequence representation and tokenization also play an important role. In the following section, we describe how we represent a diploid sequence as a diplotype and elucidate the subword tokenization method which can further improve the efficiency of the model.

\subsection{Sequence Representation}

\begin{figure}[!t]
    \centering
    \resizebox{0.9\textwidth}{!}{
        \includegraphics[scale=1.0]{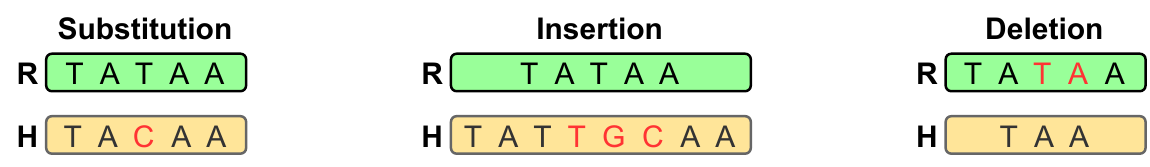}
    }
    \caption[Mutation patterns in genomic sequence]{Mutation patterns in genomic sequence. R denotes a reference sequence and H denotes a haplotype compared to the reference.}
    \label{fig:mutation_pattern}
\end{figure}

In this work, we represent our diploid sequence with diplotype representation as shown in Figure~\ref{fig:sequence_representation} (c). In diplotype representation, we represent each position as a diplotype token. A diplotype token represents the combination of haplotype tokens including mutation patterns. As shown in Figure \ref{fig:mutation_pattern}, there are three mutation patterns that can occur on a genomic sequence, i.e., substitution, insertion, and deletion. A substitution mutation is commonly called as Single Nucleotide Polymorphism (SNP), while insertion and deletion are commonly called as \textbf{indel}. To accommodate all possible mutations, the diplotype token vocabulary is extended from the haplotype vocabulary in three steps. First, insertion and deletion patterns are introduced by extending the five tokens in the haplotype vocabulary (\texttt{A}, \texttt{T}, \texttt{C}, \texttt{G}, \texttt{N}) to 11 tokens (\texttt{A}, \texttt{T}, \texttt{C}, \texttt{G}, \texttt{N}, \texttt{DEL}, \texttt{AI}, \texttt{TI}, \texttt{CI}, \texttt{GI}, \texttt{NI}). \texttt{DEL} denotes deletion and \texttt{AI}, \texttt{TI}, \texttt{CI}, \texttt{GI}, and \texttt{NI} denote insertion that comes with the corresponding \texttt{A}, \texttt{T}, \texttt{C}, \texttt{G}, \texttt{N} tokens. Second, a combination of the 11 tokens are added, e.g., \texttt{A-T}, \texttt{C-A}, \texttt{G-T}, \texttt{A-TI}, \texttt{C-AI}, and \texttt{C-TI}; producing 66 tokens in total. Third, the diplotype tokens are mapped into a single character code to allow tokenization over a sequence of diplotype tokens. The complete map of diplotype tokens vocabulary is shown in Table \ref{tab:diplotype-tokens}.

\begin{CJK*}{UTF8}{gbsn}

\begin{table*}[!t]
\centering
\resizebox{0.9\linewidth}{!}{
\begin{tabular}{lr|lr|lr|lr|lr|lr}
\toprule
\multicolumn{12}{c}{\textbf{Map of Diplotype Tokens}} \\ \midrule
\texttt{A} & \texttt{A} & \texttt{DEL-A} & 腌 & \texttt{A-C} & 嗄 & \texttt{AI-C} & 爸 & \texttt{C-G} & 嚓 & \texttt{CI-G} & 懂 \\
\texttt{C} & \texttt{C} & \texttt{DEL-AI} & 拔 & \texttt{A-G} & 阿 & \texttt{AI-CI} & 比 & \texttt{C-N} & 拆 & \texttt{CI-GI} & 答 \\
\texttt{G} & \texttt{G} & \texttt{DEL-C} & 吃 & \texttt{A-N} & 呵 & \texttt{AI-G} & 八 & \texttt{C-T} & 礤 & \texttt{CI-N} & 达 \\
\texttt{N} & \texttt{N} & \texttt{DEL-CI} & 搭 & \texttt{A-T} & 锕 & \texttt{AI-GI} & 霸 & \texttt{C-CI} & 车 & \texttt{CI-NI} & 第 \\
\texttt{T} & \texttt{T} & \texttt{DEL-G} & 想 & \texttt{A-AI} & 吖 & \texttt{AI-N} & 巴 & \texttt{C-GI} & 床 & \texttt{CI-T} & 瘩 \\
\texttt{AI} & \texttt{B} & \texttt{DEL-GI} & 香 & \texttt{A-CI} & 俺 & \texttt{AI-NI} & 逼 & \texttt{C-NI} & 穿 & \texttt{CI-TI} & 沓 \\
\texttt{CI} & \texttt{D} & \texttt{DEL-N} & 学 & \texttt{A-GI} & 安 & \texttt{AI-T} & 把 & \texttt{C-TI} & 出 & \texttt{G-GI} & 高 \\
\texttt{GI} & \texttt{H} & \texttt{DEL-NI} & 虚 & \texttt{A-NI} & 案 & \texttt{AI-TI} & 笔 & \texttt{GI-N} & \begin{CJK}{UTF8}{bkai}蝦\end{CJK} & \texttt{G-N} & 给 \\
\texttt{NI} & \texttt{O} & \texttt{DEL-T} & 徐 & \texttt{A-TI} & 按 & \texttt{N-NI} & 讷 & \texttt{GI-NI} & 合 & \texttt{G-NI} & 股 \\
\texttt{TI} & \texttt{U} & \texttt{DEL-TI} & 需 & \texttt{NI-T} & 喔 & \texttt{N-T} & 哪 & \texttt{GI-T} & 虾 & \texttt{G-T} & 个 \\
\texttt{DEL} & \texttt{X} & \texttt{T-TI} & 拓 & \texttt{NI-TI} & 侬 & \texttt{N-TI} & 娜 & \texttt{GI-TI} & 盒 & \texttt{G-TI} & 该 \\

\bottomrule
\end{tabular}
}
\caption{List of all diplotype tokens.} 
\label{tab:diplotype-tokens}
\end{table*}

\end{CJK*}

\subsection{Subword Tokenization}

Before feeding the sequence into the model, tokenization methods such as k-mer tokenization~\cite{min2017kmeremb,ji2020dnabert} and gapped k-mer tokenization~\cite{ghandi2014gkmer,shrikumar2019gkmexplain} is commonly incorporated to enrich the token representation. In this work, we incorporate a subword tokenization method called SentencePiece\cite{kudo2018sentencepiece}. SentencePiece is a subword tokenization method that doesn't require any language-specific prior. SentencePiece enables the application of subword tokenization beyond the NLP field. Internally, SentencePiece uses a slightly modified version of either Byte-pair-encoding (BPE), WordPiece, or Unigram subword tokenizations to generate the vocabulary and tokenize the sequence. Aside from the capability provided by the internal subword tokenization module, SentencePiece provides two additional pre-processing phases: 1) character normalization and 2) white-space preservation. In the character normalization phase, SentencePiece transforms a Unicode character into its semantically equivalent decomposed-Unicode character following a certain normalization standard. In white-space preservation phase, SentencePiece escapes the \texttt{<space>} character into a meta symbol \_ (U+2581). The escaped sequence is then used as the input to the internal subword tokenization method for further processing. For our experiment, we utilize SentencePiece with NFCK\footnote{https://unicode.org/reports/tr15/} normalization standard and BPE tokenization for performing the subword tokenization.
 
\begin{figure}[!t]
    \centering
    \resizebox{0.7\textwidth}{!}{
        \includegraphics[scale=1.0]{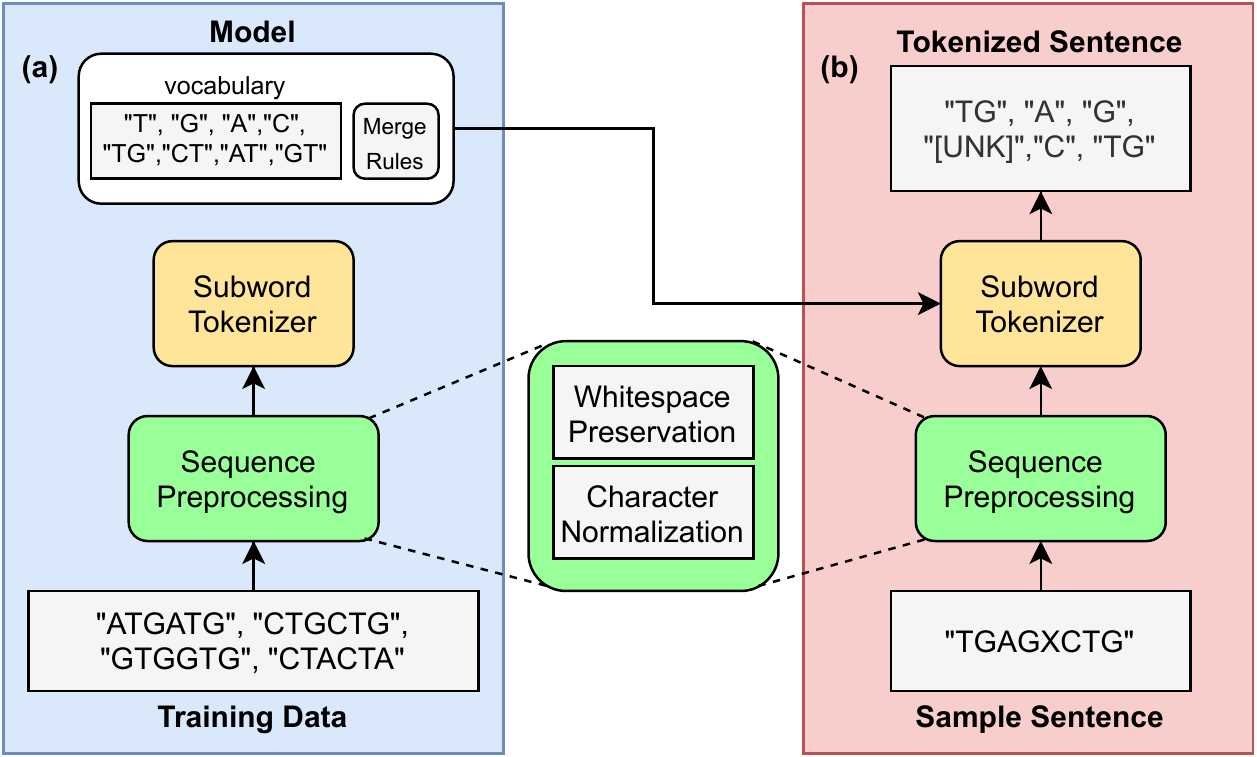}
    }
    \caption[SentencePiece with BPE subword tokenization]{SentencePiece with BPE subword tokenization. (a) Training phase. (b) Inference phase. The vocabulary is ordered by the frequency in descending order.}
    \label{fig:subword-tokenization}
\end{figure}

SentencePiece tokenization consists of two phases: training and inference. During the training phase, given training data $\mathcal{D}$, SentencePiece tokenization builds a vocabulary with a dynamic length token representation. As SentencePiece supports several subword tokenization algorithms, the vocabulary generation process is slightly different from one to another. For BPE tokenization, the vocabulary is built bottom-up starting from single-length character and iteratively add longer character pairs based on the most-frequent pairs in $\mathcal{D}$. For WordPiece tokenization, the vocabulary is also built-in bottom-up manner. But different from BPE tokenization, WordPiece scores character pairs by their likelihood in $\mathcal{D}$ instead of its frequency. Unlike BPE and WordPiece, the Unigram method starts with a large combination of tokens and iteratively reduce the vocabulary size by reducing symbols with the least contribution to the loss of the Unigram language model build from $\mathcal{D}$ The training process of SentencePiece tokenization is stopped when it reaches a specified vocabulary size. From the training phase, SentencePiece tokenization produces a model file consisting of the vocabulary and the rules to tokenize a sequence. During the inference phase, given a sequence data $\mathcal{S}$, SentencePiece tokenizes $\mathcal{S}$ into a set of tokens $(t_1, t_2, ..., t_n)$ according to the tokenization rules. Figure \ref{fig:subword-tokenization} shows the sample of the training and inference phase for the SentencePiece with BPE subword tokenization.

\section{Experimental Setup}

We conduct our experiment for Alzheimer's disease risk prediction on the Chinese Cohort by comparing the effectiveness of the Linformer model with other baseline models. In the following sections, we describe the SentencePiece training of the diplotype tokens, the pre-training phase of the Linformer model, and the fine-tuning and evaluation of Linformer models.

\subsection{Dataset Construction}

We construct a dataset of genome sequences for predicting a specific disease called late-onset Alzheimer's disease (LOAD) \cite{rabinovici2019load} on Chinese Cohort. We collect the sequencing data from 678 Hong Kong elderly with a minimum age of 65. The subjects are diagnosed with any dementia-related issue through Montreal Cognitive Assessment (MoCA) test \cite{nasreddine2005moca}. The score is adjusted according to the demography information, i.e., age, gender, and education year of the subject, and the final diagnosis is made by medical professionals to decide whether the subject has cognitive impairment. From the 678 subjects, we filter out subjects with mild cognitive impairment (MCI) and other types of dementia that are not related to Alzheimer's Disease. After the filtering, we end up with 626 subjects with 386 subjects labeled as Alzheimer's disease carrier (AD) and 240 subjects labeled as non-carrier (NC). It is also worth noting that AD diagnosis through memory test is not always accurate \cite{cecato2011mocaperf} and currently, a definite AD diagnosis can only be possible through post-mortem diagnosis \cite{deture2019neuropathological}. For each subject, we collect genome sequence information from around the APOE region~\cite{zhou2019apoe} located in chromosome 19 for each subject. The sequencing is done by using the Ilumina platform with a read depth of 5 and a read length of 150. We align the sequence data with bamtools~\cite{barnett2011bamtools} using hg19 reference genome~\cite{church2011hg19} to align the sequence data. For generating diplotype, we use pysam v0.15.4 and samtools v1.7 \cite{li2009samtools} to generate the profile for each position from the aligned read data and taking top-2 values from the profile to form the diplotype.

\subsubsection{SentencePiece Training}

For diplotype SentencePiece tokenizer, we build a training corpus generated from the GrCh37 (hg19) human reference genome\footnote{\url{https://hgdownload.soe.ucsc.edu/goldenPath/hg19/bigZips/hg19.fa.gz}} and the dbSNP154\footnote{\url{https://ftp.ncbi.nih.gov/snp/redesign/archive/b154}}. We use SNPs which are labeled \texttt{COMMON}\footnote{A \texttt{COMMON} SNP is defined as an SNP that has at least one 1000Genomes population with a minor allele of frequency (MAF) >= 1\% and for which 2 or more founders contribute to that minor allele frequency.} in the dbSNP154 to introduce mutation patterns into the training corpus. We fill the remaining position with the sequence from GrCh37 (hg19) human reference genome to generate a complete diplotype of the chromosome sequence. As our fine-tuning only focuses on the APOE region, we construct the training data for SentencePiece from only the chromosome 19 sequence. We convert the chromosome 19 sequence into a set of sentences $\mathcal{S}$. To build the sentences $S$, we first random sample $k=50$ starting points from range 0 to 5,000 from the 5' end. From each starting point, we iteratively take consecutive sequences with a random length from minimum length $a=500$ to maximum length $b=1000$ until the sequence reaches the end of the corresponding chromosome. All random samplings are taken uniformly. The sentence $S$ is then used as the training corpus for SentencePiece training. Specifically, we build a SentencePiece model for diplotype sequences with BPE tokenization to construct the vocabulary. We random sample 3,000,000 sentences to build a vocabulary of size 32000 with 5 and 66 base tokens for haplotype and diplotype vocabulary respectively. We add three additional special tokens ``[UNK]'', ``[CLS]'', and ``[SEP]'' to be used by the model in the pre-training and fine-tuning phases.

\subsubsection{Pre-Training Phase}

Following the success of the pre-training approach in many different fields such as NLP, computer vision, and speech processing, we perform pre-training to allow the model to learn the structure of a genome sequence. We use the same data as the one used for generating the SentencePiece models, but we concatenate several sentences into a single input sequence with a maximum sequence length of 4,096 tokens. Similar to language model pre-training in NLP, we use a masked language modeling (MLM) objective to train our model. For our model, we utilize a 12 layers Linformer model with 12 heads, hidden size of 768, feed-forward size of 3,072, drop out of 0.1, and low-rank factor of 256. We use AdamW optimizer~\cite{loshchilov2018decoupled} to optimize our model with initial learning rate of $1\mathrm{e}-4$ and apply a step-wise learning rate decay with $\gamma=0.999991$. We run the pre-training of our model for 200,000 steps with a batch size of 256.

\subsubsection{Fine-Tuning and Evaluation Details}

To evaluate the pre-trained models, we run a fine-tuning process to adapt the model for Alzheimer's disease risk prediction task. We feed the diplotype sequence and project the output from \texttt{[CLS]} token to get the prediction logits. We fine-tune the model for a maximum of 30 epochs with the early stopping of 5, the learning rate of $1\mathbb{e}-5$, and step-wise learning rate decay with $\gamma=0.9995$. We use cross-entropy loss and AdamW optimizer to optimize our model. We run the fine-tuning process 5 times with different stratify splitting for train, validation, and valid. We evaluate our models by calculating the AUC and F1-score from the model predictions and the phenotype label.

\subsubsection{Baseline Model}

For comparison with our Linformer model, we incorporate a baseline logistic regression model trained with APOE-$\epsilon3$ and APOE-$\epsilon4$ variants which correspond to rs7412 and rs429358 SNPs within the APOE region\cite{belloy2019apoe}. These features have been approved by the Food and Drug Administration (FDA)\footnote{\url{https://www.accessdata.fda.gov/cdrh_docs/reviews/K192073.pdf}} to be used as the measurand for risk of developing late-onset Alzheimer's disease (LOAD).

\section{Result and Discussion}

\begin{table*}[!t]
\centering
\centering
\resizebox{0.7\textwidth}{!}{
\begin{tabular}{lcc}
\toprule
\multicolumn{1}{l}{\textbf{Model}} & \textbf{AUC (\%)} & \textbf{F1-Score (\%)} \\ \midrule
\multicolumn{3}{c}{\textit{Baseline Model}} \\ \midrule
\multicolumn{1}{l}{FDA-approved feats.} & 58.88 & 49.24 \\ \midrule
\multicolumn{3}{c}{\textit{Linformer Model}} \\ \midrule
\multicolumn{1}{l}{Linformer (w/o pretraining)} & 56.49 & 56.83 \\
\multicolumn{1}{l}{Linformer (50k steps)} & 62.24 & 61.33 \\
\multicolumn{1}{l}{Linformer (100k steps)} & 62.94 & 61.17 \\
\multicolumn{1}{l}{Linformer (150k steps)} & 64.10 & 61.67 \\
\multicolumn{1}{l}{Linformer (200k steps)} & 64.35 & 61.57 \\ \bottomrule
\end{tabular}
}
\caption[Finetuning result on Alzheimer's disease prediction]{Finetuning result on Alzheimer's disease prediction. 100k and 200k denote the number of pre-training steps.}
\label{tab:results-linformer}
\end{table*}

\subsection{Evaluation Performance}

\begin{figure*}[t!]
    \centering
    \resizebox{1\textwidth}{!}{  
        \includegraphics{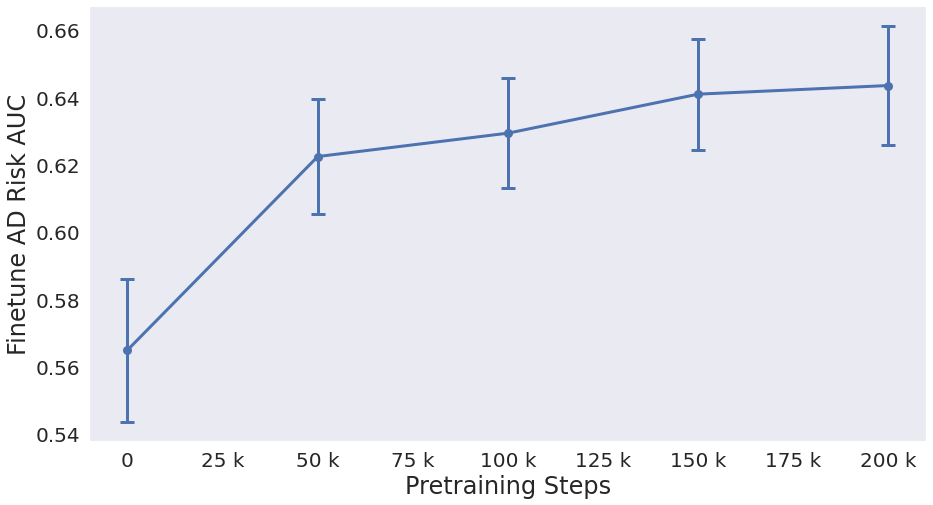}
        \includegraphics{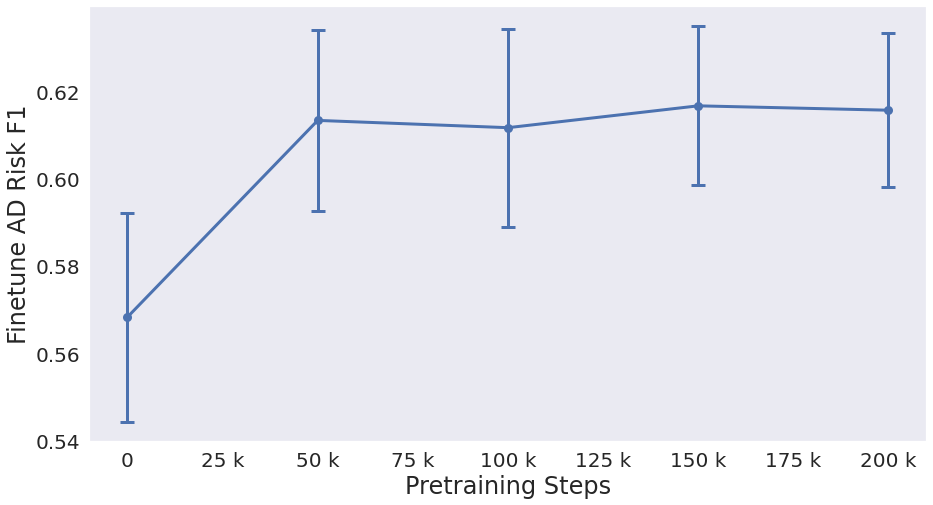}
    }
    \caption[Performance of Linformer model across different pretraining steps]{Performance of Linformer model across different pretraining steps}
    \label{fig:auc_f1_pretrain_step}
\end{figure*}

As shown from the experiment results in Table~\ref{tab:results-linformer}, the pre-trained Linformer model gains some improvement compared to the baseline model. The model without pretraining achieves a lower AUC score and a higher F1-score compared to the baseline model, this is probably due to the choice of probability threshold in the regression model which might overfit the validation data. Furthermore, we see a clear uptrend for longer training which suggests that understanding the pattern in the genomic sequence can provide benefit on the disease risk prediction task. Figure~\ref{fig:auc_f1_pretrain_step} shows the score improvement of the Linformer model corresponds to the pretraining steps.

\subsection{Effects of Sequence Length}

\begin{figure*}[t!]
    \centering
    \resizebox{1\textwidth}{!}{  
        \includegraphics{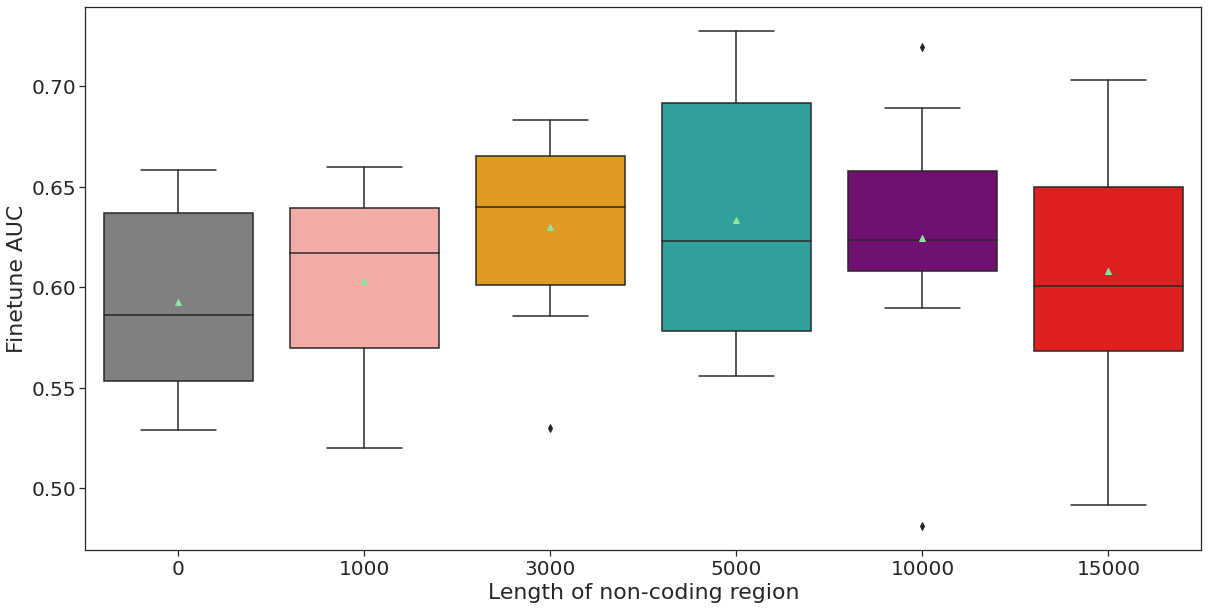}
    }
    \caption[Performance of Linformer (200k) model across different length of non-coding region.]{Performance of Linformer (200k) model across different length of non-coding region. 0 indicates the model is only trained on the APOE gene sequence without any non-coding region. Green triangle indicates the mean score over 5 runs.}
    \label{fig:auc_seq_len}
\end{figure*}

We conduct further analysis to compare the effect of adding different non-coding region size to the model performance. We first fine-tune the model with only the APOE gene sequence (3611nts), and gradually adding the upstream and downstream non-coding regions. We experiment with additional 1000, 3000, 5000, 10000, and 15000 sequence lengths from both upstream and downstream resulting in a sequence of length 5611nts, 9611nts, 13611nts, 23611nts, and 33611nts, respectively. As shown in Figure~\ref{fig:auc_seq_len}, the performance peaks when we provide an additional 5000 sequence length to the model, and it performs much worse when we only use the APOE gene or when we add even more non-coding regions. This result aligns with the top three promoters and enhancers with the highest GeneHancer score~\cite{fishilevich2017genehancer}, of which the three regions are located within the range of 5000nts upstream. This indicates our model can extract valid features from the sequence, although it is not robust against noises.

\subsection{Efficiency Contribution}

\begin{table*}[ht!]
\centering
\centering
\resizebox{0.8\textwidth}{!}{
\begin{tabular}{cccc}
\toprule
\textbf{Baseline} & \textbf{w/ Linformer} & \textbf{w/ Subword} & \textbf{w/ Linformer +} \\ 
& & \textbf{Tokenization} & \textbf{Subword Tokenization} \\
\midrule
1024 & 4096 & $\sim$9,000 & $\sim$35,000 \\ \bottomrule
\end{tabular}
}
\caption[Maximum sequence length with Linformer and Subword Tokenization]{Maximum sequence length with Linformer and Subword Tokenization.}
\label{tab:sequence-length-contrib}
\end{table*}

As shown in Table~\ref{tab:sequence-length-contrib}, with our approach we can encode a sequence by up to $\sim$35000nts with a 12-layer transformer model. This capability is achieved through two different methods: 1) subword tokenization and 2) Linformer model. Subword tokenization reduces the memory usage and computational cost by aggregating unitary tokens into a combination of tokens, based on our experiment, the average token length in our subword tokenizer is $\sim$8.7, which yields, on average, reduction the sequence length by a factor of $\sim$8.7 on a linear-attention mechanism. On the other hand, the Linformer model reduces the space and time complexity from O($n^2$) to O($n$) and based on our experiment it can process long sequence by $\sim$4$\times$ longer compared to the original with the same memory budget. This benefit will definitely increase if we compare it with a much larger sequence length. In this case, we can conclude that our approach can process 35$\times$ longer sequence by leveraging subword tokenization which reduces the sequence length by a factor of $\sim$8.7, and the Linformer model which increases the sequence length limit to $\sim$4$\times$ the original transformer model.


\section{Conclusion}

In this chapter, we show the application of the Linformer model on the Alzheimer's disease prediction task. Linformer is a memory-efficient and fast transformer variant architecture that reduce the space and time complexity of the transformer model from O($n^2$) to O($n$). Along with subword tokenization, we can increase the input sequence limit to $\sim$35000nts compared to $\sim$1000nts in most of the prior works. With additional pre-training, the Linformer model outperforms the existing baseline model with the pre-extracted feature using APOE-$\epsilon3$ and  APOE-$\epsilon4$ variants. Our analysis on extending non-coding regions suggests that our Linformer model can capture the required features in the regulatory regions correctly although the model is prone to noises. Lastly, we show that efficiency can also come from the choice of the tokenization approach which in this case, we show that subword tokenization can reduce the length of the input sequence significantly by a factor of $\sim$8.7.

\newpage

%% file: chapter/sec-6-conclusion.tex
\chapter{Conclusion}\label{sec-conclusion}

In this thesis, in light of improving the efficiency of the Transformer model, we introduce Greenformers. Greenformers is a collection of efficient methods for improving the efficiency of the transformer model in terms of computation, memory, and/or the number of parameters. We focus our Greenformers on improving the transformer model efficiency with the low-rank approximation method as low-approximation can not only greatly improve both computational and memory efficiency, but also reducing the number of the model parameters significantly. We incorporate two efficient low-rank transformer model alternatives in Greenformers, i.e., Low-Rank Transformer (LRT) and Linformer. We conduct multiple experiments to show the effectiveness of applying LRT and Linformer in comparison to the original Transformer model.

Firstly, we conduct a comprehensive study on these two Transformer model variants by performing a thorough efficiency analysis to get a deeper understanding of the best data characteristic that is suitable to apply each model variant. Based on our analysis, the LRT model is suitable for improving both the time and memory efficiency in processing short-sequence ($\leq512$) input data, while the Linformer model is suitable for improving the efficiency in processing long-sequence input data ($\geq512$). We also show that LRT is more suitable for on-device deployment, as it significantly reduces the model size as the rank decreases. Additionally, we compare the evaluation performance of the Transformer, LRT, and Linformer model in the MNIST digit recognition task where we show that LRT and Linformer can retain the evaluation performance of the Transformer model while significantly increase its computational and memory efficiency. Lastly, we estimate the efficiency factor for applying LRT to the BERT$_{BASE}$ model which shows a significant reduction for developing a pre-trained BERT$_{BASE}$ model in terms of computational, economical, and environmental cost by more than 30\% of the original cost.

Secondly, we present the effectiveness of our LRT model in the speech recognition task. Our LRT model can reduce the model size and speed up the model compared to the original transformer model while maintaining similar performance. We conduct experiments on two speech recognition datasets, AiShell-1 and HKUST, to analyze the effect of rank $r$ on the performance and efficiency trade-off of our LRT model. Our result suggests that with a model compression rate up to ~60\%, there is no degradation in terms of the evaluation performance and the low-rank approximation can even work as a regularizer and increase the generalization of the model. With LRT, we can gain speed improvement up to 1.35x in the GPU and 1.23x in the CPU compared to the original transformer model. This finding suggests that ASR models tend to be overparameterized and by reducing the noisy parameters through low-rank approximation, we can produce a model with better generalization.


Lastly, we extend the applicability of efficient transformer models to genomics studies. We apply the sequence modeling technique with transformer models to predict Alzheimer's disease in the Chinese cohort. We represent the problem as a long sequence prediction problem, where we feed the model with a genotype of individual sequencing data from a well-studied APOE region in chromosome 19. We utilize the Linformer model to handle long sequences up to $\sim$35,000 nucleotides long and we show that self-supervised pre-training of a Linformer model can help to boost the performance of the model by around 5\% AUC compared to the baseline model. We further show that our Linformer model can capture the common features in the region which is known to be related to Alzheimer's disease. Additionally, we show that the tokenization technique can also provide a huge computation and memory efficiency by chunking the sequence which reduces the overall sequence length to be processed by the model.

The main contribution of this thesis is to introduce Greenformers and emphasizes its effectiveness in improving the efficiency of the Transformer model by providing a huge reduction in terms of storage, memory, and computational cost without sacrificing the overall quality of the model. Additionally, this approach can open up further research direction on long sequence processing tasks, such as long document summarization, video processing, and genomics tasks. We hope in the future other Greenformers can be extended to cover more methods for improving the efficiency of a Transformer model and is further applied to the existing Transformer model such as BERT and GPT2 models.

\newpage

%% file: chapter/sec-publications.tex
\null\skip0.2in
\begin{center}
{\bf \Large \underline{List of Publications}}
\end{center}
\vspace{12mm}

(* denotes equal contribution)
\begin{itemize}
    \item Genta Indra Winata, \textbf{Samuel Cahyawijaya}, Zihan Liu, Zhaojiang Lin,  Andrea Madotto, Pascale Fung. "Are Multilingual Models Effective in Code-Switching?" In Proceedings of the Fifth Workshop on Computational Approaches to Linguistic Code-Switching, 2021.
    \item Wenliang Dai* Liu, \textbf{Samuel Cahyawijaya}*, Zihan Liu, Pascale Fung. "Multimodal End-to-End Sparse Model for Emotion Recognition." In Proceedings of the 2021 Conference of the North American Chapter of the Association for Computational Linguistics: Human Language Technologies, 2021.
    \item Ye Jin Bang*, Etsuko Ishii*, \textbf{Samuel Cahyawijaya}*, Ziwei Ji*, Pascale Fung, "Model Generalization on COVID-19 Fake News Detection", 1st International Workshop on Combating Online Hostile Posts in Regional Languages during Emergency Situation, CONSTRAINT 2021 co-located with 35th AAAI Conference on Artificial Intelligence, AAAI 2021.
    \item Zihan Liu, Yan Xu, Tiezheng Yu, Wenliang Dai, Ziwei Ji, \textbf{Samuel Cahyawijaya}, Andrea Madotto, Pascale Fung, "CrossNER: Evaluating Cross-Domain Named Entity Recognition", 35th AAAI Conference on Artificial Intelligence, AAAI 2021.
    \item Zihan Liu, Genta Indra Winata, \textbf{Samuel Cahyawijaya}, Andrea Madotto, Zhaojiang Lin, Pascale Fung. "On the Importance of Word Order Information in Cross-lingual Sequence Labeling." In AAAI, 2021.
    \item Andrea Madotto, \textbf{Samuel Cahyawijaya}, \textbf{Genta Indra Winata}, Yan Xu, Zihan Liu, Zhaojiang Lin, Pascale Fung. "Learning Knowledge Bases with Parameters for Task-Oriented Dialogue Systems." In Proceedings of the 2020 Conference on Empirical Methods in Natural Language Processing: Findings, 2020.
    \item Bryan Wilie*, Karissa Vincentio*, Genta Indra Winata*, \textbf{Samuel Cahyawijaya}*, Xiaohong Li, Zhi Yuan Lim, Sidik Soleman, Rahmad Mahendra, Pascale Fung, Syafri Bahar, Ayu Purwarianti. "IndoNLU: Benchmark and resources for evaluating indonesian natural language understanding." In Proceedings of the 1st Conference of the Asia-Pacific Chapter of the Association for Computational Linguistics and the 10th International Joint Conference on Natural Language Processing, 2020.
    \item Genta Indra Winata*, \textbf{Samuel Cahyawijaya}*, Zhaojiang Lin, Zihan Liu, Peng Xu, Pascale Fung. "Meta-transfer learning for code-switched speech recognition." In Proceedings of the 58th Annual Meeting of the Association for Computational Linguistics, 2020.
    \item Genta Indra Winata*, \textbf{Samuel Cahyawijaya}*, Zihan Liu*, Zhaojiang Lin, Andrea Madotto, Peng Xu, Pascale Fung. "Learning Fast Adaptation on Cross-Accented Speech Recognition." In INTERSPEECH, 2020.
    \item Genta Indra Winata*, \textbf{Samuel Cahyawijaya}*, Zhaojiang Lin, Zihan Liu, and Pascale Fung. "Lightweight and Efficient End-to-End Speech Recognition Using Low-Rank Transformer." In ICASSP 2020-2020 IEEE International Conference on Acoustics, Speech and Signal Processing (ICASSP), pp. 6144-6148. IEEE, 2020.

\end{itemize}

%% file: main.bbl
\begin{thebibliography}{100}

\bibitem{ainslie2020etc}
Joshua Ainslie, Santiago Ontanon, Chris Alberti, Vaclav Cvicek, Zachary Fisher,
  Philip Pham, Anirudh Ravula, Sumit Sanghai, Qifan Wang, and Li~Yang.
\newblock {ETC}: Encoding long and structured inputs in transformers.
\newblock In {\em Proceedings of the 2020 Conference on Empirical Methods in
  Natural Language Processing (EMNLP)}, pages 268--284, Online, November 2020.
  Association for Computational Linguistics.

\bibitem{anderson1998gata}
Kathleen~P. Anderson, Scott~C. Crable, and Jerry~B Lingrel.
\newblock Multiple proteins binding to a {GATA}-e box-{GATA} motif regulate the
  erythroid kr\"{u}ppel-like factor ({EKLF}) gene.
\newblock {\em Journal of Biological Chemistry}, 273(23):14347--14354, June
  1998.

\bibitem{abenmacher2020comparability}
Matthias Aßenmacher and Christian Heumann.
\newblock On the comparability of pre-trained language models, 2020.

\bibitem{ba2016layernorm}
Jimmy~Lei Ba, Jamie~Ryan Kiros, and Geoffrey~E. Hinton.
\newblock Layer normalization, 2016.

\bibitem{bai2020binarybert}
Haoli Bai, Wei Zhang, Lu~Hou, Lifeng Shang, Jing Jin, Xin Jiang, Qun Liu,
  Michael Lyu, and Irwin King.
\newblock Binarybert: Pushing the limit of bert quantization, 2020.

\bibitem{fairscale2021}
Mandeep Baines, Shruti Bhosale, Vittorio Caggiano, Naman Goyal, Siddharth
  Goyal, Myle Ott, Benjamin Lefaudeux, Vitaliy Liptchinsky, Mike Rabatt, Sam
  Sheiffer, Anjali Sridhar, and Min Xu.
\newblock Fairscale: A general purpose modular pytorch library for high
  performance and large scale training.
\newblock \url{https://github.com/facebookresearch/fairscale}, 2021.

\bibitem{bang2021fakenews}
Yejin Bang, Etsuko Ishii, Samuel Cahyawijaya, Ziwei Ji, and Pascale Fung.
\newblock Model generalization on covid-19 fake news detection.
\newblock In Tanmoy Chakraborty, Kai Shu, H.~Russell Bernard, Huan Liu, and
  Md~Shad Akhtar, editors, {\em Combating Online Hostile Posts in Regional
  Languages during Emergency Situation}, pages 128--140, Cham, 2021. Springer
  International Publishing.

\bibitem{barnett2011bamtools}
D.~W. Barnett, E.~K. Garrison, A.~R. Quinlan, M.~P. Stromberg, and G.~T. Marth.
\newblock {BamTools}: a c++ api and toolkit for analyzing and managing {BAM}
  files.
\newblock {\em Bioinformatics}, 27(12):1691--1692, April 2011.

\bibitem{belloy2019apoe}
Michaël~E. Belloy, Valerio Napolioni, and Michael~D. Greicius.
\newblock A quarter century of {APOE} and alzheimer's disease: Progress to date
  and the path forward.
\newblock {\em Neuron}, 101(5):820--838, March 2019.

\bibitem{beltagy2020longformer}
Iz~Beltagy, Matthew~E. Peters, and Arman Cohan.
\newblock Longformer: The long-document transformer, 2020.

\bibitem{blackwood1998enhancer}
E.~M. Blackwood.
\newblock Going the distance: A current view of enhancer action.
\newblock {\em Science}, 281(5373):60--63, July 1998.

\bibitem{brown2020gpt3}
Tom Brown, Benjamin Mann, Nick Ryder, Melanie Subbiah, Jared~D Kaplan, Prafulla
  Dhariwal, Arvind Neelakantan, Pranav Shyam, Girish Sastry, Amanda Askell,
  Sandhini Agarwal, Ariel Herbert-Voss, Gretchen Krueger, Tom Henighan, Rewon
  Child, Aditya Ramesh, Daniel Ziegler, Jeffrey Wu, Clemens Winter, Chris
  Hesse, Mark Chen, Eric Sigler, Mateusz Litwin, Scott Gray, Benjamin Chess,
  Jack Clark, Christopher Berner, Sam McCandlish, Alec Radford, Ilya Sutskever,
  and Dario Amodei.
\newblock Language models are few-shot learners.
\newblock In H.~Larochelle, M.~Ranzato, R.~Hadsell, M.~F. Balcan, and H.~Lin,
  editors, {\em Advances in Neural Information Processing Systems}, volume~33,
  pages 1877--1901. Curran Associates, Inc., 2020.

\bibitem{bu2017aishell1}
Hui Bu, Jiayu Du, Xingyu Na, Bengu Wu, and Hao Zheng.
\newblock Aishell-1: An open-source mandarin speech corpus and a speech
  recognition baseline.
\newblock In {\em Oriental COCOSDA 2017}, page Submitted, 2017.

\bibitem{cahyawijaya2021indonlg}
Samuel Cahyawijaya, Genta~Indra Winata, Bryan Wilie, Karissa Vincentio,
  Xiaohong Li, Adhiguna Kuncoro, Sebastian Ruder, Zhi~Yuan Lim, Syafri Bahar,
  Masayu~Leylia Khodra, Ayu Purwarianti, and Pascale Fung.
\newblock Indonlg: Benchmark and resources for evaluating indonesian natural
  language generation, 2021.

\bibitem{cecato2011mocaperf}
Juliana Cecato, José~Eduardo Martinelli, Ivan Aprahamian, and Mônica Yassuda.
\newblock P3-069: Moca contributions to differential diagnosis among normal
  controls, mild cognitive impairment and alzheimer's disease in brazil.
\newblock {\em Alzheimer's \& Dementia}, 7(4S\_Part\_15):S535--S535, 2011.

\bibitem{chan2016listen}
William Chan, Navdeep Jaitly, Quoc Le, and Oriol Vinyals.
\newblock Listen, attend and spell: A neural network for large vocabulary
  conversational speech recognition.
\newblock In {\em 2016 IEEE International Conference on Acoustics, Speech and
  Signal Processing (ICASSP)}, pages 4960--4964. IEEE, 2016.

\bibitem{chen2021prunedistil}
Liyang Chen, Yongquan Chen, Juntong Xi, and Xinyi Le.
\newblock Knowledge from the original network: restore a better pruned network
  with knowledge distillation.
\newblock {\em Complex {\&} Intelligent Systems}, January 2021.

\bibitem{choromanski2021performer}
Krzysztof~Marcin Choromanski, Valerii Likhosherstov, David Dohan, Xingyou Song,
  Andreea Gane, Tamas Sarlos, Peter Hawkins, Jared~Quincy Davis, Afroz
  Mohiuddin, Lukasz Kaiser, David~Benjamin Belanger, Lucy~J Colwell, and Adrian
  Weller.
\newblock Rethinking attention with performers.
\newblock In {\em International Conference on Learning Representations}, 2021.

\bibitem{church2011hg19}
Deanna~M. Church, Valerie~A. Schneider, Tina Graves, Katherine Auger, Fiona
  Cunningham, Nathan Bouk, Hsiu-Chuan Chen, Richa Agarwala, William~M. McLaren,
  Graham~R.S. Ritchie, Derek Albracht, Milinn Kremitzki, Susan Rock, Holland
  Kotkiewicz, Colin Kremitzki, Aye Wollam, Lee Trani, Lucinda Fulton, Robert
  Fulton, Lucy Matthews, Siobhan Whitehead, Will Chow, James Torrance, Matthew
  Dunn, Glenn Harden, Glen Threadgold, Jonathan Wood, Joanna Collins, Paul
  Heath, Guy Griffiths, Sarah Pelan, Darren Grafham, Evan~E. Eichler, George
  Weinstock, Elaine~R. Mardis, Richard~K. Wilson, Kerstin Howe, Paul Flicek,
  and Tim Hubbard.
\newblock Modernizing reference genome assemblies.
\newblock {\em {PLoS} Biology}, 9(7):e1001091, July 2011.

\bibitem{dai2021mesm}
Wenliang Dai, Samuel Cahyawijaya, Zihan Liu, and Pascale Fung.
\newblock Multimodal end-to-end sparse model for emotion recognition.
\newblock In {\em NAACL}, 2021.

\bibitem{dai2019transformerxl}
Zihang Dai, Zhilin Yang, Yiming Yang, Jaime Carbonell, Quoc Le, and Ruslan
  Salakhutdinov.
\newblock Transformer-{XL}: Attentive language models beyond a fixed-length
  context.
\newblock In {\em Proceedings of the 57th Annual Meeting of the Association for
  Computational Linguistics}, pages 2978--2988, Florence, Italy, July 2019.
  Association for Computational Linguistics.

\bibitem{darawi2013aspcr-ad}
Mohd~Nazif Darawi, Chin Ai-Vyrn, Kalavathy Ramasamy, Philip Poi~Jun Hua,
  Tan~Maw Pin, Shahrul~Bahyah Kamaruzzaman, and Abu Bakar~Abdul Majeed.
\newblock Allele-specific polymerase chain reaction for the detection of
  alzheimer's disease-related single nucleotide polymorphisms.
\newblock {\em {BMC} Medical Genetics}, 14(1), February 2013.

\bibitem{denil2013predictparam}
Misha Denil, Babak Shakibi, Laurent Dinh, Marc\textquotesingle~Aurelio Ranzato,
  and Nando de~Freitas.
\newblock Predicting parameters in deep learning.
\newblock In C.~J.~C. Burges, L.~Bottou, M.~Welling, Z.~Ghahramani, and K.~Q.
  Weinberger, editors, {\em Advances in Neural Information Processing Systems},
  volume~26. Curran Associates, Inc., 2013.

\bibitem{deture2019neuropathological}
Michael~A. DeTure and Dennis~W. Dickson.
\newblock The neuropathological diagnosis of alzheimer's disease.
\newblock {\em Molecular Neurodegeneration}, 14(1):32, Aug 2019.

\bibitem{devlin2019bert}
Jacob Devlin, Ming-Wei Chang, Kenton Lee, and Kristina Toutanova.
\newblock {BERT}: Pre-training of deep bidirectional transformers for language
  understanding.
\newblock In {\em Proceedings of the 2019 Conference of the North {A}merican
  Chapter of the Association for Computational Linguistics: Human Language
  Technologies, Volume 1 (Long and Short Papers)}, pages 4171--4186,
  Minneapolis, Minnesota, June 2019. Association for Computational Linguistics.

\bibitem{dodge2019show}
Jesse Dodge, Suchin Gururangan, Dallas Card, Roy Schwartz, and Noah~A. Smith.
\newblock Show your work: Improved reporting of experimental results.
\newblock In {\em Proceedings of the 2019 Conference on Empirical Methods in
  Natural Language Processing and the 9th International Joint Conference on
  Natural Language Processing (EMNLP-IJCNLP)}, pages 2185--2194, Hong Kong,
  China, November 2019. Association for Computational Linguistics.

\bibitem{dong2018speech}
Linhao Dong, Shuang Xu, and Bo~Xu.
\newblock Speech-transformer: a no-recurrence sequence-to-sequence model for
  speech recognition.
\newblock In {\em 2018 IEEE International Conference on Acoustics, Speech and
  Signal Processing (ICASSP)}, pages 5884--5888. IEEE, 2018.

\bibitem{fedus2021switch}
William Fedus, Barret Zoph, and Noam Shazeer.
\newblock Switch transformers: Scaling to trillion parameter models with simple
  and efficient sparsity, 2021.

\bibitem{finn2017maml}
Chelsea Finn, Pieter Abbeel, and Sergey Levine.
\newblock Model-agnostic meta-learning for fast adaptation of deep networks.
\newblock In Doina Precup and Yee~Whye Teh, editors, {\em Proceedings of the
  34th International Conference on Machine Learning}, volume~70 of {\em
  Proceedings of Machine Learning Research}, pages 1126--1135. PMLR, 06--11 Aug
  2017.

\bibitem{fishilevich2017genehancer}
Simon Fishilevich, Ron Nudel, Noa Rappaport, Rotem Hadar, Inbar Plaschkes,
  Tsippi~Iny Stein, Naomi Rosen, Asher Kohn, Michal Twik, Marilyn Safran, Doron
  Lancet, and Dana Cohen.
\newblock {GeneHancer}: genome-wide integration of enhancers and target genes
  in {GeneCards}.
\newblock {\em Database}, 2017, January 2017.

\bibitem{frankle2019lottery}
Jonathan Frankle and Michael Carbin.
\newblock The lottery ticket hypothesis: Finding sparse, trainable neural
  networks.
\newblock In {\em ICLR}. OpenReview.net, 2019.

\bibitem{fuchs2014gene-length}
Gilad Fuchs, Yoav Voichek, Sima Benjamin, Shlomit Gilad, Ido Amit, and Moshe
  Oren.
\newblock 4sudrb-seq: measuring genomewide transcriptional elongation rates and
  initiation frequencies within cells.
\newblock {\em Genome Biology}, 15(5), May 2014.

\bibitem{gao2018efficientsl}
Fei Gao, Lijun Wu, L.~Zhao, Tao Qin, Xueqi Cheng, and Tie-Yan Liu.
\newblock Efficient sequence learning with group recurrent networks.
\newblock In {\em NAACL-HLT}, 2018.

\bibitem{ghandi2014gkmer}
Mahmoud Ghandi, Dongwon Lee, Morteza Mohammad-Noori, and Michael~A. Beer.
\newblock Enhanced regulatory sequence prediction using gapped k-mer features.
\newblock {\em PLoS computational biology}, 10(7):e1003711--e1003711, Jul 2014.

\bibitem{golub1970svd}
G.~H. Golub and C.~Reinsch.
\newblock Singular value decomposition and least squares solutions.
\newblock {\em Numer. Math.}, 14(5):403–420, April 1970.

\bibitem{golub1996matrix}
Gene~H. Golub and Charles~F. Van~Loan.
\newblock {\em Matrix Computations (3rd Ed.)}.
\newblock Johns Hopkins University Press, USA, 1996.

\bibitem{graves2016rnn}
Alex Graves.
\newblock {\em Supervised Sequence Labelling with Recurrent Neural Networks}.
\newblock 2011.

\bibitem{guo2016svddenoising}
Qiang Guo, Caiming Zhang, Yunfeng Zhang, and Hui Liu.
\newblock An efficient svd-based method for image denoising.
\newblock {\em IEEE Transactions on Circuits and Systems for Video Technology},
  26(5):868--880, 2016.

\bibitem{han2015learningprune}
Song Han, Jeff Pool, John Tran, and William Dally.
\newblock Learning both weights and connections for efficient neural network.
\newblock In C.~Cortes, N.~Lawrence, D.~Lee, M.~Sugiyama, and R.~Garnett,
  editors, {\em Advances in Neural Information Processing Systems}, volume~28.
  Curran Associates, Inc., 2015.

\bibitem{hasibi1993sodpruning}
Babak Hassibi and David Stork.
\newblock Second order derivatives for network pruning: Optimal brain surgeon.
\newblock In S.~Hanson, J.~Cowan, and C.~Giles, editors, {\em Advances in
  Neural Information Processing Systems}, volume~5. Morgan-Kaufmann, 1993.

\bibitem{hendrycks2016gelu}
Dan Hendrycks and Kevin Gimpel.
\newblock Bridging nonlinearities and stochastic regularizers with gaussian
  error linear units.
\newblock {\em CoRR}, abs/1606.08415, 2016.

\bibitem{hernandez2020measuring}
Danny Hernandez and Tom~B. Brown.
\newblock Measuring the algorithmic efficiency of neural networks, 2020.

\bibitem{hinton2015distil}
Geoffrey Hinton, Oriol Vinyals, and Jeffrey Dean.
\newblock Distilling the knowledge in a neural network.
\newblock In {\em NIPS Deep Learning and Representation Learning Workshop},
  2015.

\bibitem{hori2017advances}
Takaaki Hori, Shinji Watanabe, Yu~Zhang, and William Chan.
\newblock Advances in joint ctc-attention based end-to-end speech recognition
  with a deep cnn encoder and rnn-lm.
\newblock {\em Proc. Interspeech 2017}, pages 949--953, 2017.

\bibitem{huynh2017ad}
Rose~Ann Huynh and Chandra Mohan.
\newblock Alzheimer's disease: Biomarkers in the genome, blood, and
  cerebrospinal fluid.
\newblock {\em Frontiers in Neurology}, 8, March 2017.

\bibitem{jacob2018quantization}
Benoit Jacob, Skirmantas Kligys, Bo~Chen, Menglong Zhu, Matthew Tang, Andrew
  Howard, Hartwig Adam, and Dmitry Kalenichenko.
\newblock Quantization and training of neural networks for efficient
  integer-arithmetic-only inference.
\newblock In {\em Proceedings of the IEEE Conference on Computer Vision and
  Pattern Recognition (CVPR)}, June 2018.

\bibitem{ji2020dnabert}
Yanrong Ji, Zhihan Zhou, Han Liu, and Ramana~V Davuluri.
\newblock {DNABERT}: pre-trained bidirectional encoder representations from
  transformers model for {DNA}-language in genome.
\newblock September 2020.

\bibitem{johannsen1911genotype}
W.~Johannsen.
\newblock The genotype conception of heredity.
\newblock {\em The American Naturalist}, 45(531):129--159, March 1911.

\bibitem{johnson2017googles}
Melvin Johnson, Mike Schuster, Quoc~V. Le, Maxim Krikun, Yonghui Wu, Zhifeng
  Chen, Nikhil Thorat, Fernanda Vi{\'e}gas, Martin Wattenberg, Greg Corrado,
  Macduff Hughes, and Jeffrey Dean.
\newblock {G}oogle{'}s multilingual neural machine translation system: Enabling
  zero-shot translation.
\newblock {\em Transactions of the Association for Computational Linguistics},
  5:339--351, 2017.

\bibitem{jung2019quantize}
S.~Jung, Changyong Son, Seohyung Lee, JinWoo Son, Jae-Joon Han, Youngjun Kwak,
  Sung~Ju Hwang, and Changkyu Choi.
\newblock Learning to quantize deep networks by optimizing quantization
  intervals with task loss.
\newblock {\em 2019 IEEE/CVF Conference on Computer Vision and Pattern
  Recognition (CVPR)}, pages 4345--4354, 2019.

\bibitem{karni2007promoter}
Shaul Karni.
\newblock Analysis of biological networks : Transcriptional networks-promoter
  sequence analysis.
\newblock Tel Aviv University, 2007.

\bibitem{katharopoulos2020lineartrf}
Angelos Katharopoulos, Apoorv Vyas, Nikolaos Pappas, and Fran{\c{c}}ois
  Fleuret.
\newblock Transformers are {RNN}s: Fast autoregressive transformers with linear
  attention.
\newblock In Hal~Daumé III and Aarti Singh, editors, {\em Proceedings of the
  37th International Conference on Machine Learning}, volume 119 of {\em
  Proceedings of Machine Learning Research}, pages 5156--5165. PMLR, 13--18 Jul
  2020.

\bibitem{kim2017joint}
Suyoun Kim, Takaaki Hori, and Shinji Watanabe.
\newblock Joint ctc-attention based end-to-end speech recognition using
  multi-task learning.
\newblock In {\em 2017 IEEE International Conference on Acoustics, Speech and
  Signal Processing (ICASSP)}, pages 4835--4839. IEEE, 2017.

\bibitem{kingma2015adam}
Diederik~P. Kingma and Jimmy Ba.
\newblock Adam: {A} method for stochastic optimization.
\newblock In Yoshua Bengio and Yann LeCun, editors, {\em 3rd International
  Conference on Learning Representations, {ICLR} 2015, San Diego, CA, USA, May
  7-9, 2015, Conference Track Proceedings}, 2015.

\bibitem{kitaev2020reformer}
Nikita Kitaev, Lukasz Kaiser, and Anselm Levskaya.
\newblock Reformer: The efficient transformer.
\newblock In {\em International Conference on Learning Representations}, 2020.

\bibitem{kobayashi2020influence}
Sosuke Kobayashi, Sho Yokoi, Jun Suzuki, and Kentaro Inui.
\newblock Efficient estimation of influence of a training instance.
\newblock In {\em Proceedings of SustaiNLP: Workshop on Simple and Efficient
  Natural Language Processing}, pages 41--47, Online, November 2020.
  Association for Computational Linguistics.

\bibitem{krizhevsky2012alexnet}
Alex Krizhevsky, Ilya Sutskever, and Geoffrey~E Hinton.
\newblock Imagenet classification with deep convolutional neural networks.
\newblock In F.~Pereira, C.~J.~C. Burges, L.~Bottou, and K.~Q. Weinberger,
  editors, {\em Advances in Neural Information Processing Systems}, volume~25.
  Curran Associates, Inc., 2012.

\bibitem{kuchaiev2017factorization}
Oleksii Kuchaiev and Boris Ginsburg.
\newblock Factorization tricks for lstm networks.
\newblock {\em ICLR Workshop}, 2017.

\bibitem{kudo2018sentencepiece}
Taku Kudo and John Richardson.
\newblock {S}entence{P}iece: A simple and language independent subword
  tokenizer and detokenizer for neural text processing.
\newblock In {\em Proceedings of the 2018 Conference on Empirical Methods in
  Natural Language Processing: System Demonstrations}, pages 66--71, Brussels,
  Belgium, November 2018. Association for Computational Linguistics.

\bibitem{lan2020albert}
Zhenzhong Lan, Mingda Chen, Sebastian Goodman, Kevin Gimpel, Piyush Sharma, and
  Radu Soricut.
\newblock Albert: A lite bert for self-supervised learning of language
  representations.
\newblock In {\em International Conference on Learning Representations}, 2020.

\bibitem{lecun1990obd}
Yann LeCun, John Denker, and Sara Solla.
\newblock Optimal brain damage.
\newblock In D.~Touretzky, editor, {\em Advances in Neural Information
  Processing Systems}, volume~2. Morgan-Kaufmann, 1990.

\bibitem{lee2001nmf}
Daniel Lee and H.~Sebastian Seung.
\newblock Algorithms for non-negative matrix factorization.
\newblock In T.~Leen, T.~Dietterich, and V.~Tresp, editors, {\em Advances in
  Neural Information Processing Systems}, volume~13. MIT Press, 2001.

\bibitem{lee2019settrf}
Juho Lee, Yoonho Lee, Jungtaek Kim, Adam Kosiorek, Seungjin Choi, and Yee~Whye
  Teh.
\newblock Set transformer: A framework for attention-based
  permutation-invariant neural networks.
\newblock In Kamalika Chaudhuri and Ruslan Salakhutdinov, editors, {\em
  Proceedings of the 36th International Conference on Machine Learning},
  volume~97 of {\em Proceedings of Machine Learning Research}, pages
  3744--3753. PMLR, 09--15 Jun 2019.

\bibitem{lee2019denoise}
Nayeon Lee, Zihan Liu, and Pascale Fung.
\newblock Team yeon-zi at {S}em{E}val-2019 task 4: Hyperpartisan news detection
  by de-noising weakly-labeled data.
\newblock In {\em Proceedings of the 13th International Workshop on Semantic
  Evaluation}, pages 1052--1056, Minneapolis, Minnesota, USA, June 2019.
  Association for Computational Linguistics.

\bibitem{li2009samtools}
H.~Li, B.~Handsaker, A.~Wysoker, T.~Fennell, J.~Ruan, N.~Homer, G.~Marth,
  G.~Abecasis, and R.~Durbin and.
\newblock The sequence alignment/map format and {SAMtools}.
\newblock {\em Bioinformatics}, 25(16):2078--2079, June 2009.

\bibitem{hao2016pruning}
Hao Li, Asim Kadav, Igor Durdanovic, Hanan Samet, and Hans~Peter Graf.
\newblock Pruning filters for efficient convnets.
\newblock {\em CoRR}, abs/1608.08710, 2016.

\bibitem{li2019speechtransformer}
Jie Li, Xiaorui Wang, Yan Li, et~al.
\newblock The speechtransformer for large-scale mandarin chinese speech
  recognition.
\newblock In {\em ICASSP 2019-2019 IEEE International Conference on Acoustics,
  Speech and Signal Processing (ICASSP)}, pages 7095--7099. IEEE, 2019.

\bibitem{li2019framewise}
Mohan Li, Yuanjiang Cao, Weicong Zhou, and Min Liu.
\newblock Framewise supervised training towards end-to-end speech recognition
  models: First results.
\newblock {\em Proc. Interspeech 2019}, pages 1641--1645, 2019.

\bibitem{li2019end}
Mohan Li, Min Liu, and Hattori Masanori.
\newblock End-to-end speech recognition with adaptive computation steps.
\newblock In {\em ICASSP 2019-2019 IEEE International Conference on Acoustics,
  Speech and Signal Processing (ICASSP)}, pages 6246--6250. IEEE, 2019.

\bibitem{liu2006hkust}
Yi~Liu, Pascale Fung, Yongsheng Yang, Christopher Cieri, Shudong Huang, and
  David Graff.
\newblock Hkust/mts: A very large scale mandarin telephone speech corpus.
\newblock In {\em Proceedings of the 5th International Conference on Chinese
  Spoken Language Processing}, ISCSLP'06, page 724–735, Berlin, Heidelberg,
  2006. Springer-Verlag.

\bibitem{loshchilov2018decoupled}
Ilya Loshchilov and Frank Hutter.
\newblock Decoupled weight decay regularization.
\newblock In {\em International Conference on Learning Representations}, 2019.

\bibitem{lundin1994gcbox}
M~Lundin, J~O Nehlin, and H~Ronne.
\newblock Importance of a flanking {AT}-rich region in target site recognition
  by the {GC} box-binding zinc finger protein {MIG}1.
\newblock {\em Molecular and Cellular Biology}, 14(3):1979--1985, March 1994.

\bibitem{madotto2020learning}
Andrea Madotto, Samuel Cahyawijaya, Genta~Indra Winata, Yan Xu, Zihan Liu,
  Zhaojiang Lin, and Pascale Fung.
\newblock Learning knowledge bases with parameters for task-oriented dialogue
  systems.
\newblock In {\em Findings of the Association for Computational Linguistics:
  EMNLP 2020}, pages 2372--2394, Online, November 2020. Association for
  Computational Linguistics.

\bibitem{miao2016empirical}
Yajie Miao, Mohammad Gowayyed, Xingyu Na, Tom Ko, Florian Metze, and Alexander
  Waibel.
\newblock An empirical exploration of ctc acoustic models.
\newblock In {\em 2016 IEEE International Conference on Acoustics, Speech and
  Signal Processing (ICASSP)}, pages 2623--2627. IEEE, 2016.

\bibitem{micikevicius2018mixed}
Paulius Micikevicius, Sharan Narang, Jonah Alben, Gregory Diamos, Erich Elsen,
  David Garcia, Boris Ginsburg, Michael Houston, Oleksii Kuchaiev, Ganesh
  Venkatesh, and Hao Wu.
\newblock Mixed precision training.
\newblock In {\em International Conference on Learning Representations}, 2018.

\bibitem{min2017kmeremb}
Xu~Min, Wanwen Zeng, Ning Chen, Ting Chen, and Rui Jiang.
\newblock {Chromatin accessibility prediction via convolutional long short-term
  memory networks with k-mer embedding}.
\newblock {\em Bioinformatics}, 33(14):i92--i101, 07 2017.

\bibitem{min2017denhancer}
Xu~Min, Wanwen Zeng, Shengquan Chen, Ning Chen, Ting Chen, and Rui Jiang.
\newblock Predicting enhancers with deep convolutional neural networks.
\newblock {\em {BMC} Bioinformatics}, 18(S13), November 2017.

\bibitem{mirza2020efficientol}
A.~Mirza, Mine Kerpicci, and S.~Kozat.
\newblock Efficient online learning with improved lstm neural networks.
\newblock {\em Digit. Signal Process.}, 102:102742, 2020.

\bibitem{nair2010relu}
Vinod Nair and Geoffrey~E. Hinton.
\newblock Rectified linear units improve restricted boltzmann machines.
\newblock In {\em Proceedings of the 27th International Conference on
  International Conference on Machine Learning}, ICML'10, page 807–814,
  Madison, WI, USA, 2010. Omnipress.

\bibitem{nasreddine2005moca}
Ziad~S. Nasreddine, Natalie~A. Phillips, Val{\~{A}}{\textcopyright}rie
  B{\~{A}}{\textcopyright}dirian, Simon Charbonneau, Victor Whitehead, Isabelle
  Collin, Jeffrey~L. Cummings, and Howard Chertkow.
\newblock The montreal cognitive assessment, {MoCA}: A brief screening tool for
  mild cognitive impairment.
\newblock {\em Journal of the American Geriatrics Society}, 53(4):695--699,
  April 2005.

\bibitem{ogilvy2007gata}
S.~Ogilvy, R.~Ferreira, S.~G. Piltz, J.~M. Bowen, B.~Gottgens, and A.~R. Green.
\newblock The {SCL} +40 enhancer targets the midbrain together with primitive
  and definitive hematopoiesis and is regulated by {SCL} and {GATA} proteins.
\newblock {\em Molecular and Cellular Biology}, 27(20):7206--7219, October
  2007.

\bibitem{armar2018imagetrf}
Niki Parmar, Ashish Vaswani, Jakob Uszkoreit, Lukasz Kaiser, Noam Shazeer,
  Alexander Ku, and Dustin Tran.
\newblock Image transformer.
\newblock In Jennifer Dy and Andreas Krause, editors, {\em Proceedings of the
  35th International Conference on Machine Learning}, volume~80 of {\em
  Proceedings of Machine Learning Research}, pages 4055--4064. PMLR, 10--15 Jul
  2018.

\bibitem{pennacchio2013enhancer}
Len~A. Pennacchio, Wendy Bickmore, Ann Dean, Marcelo~A. Nobrega, and Gill
  Bejerano.
\newblock Enhancers: five essential questions.
\newblock {\em Nature Reviews Genetics}, 14(4):288--295, March 2013.

\bibitem{peters2018elmo}
Matthew~E. Peters, Mark Neumann, Mohit Iyyer, Matt Gardner, Christopher Clark,
  Kenton Lee, and Luke Zettlemoyer.
\newblock Deep contextualized word representations.
\newblock In {\em Proc. of NAACL}, 2018.

\bibitem{pham2021aspcr-hlac}
Tran Thu~Ha Pham, Quang~Binh Tran, Chonlaphat Sukasem, Van~Dinh Nguyen,
  Chi~Hieu Chu, Thi Quynh~Nga Do, Ngoc Phuong~Mai Tran, and Thanh~Huong Phung.
\newblock A novel allele-specific pcr protocol for the detection of the
  hla-c*03:02 allele, a pharmacogenetic marker, in vietnamese kinh people.
\newblock {\em The Application of Clinical Genetics}, Volume 14:27--35,
  February 2021.

\bibitem{qiu2020blockwise}
Jiezhong Qiu, Hao Ma, Omer Levy, Scott~Wen tau Yih, Sinong Wang, and Jie Tang.
\newblock Blockwise self-attention for long document understanding, 2020.

\bibitem{rabinovici2019load}
Gil~D. Rabinovici.
\newblock Late-onset alzheimer disease.
\newblock {\em Continuum (Minneapolis, Minn.)}, 25(1):14--33, Feb 2019.

\bibitem{radford2019gpt2}
Alec Radford, Jeff Wu, Rewon Child, David Luan, Dario Amodei, and Ilya
  Sutskever.
\newblock Language models are unsupervised multitask learners.
\newblock 2019.

\bibitem{rae2020compressivetrs}
Jack~W. Rae, Anna Potapenko, Siddhant~M. Jayakumar, Chloe Hillier, and
  Timothy~P. Lillicrap.
\newblock Compressive transformers for long-range sequence modelling.
\newblock In {\em International Conference on Learning Representations}, 2020.

\bibitem{rajbhandari2020zero}
Samyam Rajbhandari, Jeff Rasley, Olatunji Ruwase, and Yuxiong He.
\newblock Zero: Memory optimizations toward training trillion parameter models.
\newblock In {\em Proceedings of the International Conference for High
  Performance Computing, Networking, Storage and Analysis}, SC '20. IEEE Press,
  2020.

\bibitem{rasley2020deepspeed}
Jeff Rasley, Samyam Rajbhandari, Olatunji Ruwase, and Yuxiong He.
\newblock Deepspeed: System optimizations enable training deep learning models
  with over 100 billion parameters.
\newblock In {\em Proceedings of the 26th ACM SIGKDD International Conference
  on Knowledge Discovery \&amp; Data Mining}, KDD '20, page 3505–3506, New
  York, NY, USA, 2020. Association for Computing Machinery.

\bibitem{reed1993pruning}
R.~Reed.
\newblock Pruning algorithms-a survey.
\newblock {\em IEEE Transactions on Neural Networks}, 4(5):740--747, 1993.

\bibitem{roy2020routingtrf}
Aurko Roy*, Mohammad~Taghi Saffar*, David Grangier, and Ashish Vaswani.
\newblock Efficient content-based sparse attention with routing transformers,
  2020.

\bibitem{sanh2019distilbert}
Victor Sanh, Lysandre Debut, Julien Chaumond, and Thomas Wolf.
\newblock {DistilBERT, a distilled version of BERT: smaller, faster, cheaper
  and lighter}.
\newblock In {\em 5th Workshop on Energy Efficient Machine Learning and
  Cognitive Computing @ NeurIPS 2019}, 2019.

\bibitem{schwartz2019green}
Roy Schwartz, Jesse Dodge, Noah~A. Smith, and Oren Etzioni.
\newblock Green ai.
\newblock {\em Commun. ACM}, 63(12):54–63, November 2020.

\bibitem{shen2021efficientattn}
Zhuoran Shen, Mingyuan Zhang, Haiyu Zhao, Shuai Yi, and Hongsheng Li.
\newblock Efficient attention: Attention with linear complexities.
\newblock In {\em Proceedings of the IEEE/CVF Winter Conference on Applications
  of Computer Vision (WACV)}, pages 3531--3539, January 2021.

\bibitem{shoeybi2020megatronlm}
Mohammad Shoeybi, Mostofa Patwary, Raul Puri, Patrick LeGresley, Jared Casper,
  and Bryan Catanzaro.
\newblock Megatron-lm: Training multi-billion parameter language models using
  model parallelism, 2020.

\bibitem{shrikumar2019gkmexplain}
Avanti Shrikumar, Eva Prakash, and Anshul Kundaje.
\newblock {GkmExplain}: fast and accurate interpretation of nonlinear gapped
  k-mer {SVMs}.
\newblock {\em Bioinformatics}, 35(14):i173--i182, July 2019.

\bibitem{simonyan2014very}
Karen Simonyan and Andrew Zisserman.
\newblock Very deep convolutional networks for large-scale image recognition.
\newblock In {\em ICLR}, 2015.

\bibitem{strubell2019energy}
Emma Strubell, Ananya Ganesh, and Andrew McCallum.
\newblock Energy and policy considerations for deep learning in {NLP}.
\newblock In {\em Proceedings of the 57th Annual Meeting of the Association for
  Computational Linguistics}, pages 3645--3650, Florence, Italy, July 2019.
  Association for Computational Linguistics.

\bibitem{sun2020mobilebert}
Zhiqing Sun, Hongkun Yu, Xiaodan Song, Renjie Liu, Yiming Yang, and Denny Zhou.
\newblock {M}obile{BERT}: a compact task-agnostic {BERT} for resource-limited
  devices.
\newblock In {\em Proceedings of the 58th Annual Meeting of the Association for
  Computational Linguistics}, pages 2158--2170, Online, July 2020. Association
  for Computational Linguistics.

\bibitem{ilya2014seq2seq}
Ilya Sutskever, Oriol Vinyals, and Quoc~V Le.
\newblock Sequence to sequence learning with neural networks.
\newblock In Z.~Ghahramani, M.~Welling, C.~Cortes, N.~Lawrence, and K.~Q.
  Weinberger, editors, {\em Advances in Neural Information Processing Systems},
  volume~27. Curran Associates, Inc., 2014.

\bibitem{tan2020efficientnet}
Mingxing Tan and Quoc Le.
\newblock {E}fficient{N}et: Rethinking model scaling for convolutional neural
  networks.
\newblock In Kamalika Chaudhuri and Ruslan Salakhutdinov, editors, {\em
  Proceedings of the 36th International Conference on Machine Learning},
  volume~97 of {\em Proceedings of Machine Learning Research}, pages
  6105--6114. PMLR, 09--15 Jun 2019.

\bibitem{tay2020sinkhorn}
Yi~Tay, Dara Bahri, Liu Yang, Donald Metzler, and Da-Cheng Juan.
\newblock Sparse {S}inkhorn attention.
\newblock In Hal~Daumé III and Aarti Singh, editors, {\em Proceedings of the
  37th International Conference on Machine Learning}, volume 119 of {\em
  Proceedings of Machine Learning Research}, pages 9438--9447. PMLR, 13--18 Jul
  2020.

\bibitem{tay2021lra}
Yi~Tay, Mostafa Dehghani, Samira Abnar, Yikang Shen, Dara Bahri, Philip Pham,
  Jinfeng Rao, Liu Yang, Sebastian Ruder, and Don Metzler.
\newblock Long range arena : A benchmark for efficient transformers.
\newblock In {\em ICLR 2021}, 2021.

\bibitem{tay2021long}
Yi~Tay, Mostafa Dehghani, Samira Abnar, Yikang Shen, Dara Bahri, Philip Pham,
  Jinfeng Rao, Liu Yang, Sebastian Ruder, and Donald Metzler.
\newblock Long range arena : A benchmark for efficient transformers.
\newblock In {\em International Conference on Learning Representations}, 2021.

\bibitem{tay2020efficient}
Yi~Tay, Mostafa Dehghani, Dara Bahri, and Donald Metzler.
\newblock Efficient transformers: A survey, 2020.

\bibitem{tuske2020attention_lstm}
Zoltán Tüske, George Saon, Kartik Audhkhasi, and Brian Kingsbury.
\newblock {Single Headed Attention Based Sequence-to-Sequence Model for
  State-of-the-Art Results on Switchboard}.
\newblock In {\em Proc. Interspeech 2020}, pages 551--555, 2020.

\bibitem{umarov2017cnnprom}
Ramzan~Kh. Umarov and Victor~V. Solovyev.
\newblock Recognition of prokaryotic and eukaryotic promoters using
  convolutional deep learning neural networks.
\newblock {\em {PLOS} {ONE}}, 12(2):e0171410, February 2017.

\bibitem{vaswani2017transformer}
Ashish Vaswani, Noam Shazeer, Niki Parmar, Jakob Uszkoreit, Llion Jones,
  Aidan~N Gomez, \L~ukasz Kaiser, and Illia Polosukhin.
\newblock Attention is all you need.
\newblock In I.~Guyon, U.~V. Luxburg, S.~Bengio, H.~Wallach, R.~Fergus,
  S.~Vishwanathan, and R.~Garnett, editors, {\em Advances in Neural Information
  Processing Systems}, volume~30. Curran Associates, Inc., 2017.

\bibitem{wang2020sustainnlp}
Alex Wang and Thomas Wolf.
\newblock Overview of the {S}ustai{NLP} 2020 shared task.
\newblock In {\em Proceedings of SustaiNLP: Workshop on Simple and Efficient
  Natural Language Processing}, pages 174--178, Online, November 2020.
  Association for Computational Linguistics.

\bibitem{wang2020linformer}
Sinong Wang, Belinda Li, Madian Khabsa, Han Fang, and Hao Ma.
\newblock Linformer: Self-attention with linear complexity, 2020.
\newblock cite arxiv:2006.04768.

\bibitem{wang2013nmf}
Yu-Xiong Wang and Yu-Jin Zhang.
\newblock Nonnegative matrix factorization: A comprehensive review.
\newblock {\em IEEE Transactions on Knowledge and Data Engineering},
  25(6):1336--1353, 2013.

\bibitem{wilie2020indonlu}
Bryan Wilie, Karissa Vincentio, Genta~Indra Winata, Samuel Cahyawijaya,
  Xiaohong Li, Zhi~Yuan Lim, Sidik Soleman, Rahmad Mahendra, Pascale Fung,
  Syafri Bahar, and Ayu Purwarianti.
\newblock {I}ndo{NLU}: Benchmark and resources for evaluating {I}ndonesian
  natural language understanding.
\newblock In {\em Proceedings of the 1st Conference of the Asia-Pacific Chapter
  of the Association for Computational Linguistics and the 10th International
  Joint Conference on Natural Language Processing}, pages 843--857, Suzhou,
  China, December 2020. Association for Computational Linguistics.

\bibitem{winata2021multilingual}
Genta~Indra Winata.
\newblock Multilingual transfer learning for code-switched language and speech
  neural modeling.
\newblock {\em arXiv preprint arXiv:2104.06268}, 2021.

\bibitem{winata2020lrt}
Genta~Indra Winata, Samuel Cahyawijaya, Zhaojiang Lin, Zihan Liu, and Pascale
  Fung.
\newblock Lightweight and efficient end-to-end speech recognition using
  low-rank transformer.
\newblock {\em ICASSP 2020 - 2020 IEEE International Conference on Acoustics,
  Speech and Signal Processing (ICASSP)}, pages 6144--6148, 2020.

\bibitem{winata2020mtl}
Genta~Indra Winata, Samuel Cahyawijaya, Zhaojiang Lin, Zihan Liu, Peng Xu, and
  Pascale Fung.
\newblock Meta-transfer learning for code-switched speech recognition.
\newblock pages 3770--3776, July 2020.

\bibitem{winata2021mme}
Genta~Indra Winata, Samuel Cahyawijaya, Zihan Liu, Zhaojiang Lin, Andrea
  Madotto, and Pascale Fung.
\newblock Are multilingual models effective in code-switching?
\newblock In {\em Proceedings of the Fifth Workshop on Computational Approaches
  to Linguistic Code-Switching}, pages 142--153, Online, June 2021. Association
  for Computational Linguistics.

\bibitem{winata2021multilingualcodeswitch}
Genta~Indra Winata, Samuel Cahyawijaya, Zihan Liu, Zhaojiang Lin, Andrea
  Madotto, and Pascale Fung.
\newblock Are multilingual models effective in code-switching?
\newblock In {\em Proceedings of the Fifth Workshop on Computational Approaches
  to Linguistic Code-Switching}, pages 142--153, Online, June 2021. Association
  for Computational Linguistics.

\bibitem{winata2020fa-casr}
Genta~Indra Winata, Samuel Cahyawijaya, Zihan Liu, Zhaojiang Lin, Andrea
  Madotto, Peng Xu, and Pascale Fung.
\newblock {Learning Fast Adaptation on Cross-Accented Speech Recognition}.
\newblock In {\em Proc. Interspeech 2020}, pages 1276--1280, 2020.

\bibitem{xu2021networkpruning}
Dongkuan Xu, Ian En-Hsu Yen, Jinxi Zhao, and Zhibin Xiao.
\newblock Rethinking network pruning {--} under the pre-train and fine-tune
  paradigm.
\newblock In {\em Proceedings of the 2021 Conference of the North American
  Chapter of the Association for Computational Linguistics: Human Language
  Technologies}, pages 2376--2382, Online, June 2021. Association for
  Computational Linguistics.

\bibitem{yin2019dhistone}
Qijin Yin, Mengmeng Wu, Qiao Liu, Hairong Lv, and Rui Jiang.
\newblock Deephistone: a deep learning approach to predicting histone
  modifications.
\newblock {\em BMC Genomics}, 20(2):193, Apr 2019.

\bibitem{zaheer2020bigbird}
Manzil Zaheer, Guru Guruganesh, Kumar~Avinava Dubey, Joshua Ainslie, Chris
  Alberti, Santiago Ontanon, Philip Pham, Anirudh Ravula, Qifan Wang, Li~Yang,
  and Amr Ahmed.
\newblock Big bird: Transformers for longer sequences.
\newblock In H.~Larochelle, M.~Ranzato, R.~Hadsell, M.~F. Balcan, and H.~Lin,
  editors, {\em Advances in Neural Information Processing Systems}, volume~33,
  pages 17283--17297. Curran Associates, Inc., 2020.

\bibitem{zhang020ternarybert}
Wei Zhang, Lu~Hou, Yichun Yin, Lifeng Shang, Xiao Chen, Xin Jiang, and Qun Liu.
\newblock {T}ernary{BERT}: Distillation-aware ultra-low bit {BERT}.
\newblock In {\em Proceedings of the 2020 Conference on Empirical Methods in
  Natural Language Processing (EMNLP)}, pages 509--521, Online, November 2020.
  Association for Computational Linguistics.

\bibitem{zhou2015dsea}
Jian Zhou and Olga~G Troyanskaya.
\newblock Predicting effects of noncoding variants with deep
  learning{\textendash}based sequence model.
\newblock {\em Nature Methods}, 12(10):931--934, August 2015.

\bibitem{zhou2019apoe}
Xiaopu Zhou, , Yu~Chen, Kin~Y. Mok, Timothy C.~Y. Kwok, Vincent C.~T. Mok,
  Qihao Guo, Fanny~C. Ip, Yuewen Chen, Nandita Mullapudi, Paola
  Giusti-Rodr{\'{\i}}guez, Patrick~F. Sullivan, John Hardy, Amy K.~Y. Fu, Yun
  Li, and Nancy~Y. Ip.
\newblock Non-coding variability at the {APOE} locus contributes to the
  alzheimer's risk.
\newblock {\em Nature Communications}, 10(1), July 2019.

\end{thebibliography}
